\pdfoutput=1
\documentclass[11pt]{article}

\usepackage[final]{acl}

\usepackage{times}
\usepackage{latexsym}

\usepackage[T1]{fontenc}
\usepackage[utf8]{inputenc}

\usepackage{microtype}

\usepackage{inconsolata}

\usepackage{graphicx}
\usepackage{booktabs}
\usepackage{multirow}
\usepackage{makecell}
\usepackage{colortbl}
\usepackage{url}
\usepackage{subcaption}

\usepackage[normalem]{ulem}
\useunder{\uline}{\ul}{}

\title{A Comprehensive Evaluation of Quantization Strategies\\ for Large Language Models}

\author{
Renren Jin$^{1}$\thanks{Equal contribution.}\thanks{Work done during internship at Xiaomi AI Lab.}, Jiangcun Du$^{1}$\footnotemark[1], Wuwei Huang$^{2}$, Wei Liu$^{2}$,\\ 
\textbf{Jian Luan}$^{2}$\textbf{,} \textbf{Bin Wang}$^{2}$\textbf{,} \textbf{Deyi Xiong}$^{1}$\thanks{~Corresponding author.}\\
$^1$College of Intelligence and Computing, Tianjin University, Tianjin, China\\
$^{2}$Xiaomi AI Lab, Beijing, China\\
\texttt{\{rrjin, d2000, dyxiong\}@tju.edu.cn}\\
\texttt{\{huangwuwei, liuwei40, luanjian, wangbin11\}@xiaomi.com}\\
}

\begin{document}
\maketitle
\begin{abstract}
Increasing the number of parameters in large language models (LLMs) usually improves performance in downstream tasks but raises compute and memory costs, making deployment difficult in resource-limited settings. Quantization techniques, which reduce the bits needed for model weights or activations with minimal performance loss, have become popular due to the rise of LLMs. However, most quantization studies use pre-trained LLMs, and the impact of quantization on instruction-tuned LLMs and the relationship between perplexity and benchmark performance of quantized LLMs are not well understood. Evaluation of quantized LLMs is often limited to language modeling and a few classification tasks, leaving their performance on other benchmarks unclear. To address these gaps, we propose a structured evaluation framework consisting of three critical dimensions: (1) knowledge \& capacity, (2) alignment, and (3) efficiency, and conduct extensive experiments across ten diverse benchmarks. Our experimental results indicate that LLMs with 4-bit quantization can retain performance comparable to their non-quantized counterparts, and perplexity can serve as a proxy metric for quantized LLMs on most benchmarks. Furthermore, quantized LLMs with larger parameter scales can outperform smaller LLMs. Despite the memory savings achieved through quantization, it can also slow down the inference speed of LLMs. Consequently, substantial engineering efforts and hardware support are imperative to achieve a balanced optimization of decoding speed and memory consumption in the context of quantized LLMs.

\end{abstract}

\begin{figure}[!ht]
\centering
\includegraphics[width=0.8\linewidth]{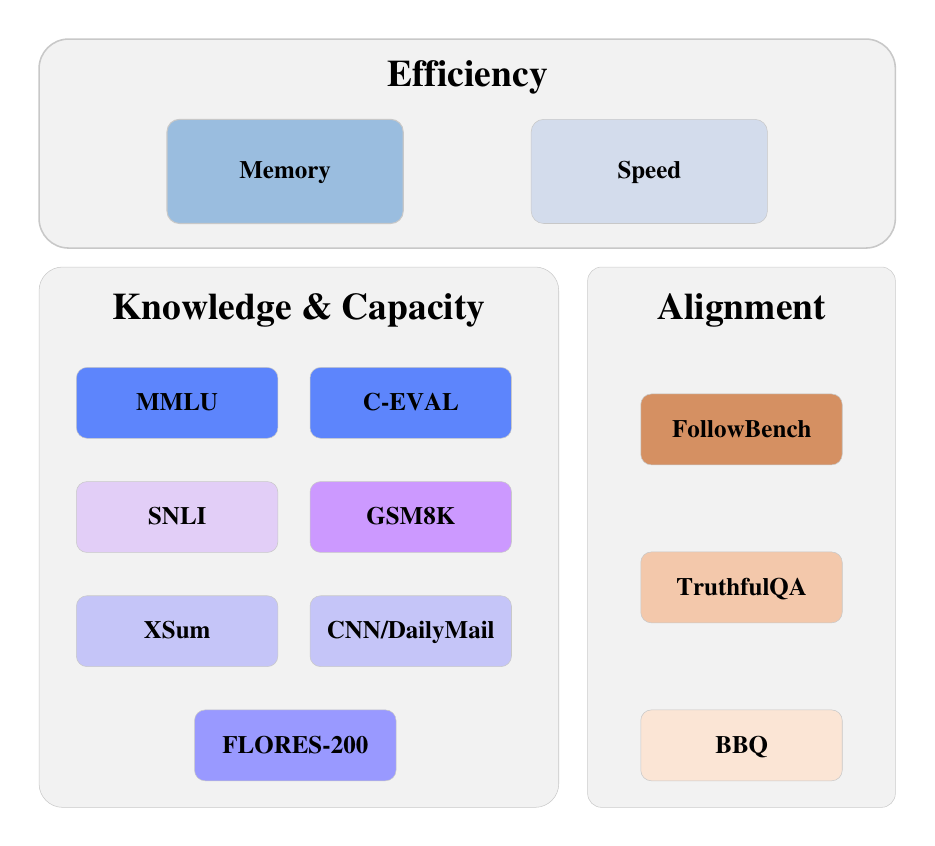} 
\caption{The evaluation framework employed in our study to assess the quantized LLMs from three key dimensions: efficiency, knowledge \& capacity and alignment.}
\label{quant_eval_overview}
\end{figure}

\section{Introduction}
In recent years, LLMs have seen substantial growth in the number of parameters, scaling up to billions or even trillions \citep{DBLP:conf/nips/BrownMRSKDNSSAA20,DBLP:conf/icml/DuHDTLXKZYFZFBZ22,DBLP:journals/corr/abs-2211-05100,DBLP:journals/corr/abs-2302-13971,DBLP:journals/corr/abs-2307-09288,DBLP:journals/corr/abs-2303-10845}, yielding exceptional performance across various tasks and real-world applications \citep{DBLP:journals/corr/abs-2303-18223,DBLP:conf/acl/LaskarBRBJH23,DBLP:journals/corr/abs-2302-04023,DBLP:conf/emnlp/LaiNVMDBN23,DBLP:journals/corr/abs-2308-12488,DBLP:journals/corr/abs-2211-09110,DBLP:conf/coling/ZhuCX24,DBLP:journals/corr/abs-2310-19736}. However, the huge number of parameters also results in significant compute and memory requirements, hindering their deployment on devices with limited resources. To mitigate these challenges, researchers have proposed various approaches to model quantization, which aim to optimize model inference and memory usage while minimizing performance degradation.

\begin{table*}[!ht]
\centering
\resizebox{\textwidth}{!}{
\begin{tabular}{clcccccc}
    \toprule
    \textbf{Category} & \textbf{Benchmarks} & \textbf{Split} & \textbf{\#Samples} & \textbf{Languages} & \textbf{Evaluation Dimension} & \textbf{Metrics} & \textbf{Evaluation Methods} \\
    \midrule
    \multicolumn{1}{c}{\multirow{7}{*}{Knowledge \& Capacity}} & MMLU \citep{DBLP:conf/iclr/HendrycksBBZMSS21} & Test & 14,042 & English & Knowledge & Accuracy $\uparrow$ & Rule-based \\
    \arrayrulecolor{lightgray}\cmidrule{2-8}
    \multicolumn{1}{c}{} & C-EVAL \citep{huang2023ceval} & Test & 12,342 & Chinese & Knowledge & Accuracy $\uparrow$ & Rule-based \\
    \arrayrulecolor{lightgray}\cmidrule{2-8}
    \multicolumn{1}{c}{} & FLORES-200 \citep{DBLP:journals/corr/abs-2207-04672} & Test & 1,012 & English, Chinese & Translation & BLEU $\uparrow$ & Rule-based \\
    \arrayrulecolor{lightgray}\cmidrule{2-8}
    \multicolumn{1}{c}{} & CNN/DailyMail \citep{DBLP:conf/acl/SeeLM17} & Test & 11,490 & English & Summarization & ROUGE $\uparrow$ & Rule-based \\
    \arrayrulecolor{lightgray}\cmidrule{2-8}
    \multicolumn{1}{c}{} & XSum \citep{DBLP:conf/emnlp/NarayanCL18} & Test & 11,334 & English & Summarization & ROUGE $\uparrow$ & Rule-based \\
    \arrayrulecolor{lightgray}\cmidrule{2-8}
    \multicolumn{1}{c}{} & GSM8K \citep{DBLP:journals/corr/abs-2110-14168} & Test & 1,319 & English & Mathematical Reasoning & Accuracy $\uparrow$ & Rule-based \\
    \arrayrulecolor{lightgray}\cmidrule{2-8}
    \multicolumn{1}{c}{} & SNLI \citep{DBLP:conf/emnlp/BowmanAPM15} & Test & 10,000 & English & Language Understanding & Accuracy $\uparrow$ & Rule-based \\
    \arrayrulecolor{black}\midrule
    \multicolumn{1}{c}{\multirow{3}{*}{Alignment}} & FollowBench \citep{DBLP:journals/corr/abs-2310-20410} & Test & 820 & English & Instruction Following & \makecell{Hard Satisfaction Rate (HSR) $\uparrow$\\Soft Satisfaction Rate (SSR) $\uparrow$\\Consistent Satisfaction Levels (CSL) $\uparrow$} & \makecell{Rule-based\\GPT-4-as-a-judge} \\
    \arrayrulecolor{lightgray}\cmidrule{2-8}
    \multicolumn{1}{c}{} & TruthfulQA \citep{DBLP:conf/acl/LinHE22} & Test & 817 & English & Truthfulness & Accuracy $\uparrow$ & Rule-based \\
    \arrayrulecolor{lightgray}\cmidrule{2-8}
    \multicolumn{1}{c}{} & BBQ \citep{DBLP:conf/acl/ParrishCNPPTHB22} & Test & 58,492 & English & Social biases & Bias Score $\rightarrow 0 \leftarrow$ & Rule-based \\
    \arrayrulecolor{black}\bottomrule
\end{tabular}
}
\caption{Comprehensive overview of benchmarks used in our evaluation experiments.}
\label{tab:benchmark_overview}
\end{table*}

The central idea of model quantization is representing the weights or activations of a model in a lower-precision format (such as 8-bit integers) rather than their original high-precision floating-point format (typically 16-bit or 32-bit) \citep{DBLP:journals/corr/abs-2103-13630,DBLP:journals/corr/abs-2308-07633}. Quantization approaches can be broadly classified into two primary categories: quantization-aware training (QAT) and post-training quantization (PTQ). QAT incorporates the quantization process into the training phase of the model, thereby allowing the model to adapt to lower-precision representations \citep{DBLP:journals/corr/abs-2305-17888,DBLP:journals/corr/abs-2305-14314,DBLP:journals/corr/abs-2305-14152}. Conversely, PTQ applies quantization techniques after the training phase has finished \citep{DBLP:conf/nips/DettmersLBZ22,DBLP:journals/corr/abs-2210-17323,DBLP:journals/corr/abs-2306-00978,DBLP:journals/corr/abs-2306-02272,DBLP:journals/corr/abs-2306-03078,DBLP:conf/icml/XiaoLSWDH23,DBLP:conf/nips/YaoAZWLH22}.

Despite the risk of performance degradation, PTQ is more prevalent due to the prohibitive training costs associated with QAT. However, several aspects pertaining to the evaluation of PTQ require further exploration. Firstly, the majority of PTQ methods are evaluated solely by assessing the performance of the quantized pre-trained LLMs on benchmarks, leaving the performance of quantized LLMs that have undergone instruction tuning unclear - despite the latter being more commonly used in real-world scenarios \citep{DBLP:conf/nips/Ouyang0JAWMZASR22,DBLP:journals/corr/abs-2204-05862,DBLP:journals/corr/abs-2304-03277}. Secondly, the evaluation of quantized models is limited to the language modeling task and a few classification tasks. This restricts our understanding of their performance on other benchmarks that are more closely related to real-world applications. Lastly, while perplexity is predominantly employed as the evaluation metric for verifying the effectiveness of quantization methods and has been demonstrated as an indicator of the performance of LLMs on other benchmarks in previous studies \citep{DBLP:conf/acl/XiaAZLPCZS23}, the correlation between the perplexity of quantized LLMs and their performance on other benchmarks remains poorly understood.

In this paper, we conduct a comprehensive evaluation of the quantized LLMs that undergo instruction tuning, utilizing a diverse range of publicly available benchmarks. These benchmarks cover language understanding and generation, as well as two critical dimensions of LLMs: knowledge \& capacity and alignment. Additionally, we evaluate various quantization strategies for their efficiency in terms of generation speed and memory consumption. The comprehensive framework for this evaluation is illustrated in Figure~\ref{quant_eval_overview}, while Table~\ref{tab:benchmark_overview} provides a detailed summary of the benchmarks employed in our experiments.\footnote{Our code is publicly available at \url{https://github.com/cordercorder/quant_eval}.}

Our contributions can be summarized as follows:

\begin{itemize}
    \item We propose a structured evaluation framework and conduct extensive experiments to evaluate instruction-tuned LLMs and their quantized counterparts employing various quantization strategies across different parameter scales (7B, 14B, 72B). 
    \item Our empirical findings suggest that LLMs utilizing 4-bit quantization can maintain performance comparable to their non-quantized counterparts on the evaluated benchmarks. Additionally, quantized LLMs with a larger parameter scale demonstrate superior performance compared to their non-quantized counterparts with smaller parameter sizes. Furthermore, we find that perplexity serves as a reliable performance indicator for quantized LLMs across the majority of the benchmarks.
    \item We identify isolating outlier weights as a key factor enabling SpQR to effectively quantize LLMs to an extreme 2-bit level, significantly outperforming GPTQ at the same level.
    \item Despite the impressive performance of contemporary quantization approaches, our further analysis reveals substantial engineering challenges. Specifically, these approaches require significant engineering effort and hardware support to be effectively applied in practical scenarios, particularly in terms of memory and speed requirements.
\end{itemize}

\section{Related Work}

\paragraph{LLMs Quantization}
There are currently two main formalisms of model quantization: QAT \citep{DBLP:conf/cvpr/JacobKCZTHAK18} and PTQ. 

PTQ applies quantization after model training, while QAT considers the effects of quantization during the training process, necessitating considerable resources and expertise, thereby restricting its broader application. Consequently, our research primarily concentrates on PTQ.
 
Concerning the identification and protection of outlier values, GPT3.int8() \citep{DBLP:conf/nips/DettmersLBZ22} (also known as LLM.int8()) identifies outliers by magnitude while SpQR \citep{DBLP:journals/corr/abs-2306-03078} employs Hessian matrix to identify outlier values.
By equivalently scaling weights and activation values, SmoothQuant \citep{DBLP:conf/icml/XiaoLSWDH23} greatly reduces quantization error of activation and thus results in a great reduction in the quantization loss of the model. 
Outlier Suppression+ \citep{DBLP:conf/emnlp/WeiZLZGG023} suppresses the outlier of weights by performing channel-wise shift and scale. 
QLoRA \citep{DBLP:journals/corr/abs-2305-14314} proposes to use the NF4 data format to reduce quantization rounding errors further.
OPTQ \citep{DBLP:conf/iclr/FrantarAHA23} (generally known as GPTQ) adjusting the weights during the quantization process to reduce quantization errors.

\paragraph{LLMs Evaluation}

As the technology behind LLMs continues to advance, these models have shown remarkable performance in many tasks \citep{DBLP:journals/corr/abs-2302-04023,DBLP:journals/corr/abs-2308-12488}, sometimes surpassing human proficiency \citep{DBLP:journals/corr/abs-2206-04615,DBLP:conf/acl/LaskarBRBJH23}. Additionally, as the number of parameters in these models increases, they exhibit emergent abilities \citep{DBLP:journals/tmlr/WeiTBRZBYBZMCHVLDF22,schaeffer2023are,DBLP:journals/corr/abs-2307-08072,DBLP:journals/corr/abs-2309-01809,DBLP:journals/corr/abs-2310-03262}, making it challenging to compare their performance to that of other models and understand their behavior. As a result, numerous benchmarks have been curated to rigorously assess the performance of LLMs \citep{10.1145/3641289,ziyu-etal-2023-lens,DBLP:journals/corr/abs-2308-05374,DBLP:conf/coling/Yu0X24}. These benchmarks can be divided into two primary categories: (1) knowledge \& capacity evaluation \citep{DBLP:conf/iclr/HendrycksBBZMSS21,DBLP:journals/corr/abs-2306-09212,DBLP:journals/corr/abs-2304-12986,huang2023ceval,DBLP:journals/corr/abs-2307-16789,DBLP:journals/corr/abs-2403-07747,DBLP:conf/coling/LiuJRX24,DBLP:journals/corr/abs-2312-16132,DBLP:journals/corr/abs-2305-10263,DBLP:conf/aaai/ShiYHLX24}, which examines the model’s ability to understand and generate correct responses; and (2) alignment evaluation, which measures how well the model’s outputs align with human preference and values \citep{DBLP:conf/emnlp/GehmanGSCS20,DBLP:conf/acl/LinHE22,DBLP:conf/acl/ParrishCNPPTHB22,DBLP:journals/corr/abs-2306-16244,DBLP:journals/corr/abs-2311-18743,DBLP:conf/acl/YinSGWQH23,DBLP:journals/corr/abs-2311-07911}. Although these benchmarks are commonly employed to assess LLMs, their quantized counterparts are often excluded from these evaluations. As a result, it can be challenging to comprehend the behavior of quantized LLMs and determine the extent of the performance gap between them and their non-quantized counterparts.

In addressing these challenges, our research primarily focuses on evaluating quantized LLMs. Our goal is to conduct a thorough examination of the performance of LLMs that have been quantized using various methods. In doing so, we hope to yield valuable insights that will inform and enhance future advancements in quantization methodologies.

\section{Evaluation Protocol}
The comprehensive evaluation of LLMs presents a long-standing challenge due to their versatility, widespread application, and poor explainability. To address this, we propose a structured evaluation framework that encompasses three critical dimensions: (1) knowledge and capacity, (2) alignment, and (3) efficiency. 

For the evaluation of knowledge and capacity, we consider two types of benchmarks: (i) those requiring LLMs to demonstrate extensive knowledge across various domains to achieve satisfactory performance, and (ii) those assessing the ability of LLMs to perform specific tasks such as language generation and understanding. In this context, we employ the MMLU \citep{DBLP:conf/iclr/HendrycksBBZMSS21} and C-EVAL \citep{huang2023ceval} benchmarks for the former, covering diverse subjects including but not limited to history, chemistry, and economics. For the latter, we select the FLORES-200 \citep{DBLP:journals/corr/abs-2207-04672}, CNN/DailyMail \citep{DBLP:conf/acl/SeeLM17}, and XSum \citep{DBLP:conf/emnlp/NarayanCL18} benchmarks, which focus on essential language generation tasks like translation and summarization, and the GSM8K \citep{DBLP:journals/corr/abs-2110-14168} and SNLI \citep{DBLP:conf/emnlp/BowmanAPM15} benchmarks for evaluating language understanding and reasoning capabilities.

\begin{table*}[!ht]
\centering
\resizebox{\textwidth}{!}{
\begin{tabular}{ccccccccccc}
    \toprule
    \textbf{Model} & \textbf{Datatype} & \textbf{Quantization Method} & \textbf{Average Accuracy} & \textbf{Average BLEU} & \textbf{Average ROUGE-1/ROUGE-2/ROUGE-3} & \textbf{Average HSR/SSR/CSL} & \textbf{Average Bias Score} & \textbf{Average Perplexity} & \textbf{Memory} & \textbf{Speed} \\
    \midrule
    \multicolumn{1}{c}{\multirow{10}{*}{Qwen-7B-Chat}} & BFloat16 & - & 57.10 & 29.63 & 0.257/0.086/0.168 & 40.23/51.71/1.57 & 6.20/3.87 & 11.76 & 15.14 & 37.67 \\
    \cmidrule{2-11}
    \multicolumn{1}{c}{} & \multicolumn{1}{c}{\multirow{3}{*}{INT-8}} & LLM.int8() & 56.67 & 28.97 & 0.256/0.086/0.168 & 40.52/52.38/1.62 & 6.41/3.49 & 11.77 & 9.23 & 7.19 \\
    \multicolumn{1}{c}{} & \multicolumn{1}{c}{} & GPTQ & 57.21 & 29.52 & 0.257/0.087/0.169 & 40.78/53.00/1.52 & 5.98/3.76 & 11.76 & 10.91 & 13.57 \\
    \multicolumn{1}{c}{} & \multicolumn{1}{c}{} & SpQR & 56.49 & 29.51 & 0.257/0.086/0.168 & 40.20/52.10/1.62 & 6.36/3.95 & 11.83 & 15.60 & 37.65 \\
    \cmidrule{2-11}
    \multicolumn{1}{c}{} & \multicolumn{1}{c}{\multirow{2}{*}{INT-4}} & GPTQ & 54.86 & 28.43 & 0.254/0.084/0.167 & 39.84/52.13/1.38 & 5.49/3.69 & 12.31 & 7.83 & 37.43 \\
    \multicolumn{1}{c}{} & \multicolumn{1}{c}{} & SpQR & 56.41 & 29.59 & 0.256/0.086/0.168 & 40.26/51.59/1.48 & 6.34/3.77 & 11.97 & 15.60 & 37.73 \\
    \cmidrule{2-11}
    \multicolumn{1}{c}{} & \multicolumn{1}{c}{\multirow{2}{*}{INT-3}} & GPTQ & 51.42 & 24.22 & 0.228/0.067/0.149 & 35.82/47.77/1.27 & 4.21/4.90 & 15.10 & 7.12 & 8.21 \\
    \multicolumn{1}{c}{} & \multicolumn{1}{c}{} & SpQR & 55.45 & 28.39 & 0.253/0.083/0.166 & 36.03/49.44/1.30 & 6.31/4.08 & 13.40 & 15.61 & 37.73 \\
    \cmidrule{2-11}
    \multicolumn{1}{c}{} & \multicolumn{1}{c}{\multirow{2}{*}{INT-2}} & GPTQ & 16.52 & 0.01 & 0.042/0.000/0.029 & 0.24/0.64/0.00 & -0.54/-0.97 & 84396.73 & 6.26 & 19.36 \\
    \multicolumn{1}{c}{} & \multicolumn{1}{c}{} & SpQR & 53.18 & 27.22 & 0.242/0.077/0.158 & 37.52/49.69/1.55 & 4.13/5.76 & 13.77 & 15.66 & 37.51 \\
    \midrule
    \multicolumn{1}{c}{\multirow{10}{*}{Qwen-14B-Chat}} & BFloat16 & - & 62.92 & 31.13 & 0.254/0.085/0.196 & 53.16/62.25/2.15 & 8.35/3.69 & 9.84 & 27.60 & 25.15 \\
    \cmidrule{2-11}
    \multicolumn{1}{c}{} & \multicolumn{1}{c}{\multirow{3}{*}{INT-8}} & LLM.int8() & 62.48 & 31.63 & 0.254/0.084/0.166 & 48.35/57.69/1.75 & 7.92/3.89 & 9.86 & 15.91 & 5.85 \\
    \multicolumn{1}{c}{} & \multicolumn{1}{c}{} & GPTQ & 62.67 & 31.84 & 0.254/0.084/0.196 & 49.22/58.76/1.90 & 8.22/3.70 & 9.85 & 17.92 & 14.37 \\
    \multicolumn{1}{c}{} & \multicolumn{1}{c}{} & SpQR & 62.86 & 31.97 & 0.255/0.085/0.167 & 47.53/57.59/1.87 & 8.60/3.65 & 9.85 & 27.95 & 25.42 \\
    \cmidrule{2-11}
    \multicolumn{1}{c}{} & \multicolumn{1}{c}{\multirow{2}{*}{INT-4}} & GPTQ & 61.53 & 31.40 & 0.252/0.082/0.165 & 48.66/57.47/1.90 & 7.82/4.11 & 10.29 & 12.03 & 24.38 \\
    \multicolumn{1}{c}{} & \multicolumn{1}{c}{} & SpQR & 62.66 & 31.47 & 0.252/0.083/0.165 & 46.84/56.27/1.78 & 7.96/3.86 & 9.94 & 27.95 & 24.62 \\
    \cmidrule{2-11}
    \multicolumn{1}{c}{} & \multicolumn{1}{c}{\multirow{2}{*}{INT-3}} & GPTQ & 58.34 & 28.92 & 0.237/0.073/0.155 & 43.91/53.38/1.62 & 8.41/3.88 & 13.94 & 10.77 & 4.71 \\
    \multicolumn{1}{c}{} & \multicolumn{1}{c}{} & SpQR & 61.43 & 31.09 & 0.253/0.082/0.165 & 47.83/57.65/1.85 & 8.03/3.11 & 10.19 & 29.97 & 25.20 \\
    \cmidrule{2-11}
    \multicolumn{1}{c}{} & \multicolumn{1}{c}{\multirow{2}{*}{INT-2}} & GPTQ & 16.78 & 0.01 & 0.044/0.000/0.030 & 0.54/1.12/0.02 & -0.17/-0.81 & 192872.47 & 8.99 & 18.26 \\
    \multicolumn{1}{c}{} & \multicolumn{1}{c}{} & SpQR & 59.82 & 29.20 & 0.247/0.080/0.162 & 47.76/57.47/1.82 & 8.08/5.33 & 11.00 & 28.04 & 24.82 \\
    \midrule
    \multicolumn{1}{c}{\multirow{10}{*}{Qwen-72B-Chat}} & BFloat16 & - & 71.76 & 34.81 & 0.300/0.114/0.203 & 53.16/62.25/2.15 & 9.07/1.57 & 8.52 & 138.44 & 8.97 \\
    \cmidrule{2-11}
    \multicolumn{1}{c}{} & \multicolumn{1}{c}{\multirow{3}{*}{INT-8}} & LLM.int8() & 71.74 & 34.39 & 0.301/0.115/0.204 & 56.00/64.03/2.28 & 8.81/1.68 & 8.51 & 74.96 & 3.07 \\
    \multicolumn{1}{c}{} & \multicolumn{1}{c}{} & GPTQ & 71.20 & 34.82 & 0.300/0.114/0.204 & 54.66/63.28/2.08 & 8.95/1.31 & 8.71 & 77.85 & 1.43 \\
    \multicolumn{1}{c}{} & \multicolumn{1}{c}{} & SpQR & 71.90 & 34.67 & 0.300/0.115/0.203 & 54.27/62.67/2.33 & 9.07/1.51 & 8.54 & 143.20 & 6.57 \\
    \cmidrule{2-11}
    \multicolumn{1}{c}{} & \multicolumn{1}{c}{\multirow{2}{*}{INT-4}} & GPTQ & 71.38 & 34.03 & 0.298/0.112/0.201 & 52.81/61.43/2.13 & 10.11/1.76 & 8.77 & 44.11 & 14.88 \\
    \multicolumn{1}{c}{} & \multicolumn{1}{c}{} & SpQR & 71.76 & 34.72 & 0.299/0.114/0.201 & 53.81/62.14/2.10 & 8.73/1.52 & 8.64 & 143.21 & 6.56 \\
    \cmidrule{2-11}
    \multicolumn{1}{c}{} & \multicolumn{1}{c}{\multirow{2}{*}{INT-3}} & GPTQ & 66.89 & 30.98 & 0.269/0.096/0.178 & 52.73/61.38/2.02 & 8.77/3.11 & 10.19 & 35.93 & 0.84 \\
    \multicolumn{1}{c}{} & \multicolumn{1}{c}{} & SpQR & 70.67 & 34.08 & 0.292/0.109/0.196 & 52.82/61.42/2.22 & 8.24/1.97 & 8.84 & 143.37 & 6.57 \\
    \cmidrule{2-11}
    \multicolumn{1}{c}{} & \multicolumn{1}{c}{\multirow{2}{*}{INT-2}} & GPTQ & 18.40 & 0.01 & 0.021/0.000/0.017 & 0.08/0.48/0.00 & -0.36/0.89 & 48714.04 & 27.74 & 2.23 \\
    \multicolumn{1}{c}{} & \multicolumn{1}{c}{} & SpQR & 67.07 & 32.74 & 0.278/0.100/0.186 & 53.48/61.76/2.20 & 7.67/1.69 & 9.57 & 144.59 & 6.56 \\
    \bottomrule
\end{tabular}
}
\caption{Evaluation results of the Qwen-Chat series models and their quantized counterparts across ten benchmarks designed to evaluate LLMs in terms of knowledge \& capacity and alignment, as well as metrics for memory consumption and decoding speed during inference. The benchmarks are grouped by the type of metrics used, and the average score for each metric within its respective group is presented. The ``Average Accuracy'' represents the mean accuracy of LLMs across the MMLU \citep{DBLP:conf/iclr/HendrycksBBZMSS21}, C-EVAL \citep{huang2023ceval}, GSM8K \citep{DBLP:journals/corr/abs-2110-14168}, SNLI \citep{DBLP:conf/emnlp/BowmanAPM15}, and TruthfulQA \citep{DBLP:conf/acl/LinHE22} benchmarks. The ``Average BLEU'' indicates the mean BLEU score for Chinese-English and English-Chinese translations on the FLORES-200 benchmark \citep{DBLP:journals/corr/abs-2207-04672}. The ``Average ROUGE-1/ROUGE-2/ROUGE-L'' displays the mean ROUGE-1, ROUGE-2, and ROUGE-L scores on the XSum \citep{DBLP:conf/emnlp/NarayanCL18} and CNN/DailyMail \citep{DBLP:conf/acl/SeeLM17} benchmarks. The ``Average HSR/SSR/CSL'' represents the mean hard satisfaction rates (HSR), soft satisfaction rates (SSR) across five difficulty levels, and consistent satisfaction levels (CSL) across five constraints on the FollowBench benchmark \citep{DBLP:journals/corr/abs-2310-20410}. The ``Average Bias Score'' is shown as x/y, where x and y represent the mean bias score across various categories in ambiguous and unambiguous contexts, respectively, on the BBQ benchmark \citep{DBLP:conf/acl/ParrishCNPPTHB22}. The ``Average Perplexity'' indicates the mean perplexity on WikiText2 \citep{DBLP:conf/iclr/MerityX0S17}, C4 \citep{DBLP:journals/jmlr/RaffelSRLNMZLL20}, and PTB \citep{DBLP:conf/naacl/MarcusKMMBFKS94}. The ``Memory'' refers to the memory consumed (in GB) during inference when the input consists of 256 tokens and the output contains 512 tokens. The ``Speed'' represents the number of tokens generated per second when the input consists of 256 tokens and the output contains 512 tokens.}
\label{tab:experimental_results_overview}
\end{table*}

For the evaluation of alignment, we adopt the HHH criteria proposed by \citet{DBLP:journals/corr/abs-2112-00861}, which assess LLMs from three distinct perspectives: helpfulness, honesty, and harmlessness. Accordingly, we have chosen the FollowBench \citep{DBLP:journals/corr/abs-2310-20410}, TruthfulQA \citep{DBLP:conf/acl/LinHE22}, and BBQ \citep{DBLP:conf/acl/ParrishCNPPTHB22} benchmarks to assess these aspects, respectively.

For the evaluation of efficiency, we consider metrics such as memory usage and generation speed during inference, which are crucial for the practical application of LLMs in real-world scenarios.

It is important to note that while these three dimensions provide a comprehensive framework for evaluating LLMs, other benchmarks or metrics can also be employed as long as they align with these dimensions.

\section{Evaluation Setup}

\subsection{LLMs}
We predominantly employ quantization techniques on the Qwen-Chat series of models \citep{DBLP:journals/corr/abs-2309-16609}, which have undergone instruction tuning, taking into account the following considerations: (1) The Qwen-Chat models have demonstrated exceptional performance across a variety of tasks. (2) The Qwen-Chat series includes LLMs of varying parameter scales, specifically models with 7 billion, 14 billion, and 72 billion parameters. (3) The models in the Qwen-Chat series have been pre-trained on an extensive corpus of multilingual data, with a particular focus on Chinese and English. This extensive pre-training enables the models to support a multitude of languages beyond English.

\begin{figure*}[!t]
\centering
\begin{subfigure}{0.475\textwidth}
    \centering
    \includegraphics[width=\linewidth]{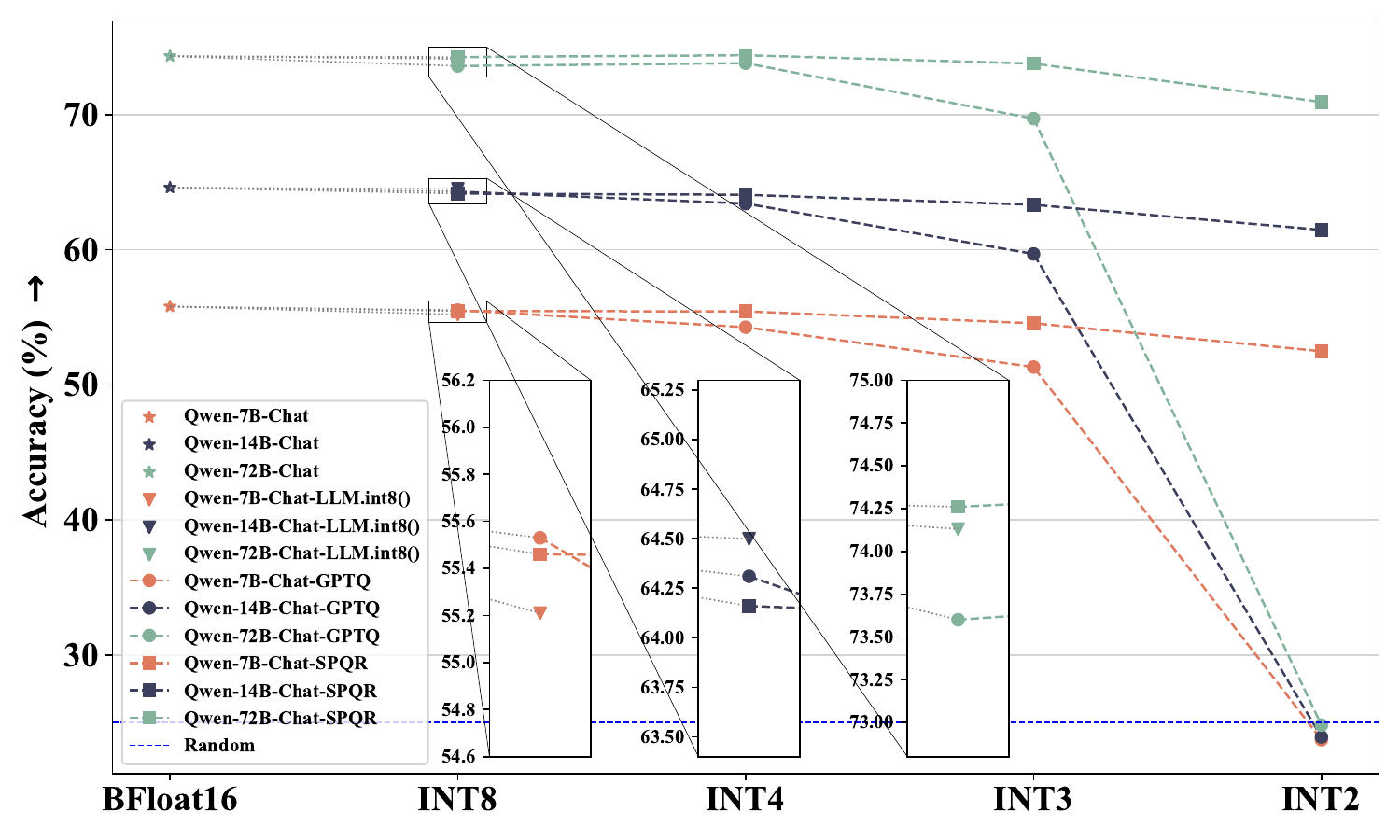}
    \caption{MMLU}
\end{subfigure}
\begin{subfigure}{0.475\textwidth}
    \centering
    \includegraphics[width=\linewidth]{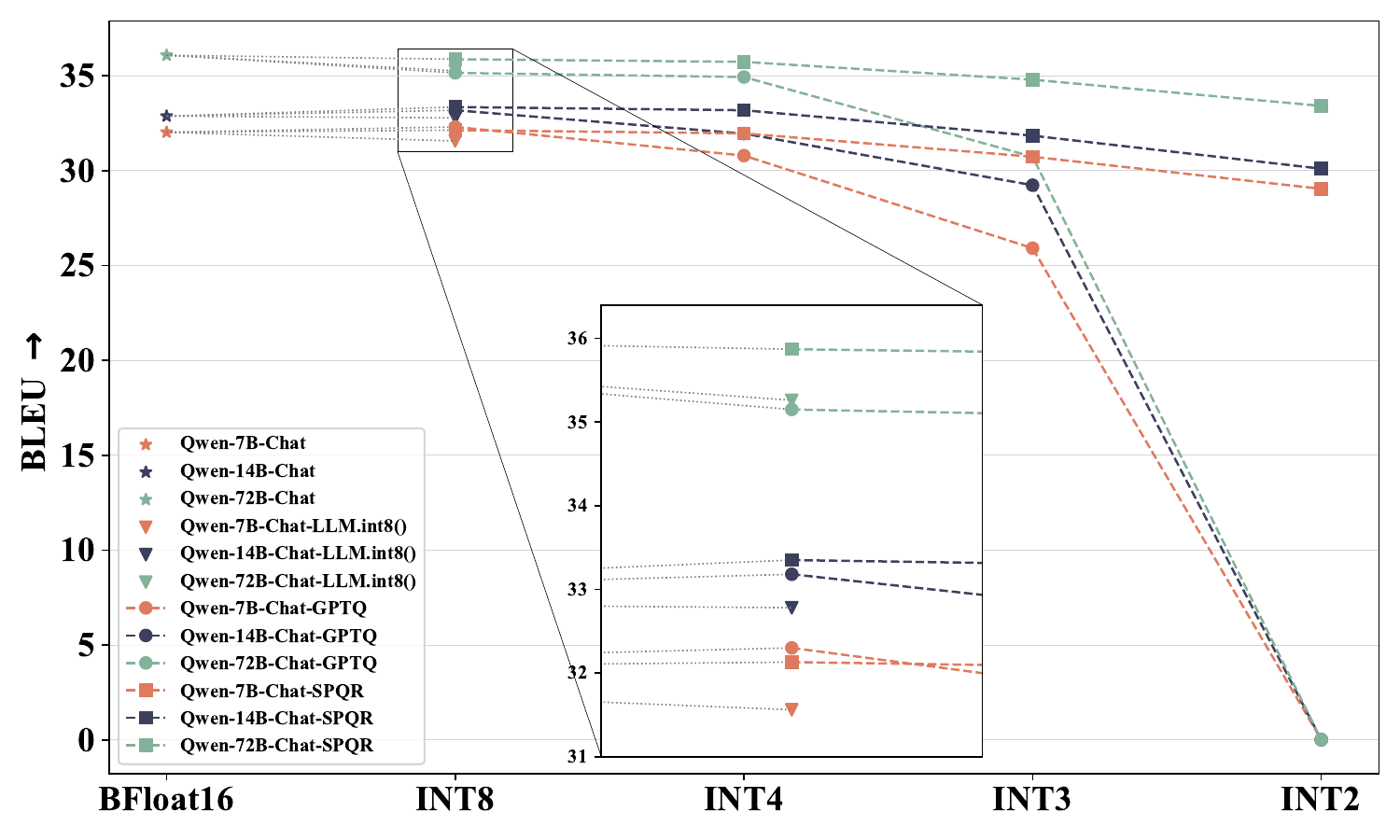}
    \caption{En $\rightarrow$ Zh}
\end{subfigure}
\caption{Performance of the Qwen-Chat series models and their quantized counterparts on the MMLU \citep{DBLP:conf/iclr/HendrycksBBZMSS21} benchmark (a) and the English-to-Chinese (En $\rightarrow$ Zh) translation task of the FLORES-200 \citep{DBLP:journals/corr/abs-2207-04672} (b) benchmark. The x-axis represents the data format of the model's weight, where $x$ in INT$x$ denotes the number of integer bits used for weight representations. To highlight the nuanced differences between LLM.int8() and other methodologies, a magnified view is integrated into the figure.}
\label{figure:mmlu_avg_flores_200_en2zh}
\end{figure*}

\begin{figure*}[!ht]
\centering
\begin{subfigure}{0.325\textwidth}
    \centering
    \includegraphics[width=\linewidth]{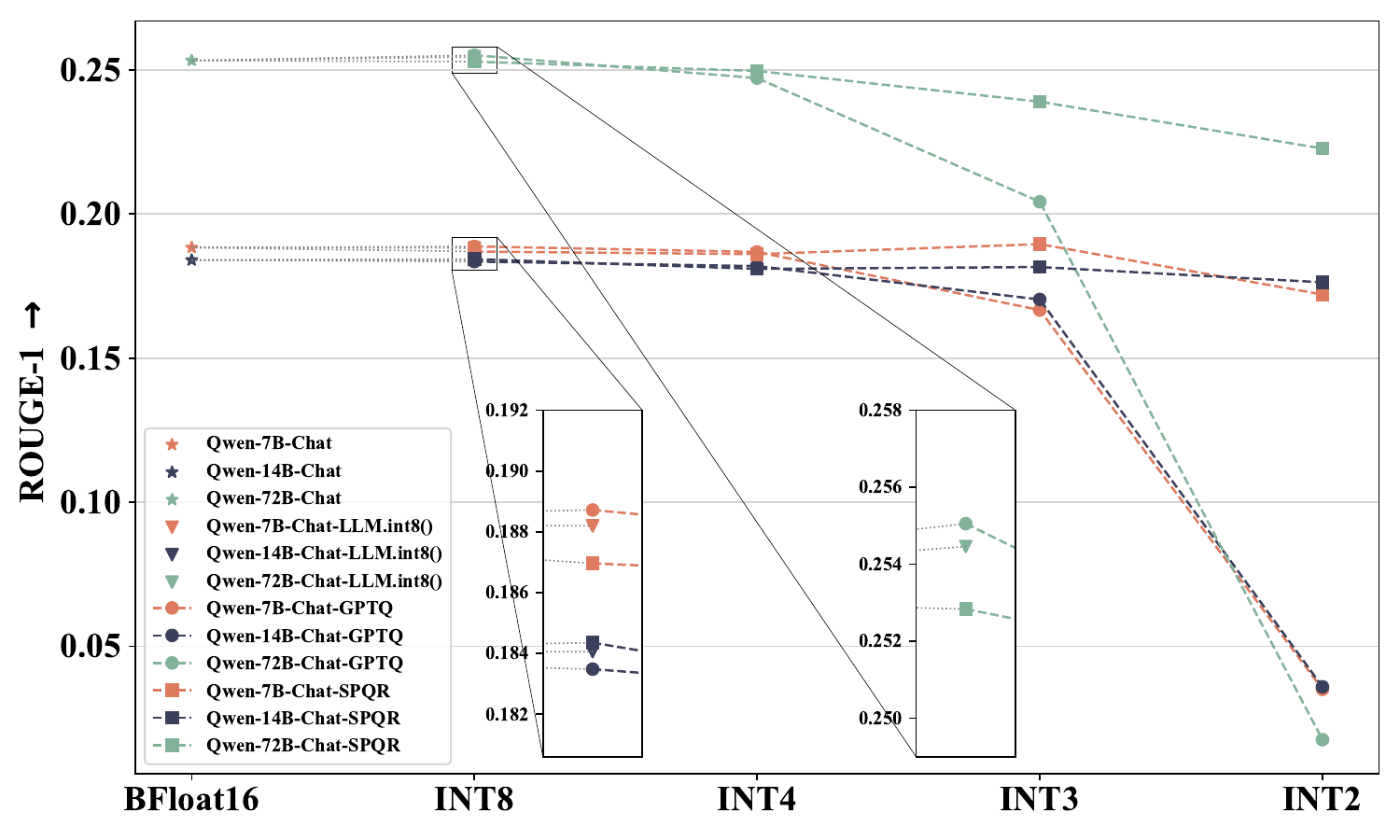}
    \caption{ROUGE-1}
\end{subfigure}
\begin{subfigure}{0.325\textwidth}
    \centering
    \includegraphics[width=\linewidth]{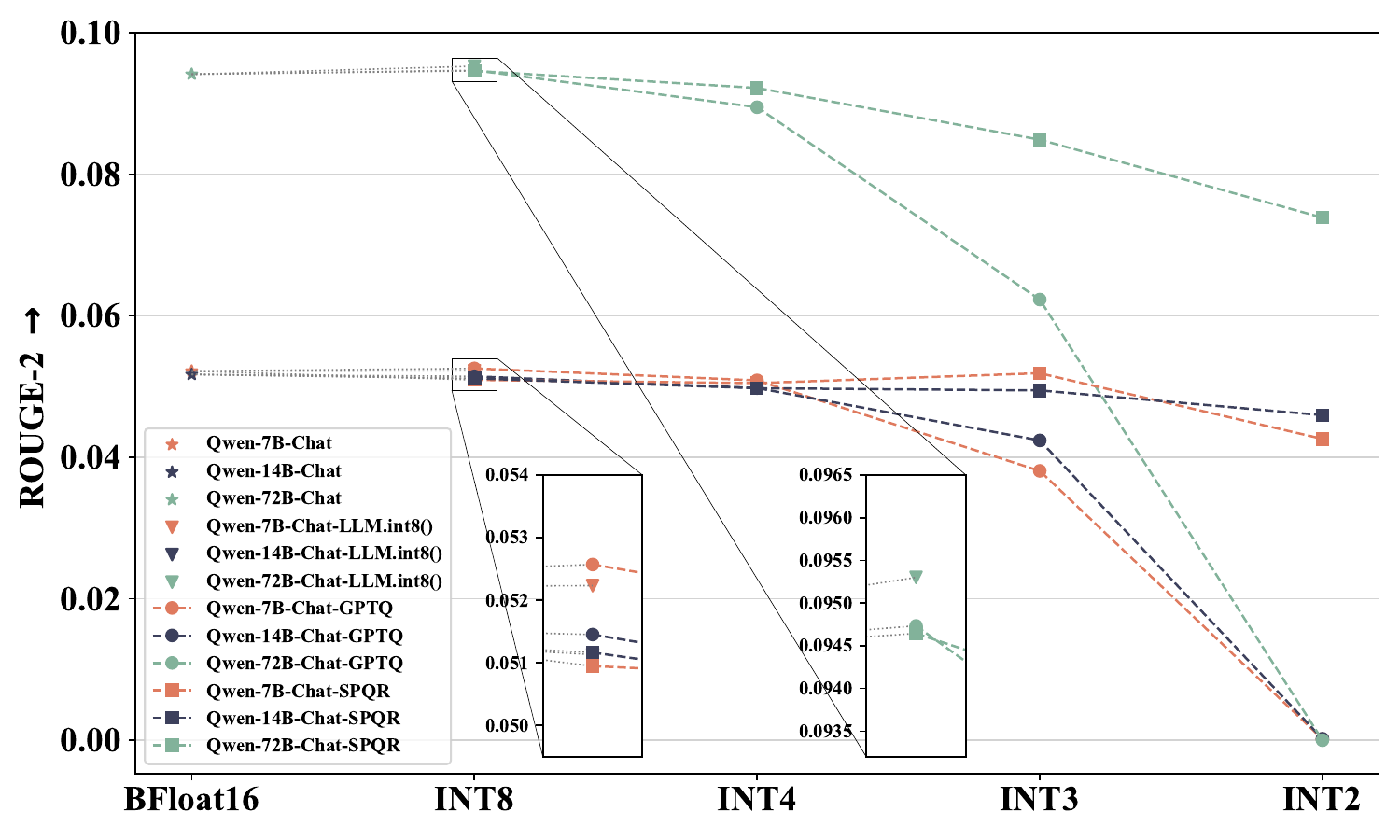}
    \caption{ROUGE-2}
\end{subfigure}
\begin{subfigure}{0.325\textwidth}
    \centering
    \includegraphics[width=\linewidth]{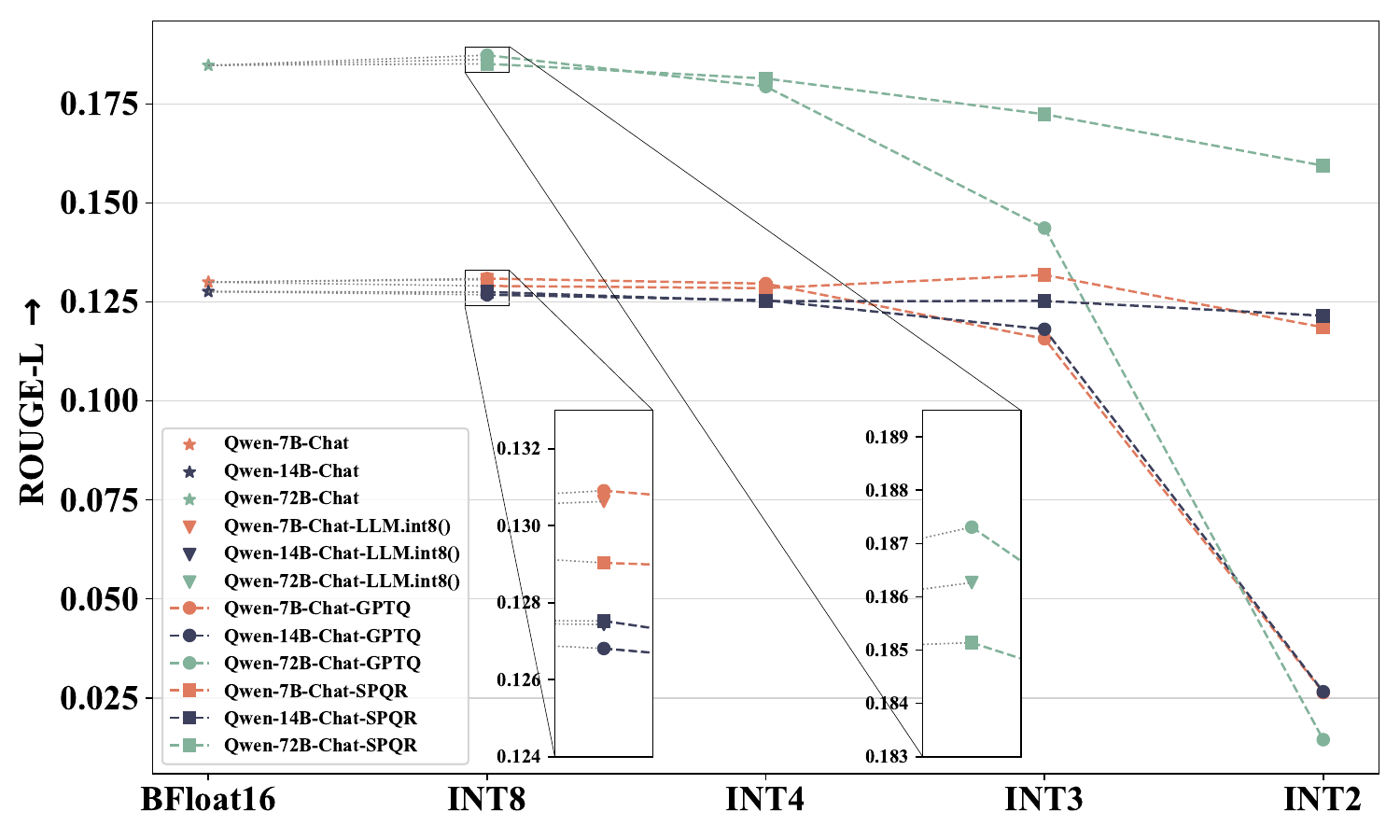}
    \caption{ROUGE-L}
\end{subfigure}
\caption{ROUGE-1 (a), ROUGE-2 (b), and ROUGE-L (c) scores for the Qwen-Chat series models and their quantized counterparts on the test sets of XSum \citep{DBLP:conf/emnlp/NarayanCL18}.}
\label{figure:xsum}
\end{figure*}

\subsection{Quantization Strategies}

We select three prominent quantization approaches accompanied by dedicated open-source implementations for evaluation: LLM.int8() \citep{DBLP:conf/nips/DettmersLBZ22},  GPTQ \citep{DBLP:conf/iclr/FrantarAHA23}, and SpQR \citep{DBLP:journals/corr/abs-2306-03078}. These approaches have been either deeply integrated into the Hugging Face Transformers\footnote{\url{https://github.com/huggingface/transformers}} library \citep{DBLP:conf/emnlp/WolfDSCDMCRLFDS20} or widely used, thereby enabling them to support a variety of open-source LLMs. Specifically, we employ GPTQ and SpQR to quantize the LLMs to 8, 4, 3, and 2 bits, respectively, except LLM.int8(), which exclusively quantizes them to 8 bits. For the calibration data required by SpQR and GPTQ, we randomly sampled 128 examples from the dataset collected by \citet{alpaca} and \citet{DBLP:journals/corr/abs-2304-03277}. For a detailed introduction to these quantization approaches, please refer to Appendix~\ref{appendix_sec:quantization_strategies}.

\subsection{Benchmarks}
We utilize ten distinct benchmarks to facilitate a comprehensive assessment of LLMs and their quantized counterparts. These benchmarks encompass knowledge \& capacity evaluation, as well as alignment evaluation. By leveraging this broad spectrum of benchmarks, we aim to gain a holistic understanding of the models' performance across various dimensions, thereby enabling a detailed comparison between the original and quantized versions of LLMs. For a comprehensive overview these benchmarks and the associated prompts employed in our study, please see Appendix~\ref{appendix_sec:benchmarks} and Appendix~\ref{appendix_sec:prompts}.

\section{Experiment Results and Discussion}
Table~\ref{tab:experimental_results_overview} presents a comprehensive performance summary of the Qwen-Chat series models and their quantized counterparts across ten benchmarks designed to evaluate LLMs in terms of knowledge \& capacity and alignment. It also includes metrics for memory consumption and decoding speed during inference. Detailed experimental results for each benchmark and metric are illustrated in Figures~\ref{figure:mmlu_avg_flores_200_en2zh} through \ref{figure:memory_and_speed} and Figures~\ref{figure:cnndm} through \ref{figure:ppl_c4_ptb} in Appendix~\ref{appendix_sec:detailed_experimental_results}.

\begin{figure*}[!ht]
\centering
\begin{subfigure}{0.325\textwidth}
    \centering
    \includegraphics[width=\linewidth]{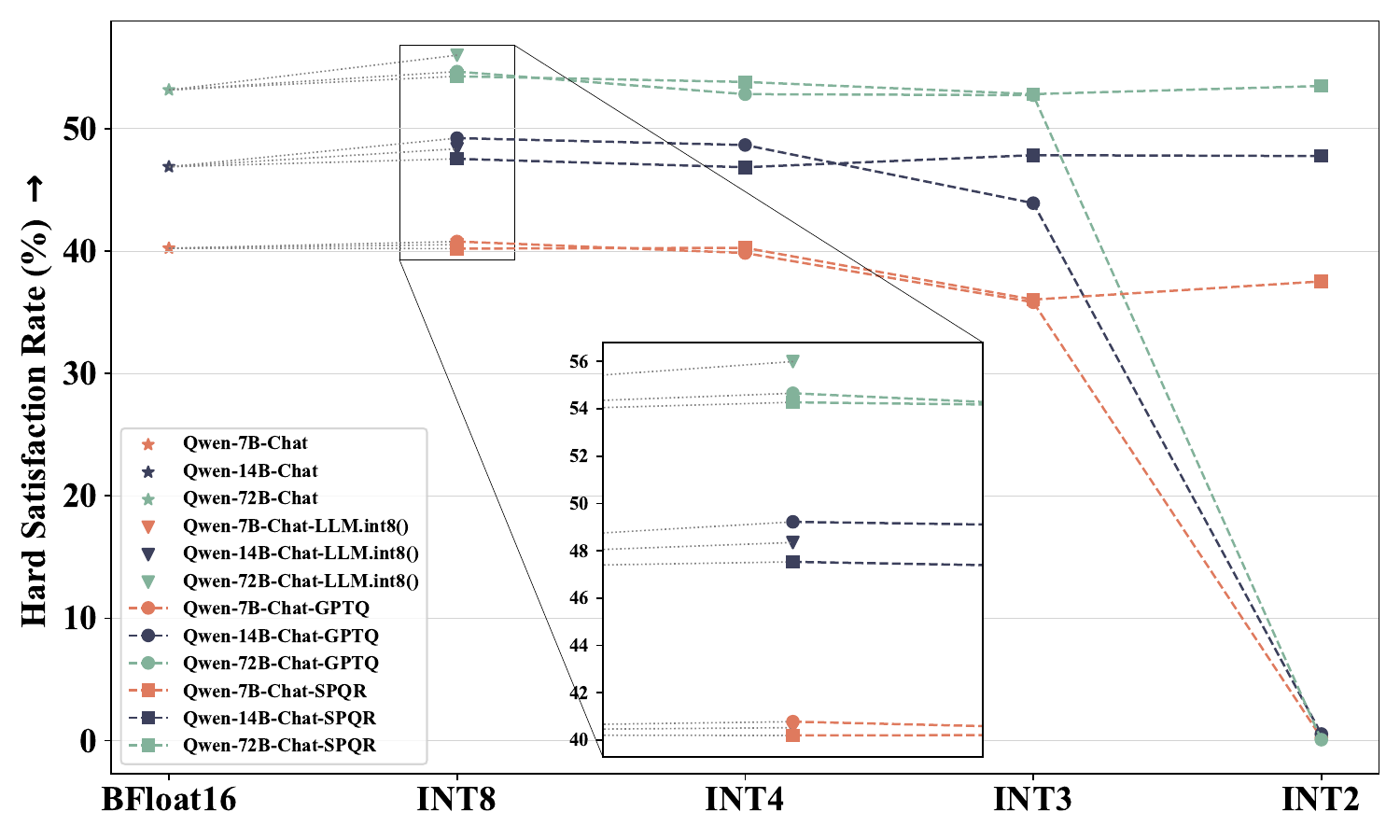}
    \caption{Hard Satisfaction Rate (HSR)}
\end{subfigure}
\begin{subfigure}{0.325\textwidth}
    \centering
    \includegraphics[width=\linewidth]{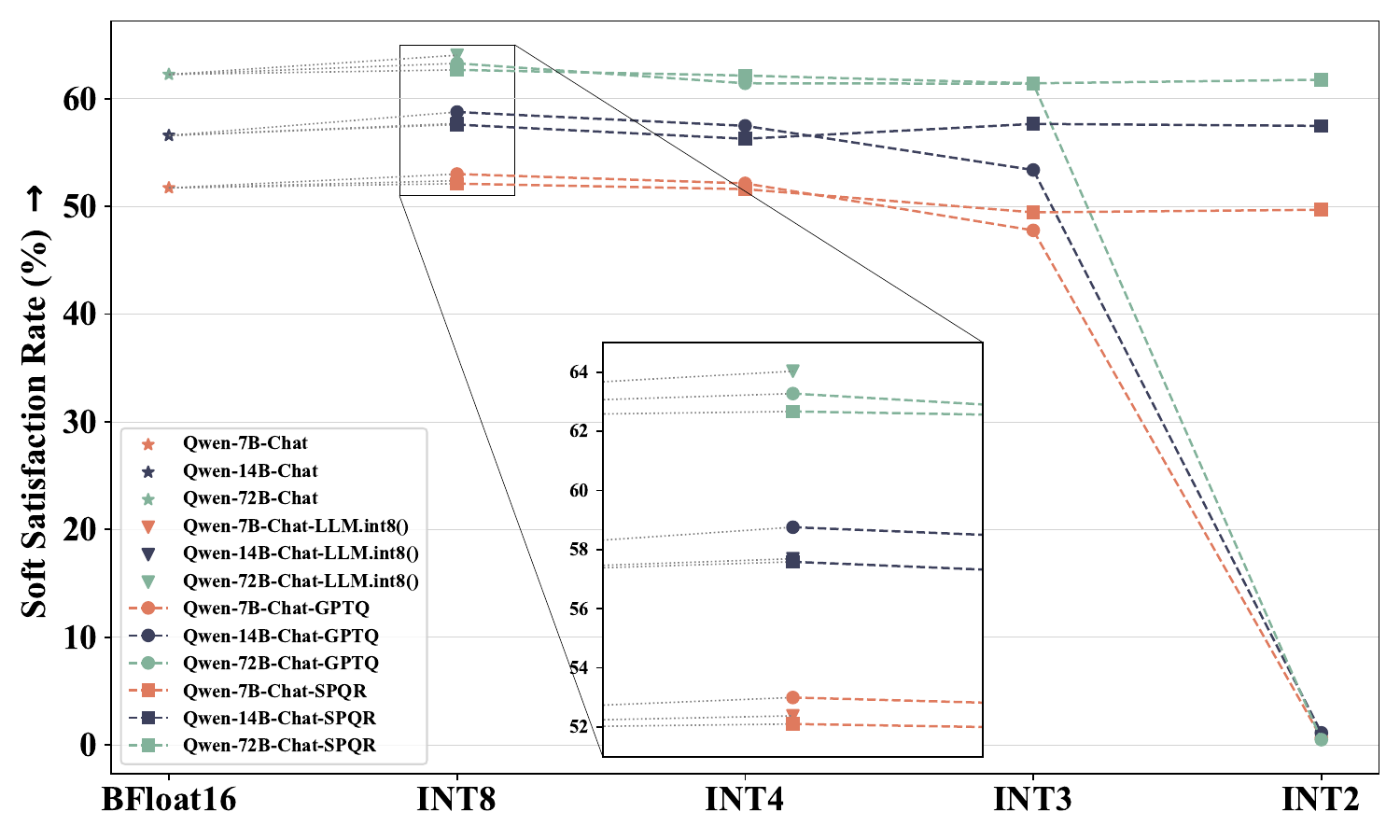}
    \caption{Soft Satisfaction Rate (SSR)}
\end{subfigure}
\begin{subfigure}{0.325\textwidth}
    \centering
    \includegraphics[width=\linewidth]{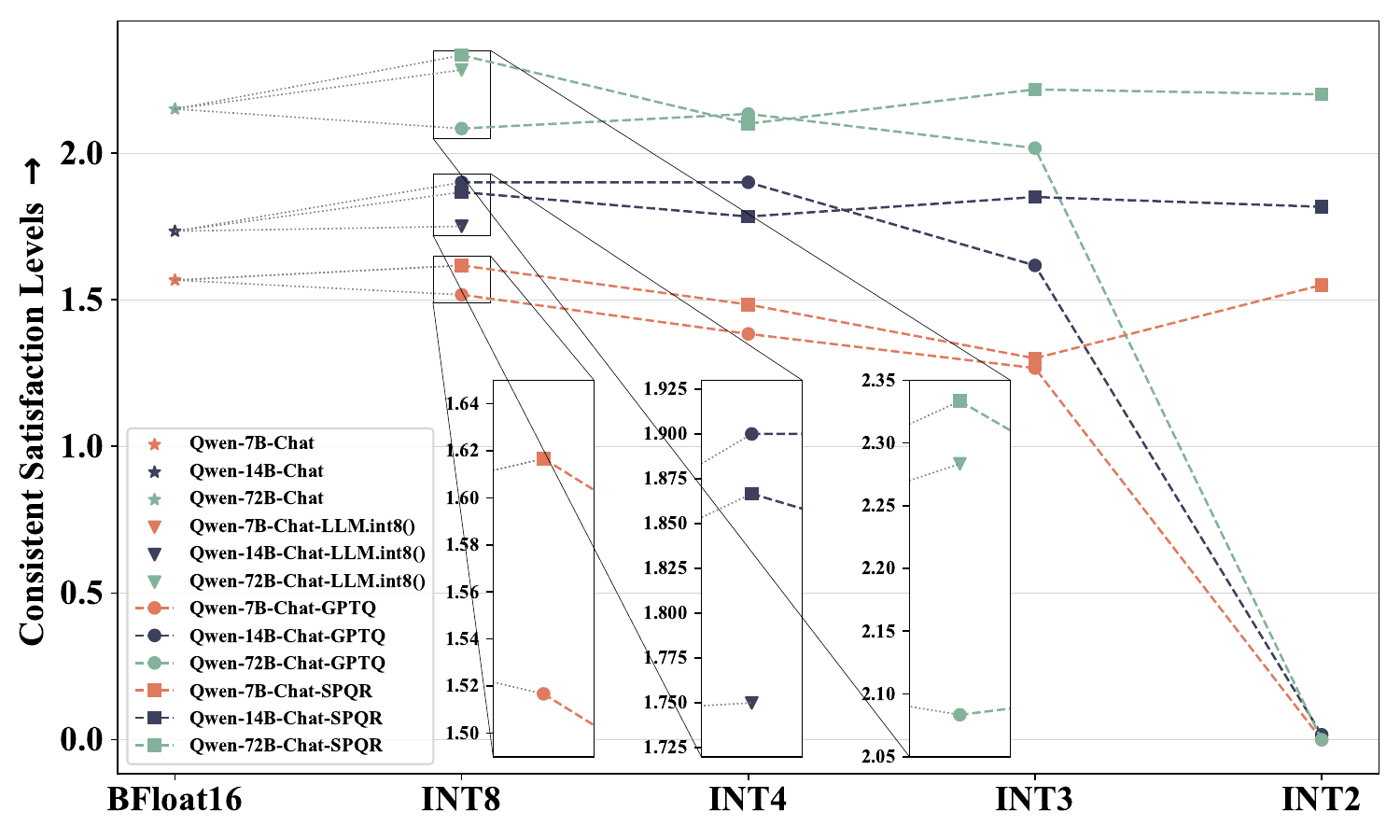}
    \caption{Consistent Satisfaction Levels (CSL)}
\end{subfigure}
\caption{Average hard satisfaction rates (a), soft satisfaction rates (b), and consistent satisfaction levels (c) across five difficulty levels for the Qwen-Chat series models and their quantized counterparts on the FollowBench benchmark \citep{DBLP:journals/corr/abs-2310-20410}.}
\label{figure:followbench}
\end{figure*}

\begin{figure*}[!ht]
\centering
\begin{subfigure}{0.325\textwidth}
    \centering
    \includegraphics[width=\linewidth]{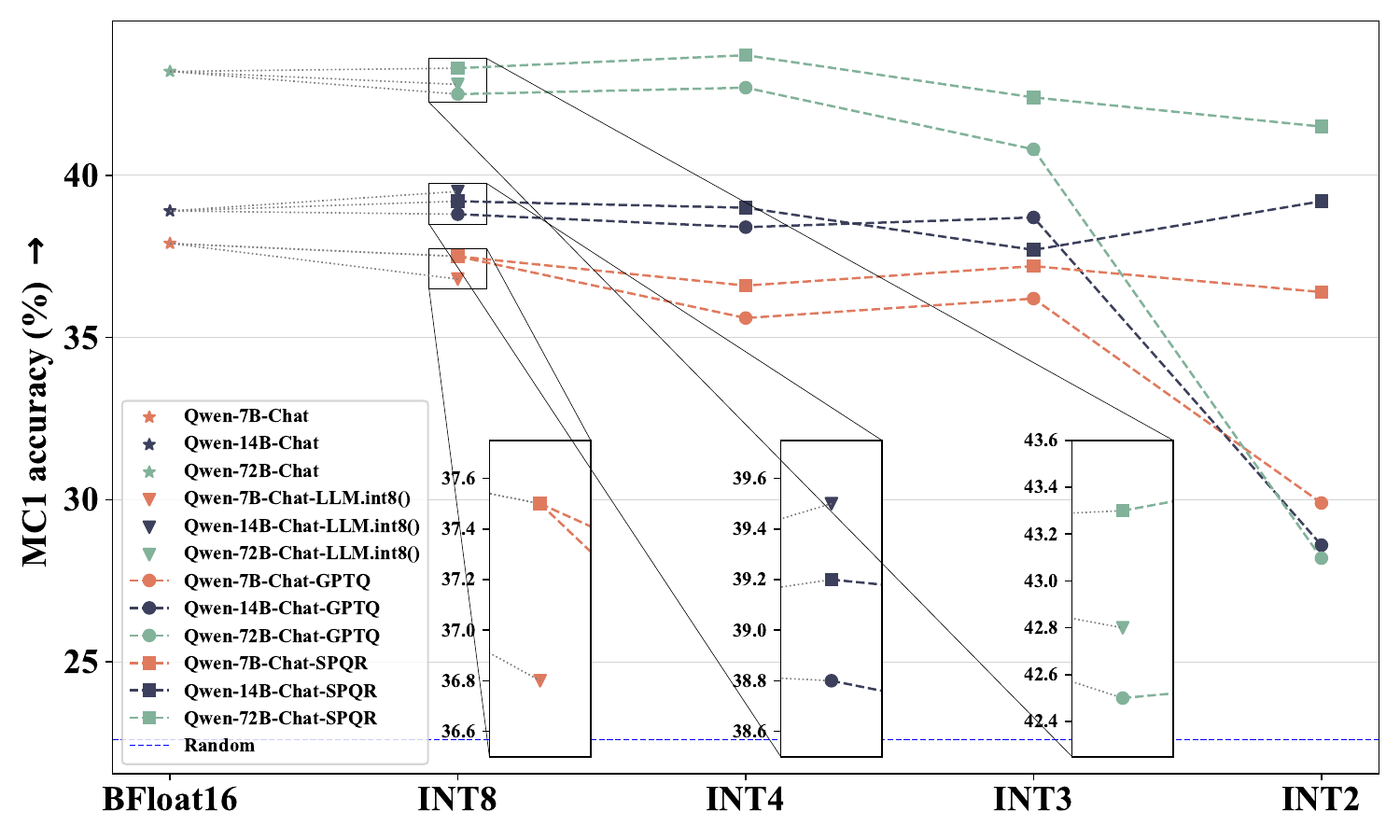}
    \caption{TruthfulQA}
    \label{figure:truthful_qa}
\end{subfigure}
\begin{subfigure}{0.325\textwidth}
    \centering
    \includegraphics[width=\linewidth]{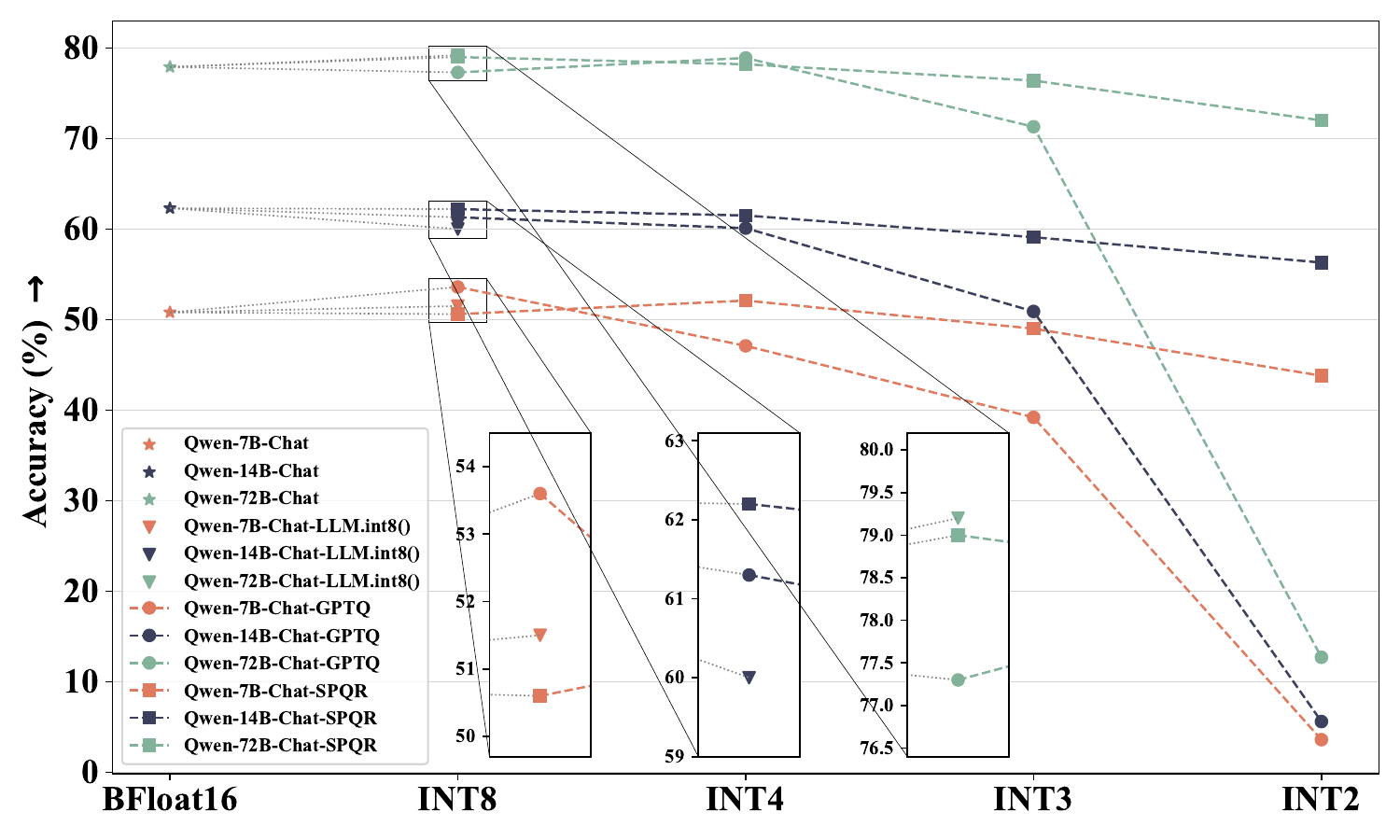}
    \caption{GSM8K}
\end{subfigure}
\begin{subfigure}{0.325\textwidth}
    \centering
    \includegraphics[width=\linewidth]{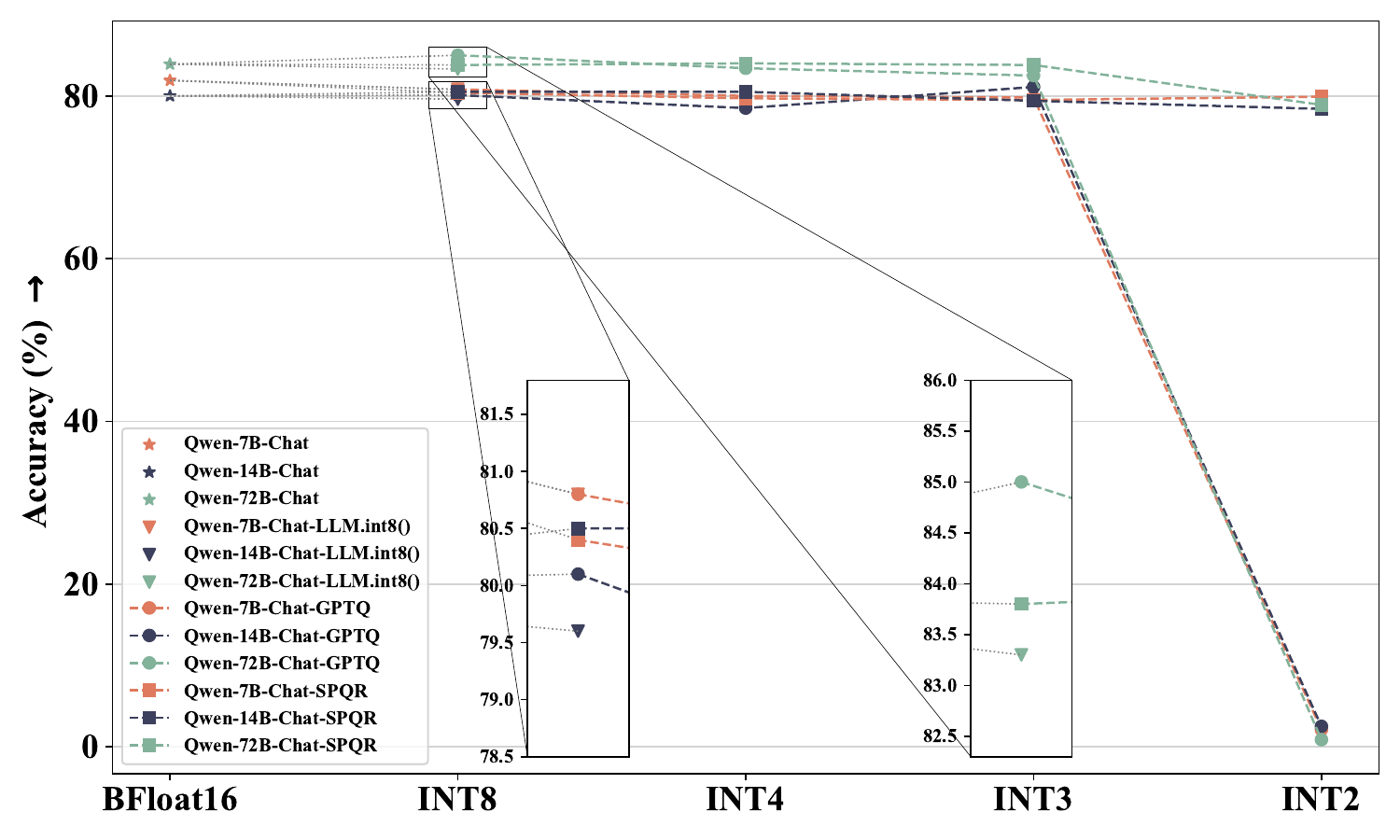}
    \caption{SNLI}
\end{subfigure}
\caption{Performance of Qwen-Chat series models and their quantized counterparts on the TruthfulQA benchmark \citep{DBLP:conf/acl/LinHE22} (a), as well as the test sets of GSM8K \citep{DBLP:journals/corr/abs-2110-14168} (b) and SNLI \citep{DBLP:conf/emnlp/BowmanAPM15} (c).}
\label{figure:truthful_qa_gsm8k_snli}
\end{figure*}

Overall, the experimental results indicate that LLMs with a greater number of parameters generally outperform those with fewer parameters across most benchmarks. Furthermore, we observe a downward trend in the performance of these LLMs across most benchmarks when they are quantized to fewer bits. Here are the detailed observations:

\paragraph{4-bit quantization offers a trade-off between the LLMs' capacity and the number of bits in the low-precision format. As the number of quantized bits decreases to 3 bits or lower, there is a noticeable performance discrepancy between the LLMs and their quantized counterparts.} Experimental results suggest that when the LLMs are quantized to 8 bits, the majority of LLMs, irrespective of their parameter scales, can maintain a performance level comparable to their non-quantized equivalents. Moreover, LLMs that are quantized to 4 bits can also uphold similar performance to their non-quantized versions across most benchmarks. However, if these LLMs are further quantized to 3 bits or lower, the capacity of these models begins to deteriorate. Notably, our investigation reveals that when the LLMs are quantized to 2 bits using GPTQ, they lose their ability to comprehend and follow user instructions, resulting in the generation of incoherent text.

\begin{figure*}[!ht]
\centering
\begin{subfigure}{0.475\textwidth}
    \centering
    \includegraphics[width=\linewidth]{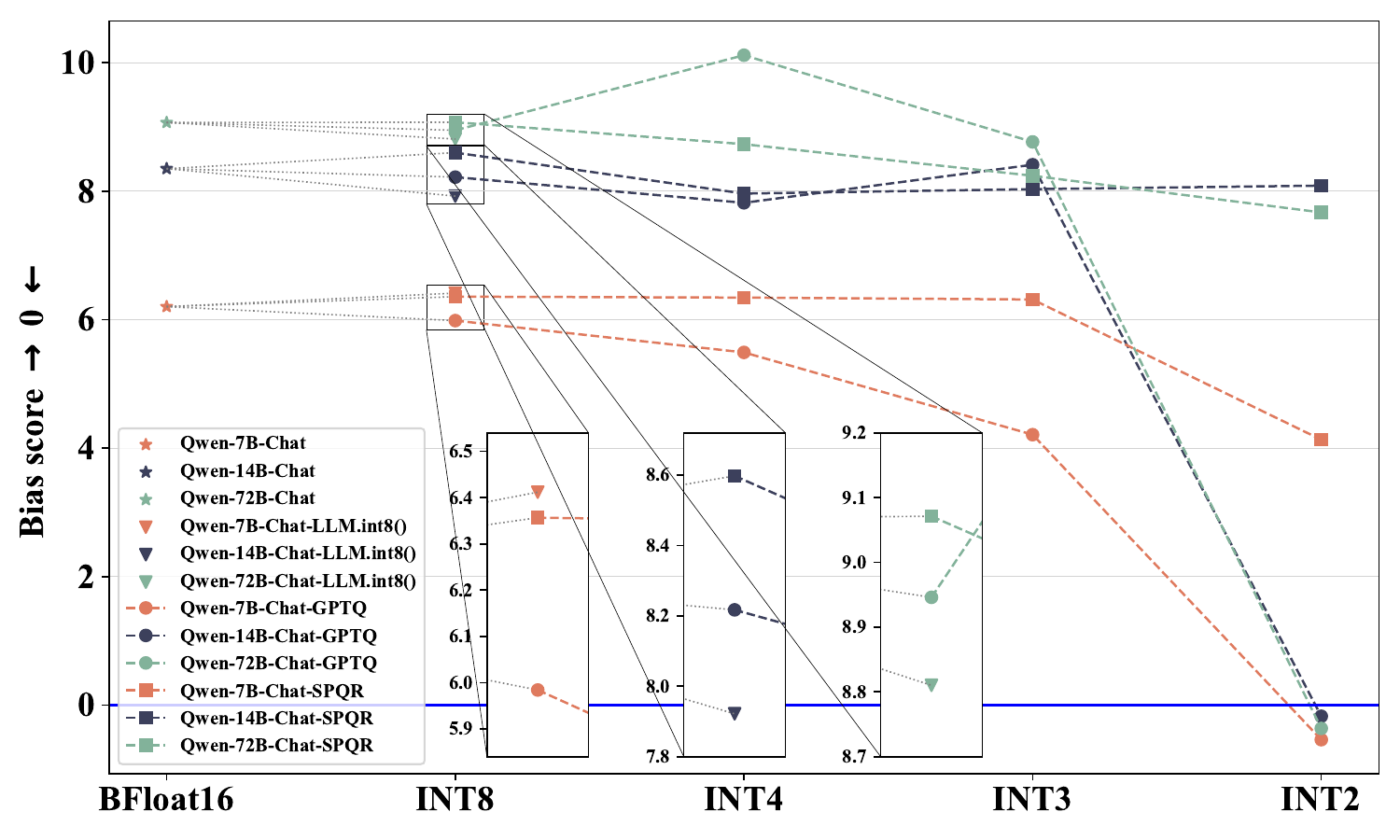}
    \caption{Bias scores in ambiguous context.}
\end{subfigure}
\begin{subfigure}{0.475\textwidth}
    \centering
    \includegraphics[width=\linewidth]{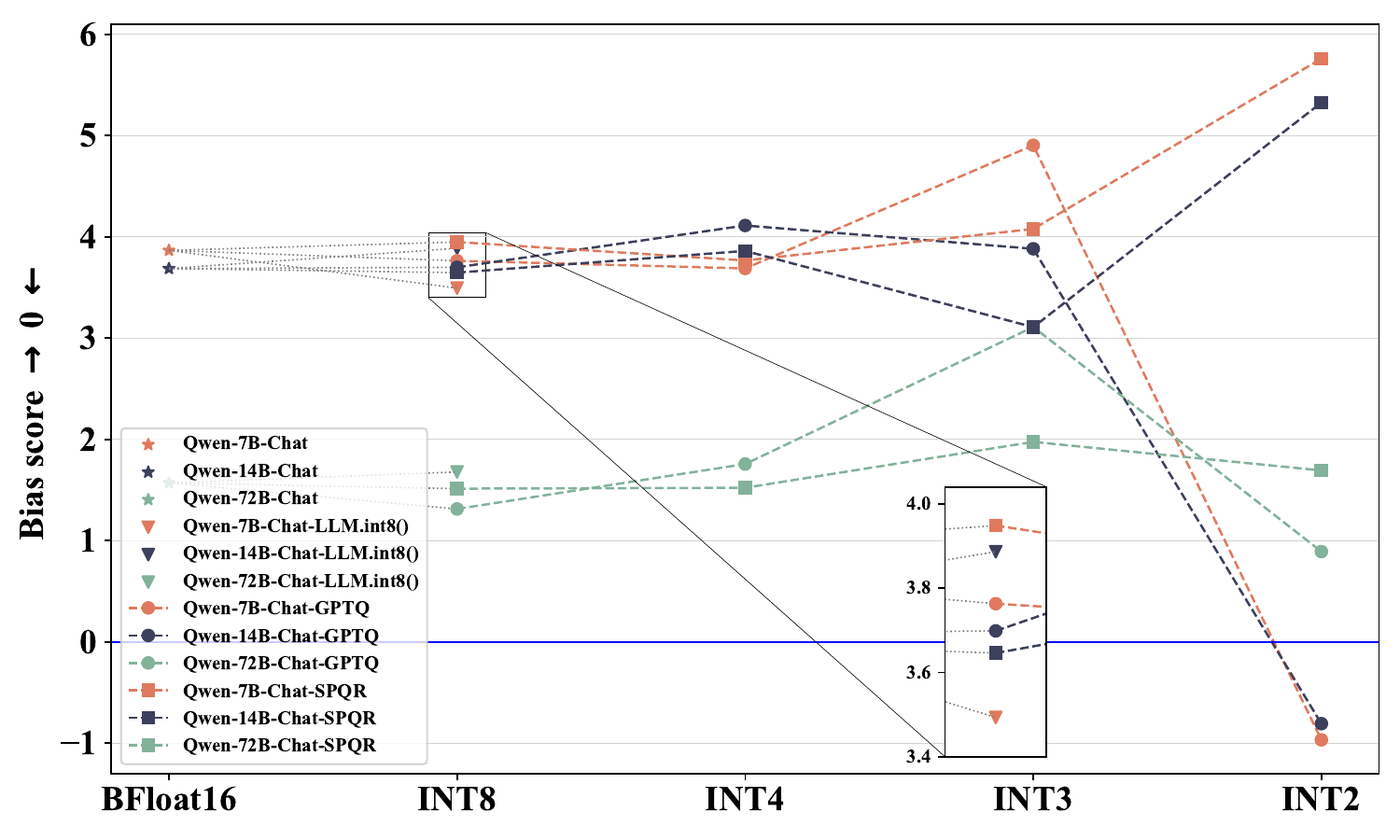}
    \caption{Bias scores in disambiguated context.}
\end{subfigure}
\caption{Bias scores of the Qwen-Chat series models and their quantized counterparts in ambiguous and disambiguated contexts on the BBQ benchmark \citep{DBLP:conf/acl/ParrishCNPPTHB22}.}
\label{figure:bbq}
\end{figure*}

\begin{figure}[!ht]
\centering
\includegraphics[width=\linewidth]{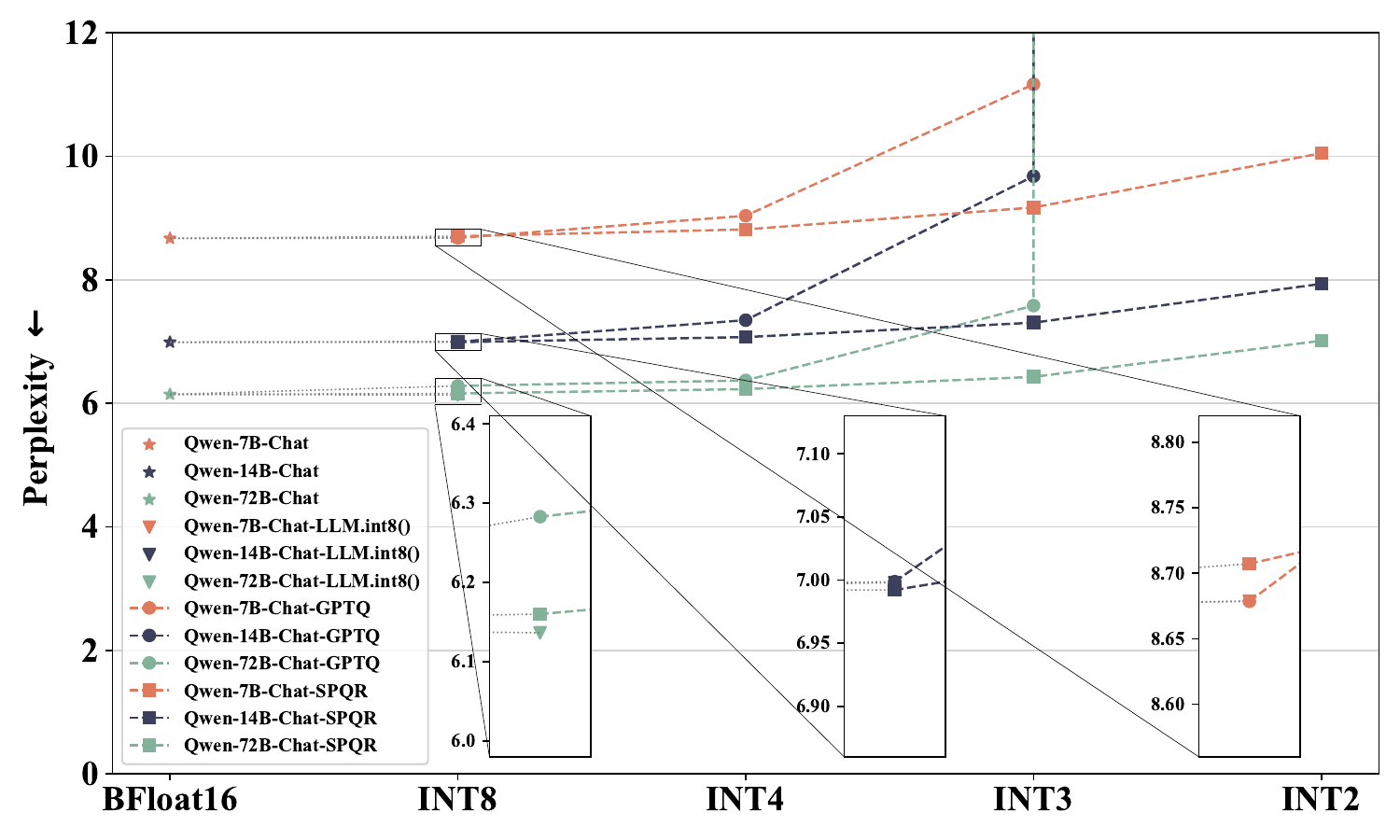}
\caption{Perplexity of Qwen-Chat Series models and their quantized counterparts on the WikiText2 dataset \citep{DBLP:conf/iclr/MerityX0S17}.}
\label{figure:ppl_wikitext}
\end{figure}

\paragraph{Perplexity is a reliable performance indicator for quantized LLMs on evaluation benchmarks.} Figure~\ref{figure:ppl_wikitext} illustrates the perplexity for both the original LLMs and their quantized versions on WikiText2 \citep{DBLP:conf/iclr/MerityX0S17}. For more experimental results of perplexity on the C4 \citep{DBLP:journals/jmlr/RaffelSRLNMZLL20} and PTB \citep{DBLP:conf/naacl/MarcusKMMBFKS94} datasets, please refer to Figure~\ref{figure:ppl_c4_ptb} in Appendix~\ref{appendix_sec:detailed_experimental_results}. It is evident that the perplexity of 8-bit quantized models closely matches that of their non-quantized counterparts. Moreover, as the LLMs are further quantized to 4 and 3 bits, there’s a slight increase in perplexity. However, perplexity sharply increases, exceeding 38,000, when the models are quantized to 2 bits using GPTQ. This sharp increase in perplexity aligns with our observation that models quantized to 2 bits with GPTQ struggle to generate coherent text. In summary, as LLMs are quantized to fewer bits, there is an upward trend in perplexity, which corresponds to a decline in their performance on evaluated benchmarks. Interestingly, despite a noticeable increase in perplexity, 4-bit quantized models still perform comparably to their non-quantized counterparts on these benchmarks. We speculate that this could be due to the nonlinear or discontinuous metrics used by these benchmarks, which may not reflect minor changes in perplexity. Furthermore, as demonstrated in Table~\ref{tab:pearson_corr}, there is a strong correlation between perplexity and the performance of quantized LLMs. The average absolute value of the Pearson correlation coefficient is notably high at 0.7895. This evidence reinforces our claim that perplexity serves as a reliable performance indicator for quantized LLMs on evaluation benchmarks.

\begin{table}[!t]
\centering
\resizebox{0.5\textwidth}{!}{
\begin{tabular}{ccc}
\toprule
\textbf{Benchmark} & \textbf{Metric} & \textbf{Pearson Correlation Coefficient} \\ \midrule
MMUL & Accuracy & -0.892 \\
\midrule
C-EVAL & Accuracy & -0.930 \\
\midrule
\multirow{2}{*}{FLORES-200} & BLEU (English to Chinese) & -0.884 \\
\arrayrulecolor{lightgray}\cmidrule{2-3}
 & BLEU (Chinese to English) & -0.904 \\
\arrayrulecolor{black}\midrule
\multirow{3}{*}{XSum} & ROUGE-1 & -0.768 \\
\arrayrulecolor{lightgray}\cmidrule{2-3}
 & ROUGE-2 & -0.493 \\
\arrayrulecolor{lightgray}\cmidrule{2-3}
 & ROUGE-L & -0.222 \\
\arrayrulecolor{black}\midrule
\multirow{3}{*}{CNN/DailyMail} & ROUGE-1 & -0.890 \\
\arrayrulecolor{lightgray}\cmidrule{2-3}
 & ROUGE-2 & -0.849 \\
\arrayrulecolor{lightgray}\cmidrule{2-3}
 & ROUGE-L & -0.885 \\
\arrayrulecolor{black}\midrule
GSM8K & Accuracy & -0.911 \\
\arrayrulecolor{black}\midrule
SNLI & Accuracy & -0.583 \\
\arrayrulecolor{black}\midrule
\multirow{3}{*}{FollowBench} & HSR (hard satisfaction rates) & -0.864 \\
\arrayrulecolor{lightgray}\cmidrule{2-3}
 & SSR (soft satisfaction rates) & -0.899 \\
\arrayrulecolor{lightgray}\cmidrule{2-3}
 & CSL (consistent satisfaction levels) & -0.877 \\
\arrayrulecolor{black}\midrule
TruthfulQA & MC1 Accuracy & -0.789 \\
\arrayrulecolor{black}\midrule
\multirow{2}{*}{BBQ} & Bias scores in ambiguous context & -0.765 \\
\arrayrulecolor{lightgray}\cmidrule{2-3}
 & Bias scores in disambiguated context & 0.806 \\
\arrayrulecolor{black}\bottomrule
\end{tabular}
}
\caption{The Pearson correlation coefficient between the average perplexity on the WikiText2, C4, and PTB datasets of both 4-bit and 3-bit quantized LLMs (quantized with GPTQ and SpQR) and their performance across various benchmarks.}
\label{tab:pearson_corr}
\end{table}

\begin{figure*}[!ht]
\centering
\begin{subfigure}{0.475\textwidth}
    \centering
    \includegraphics[width=\linewidth]{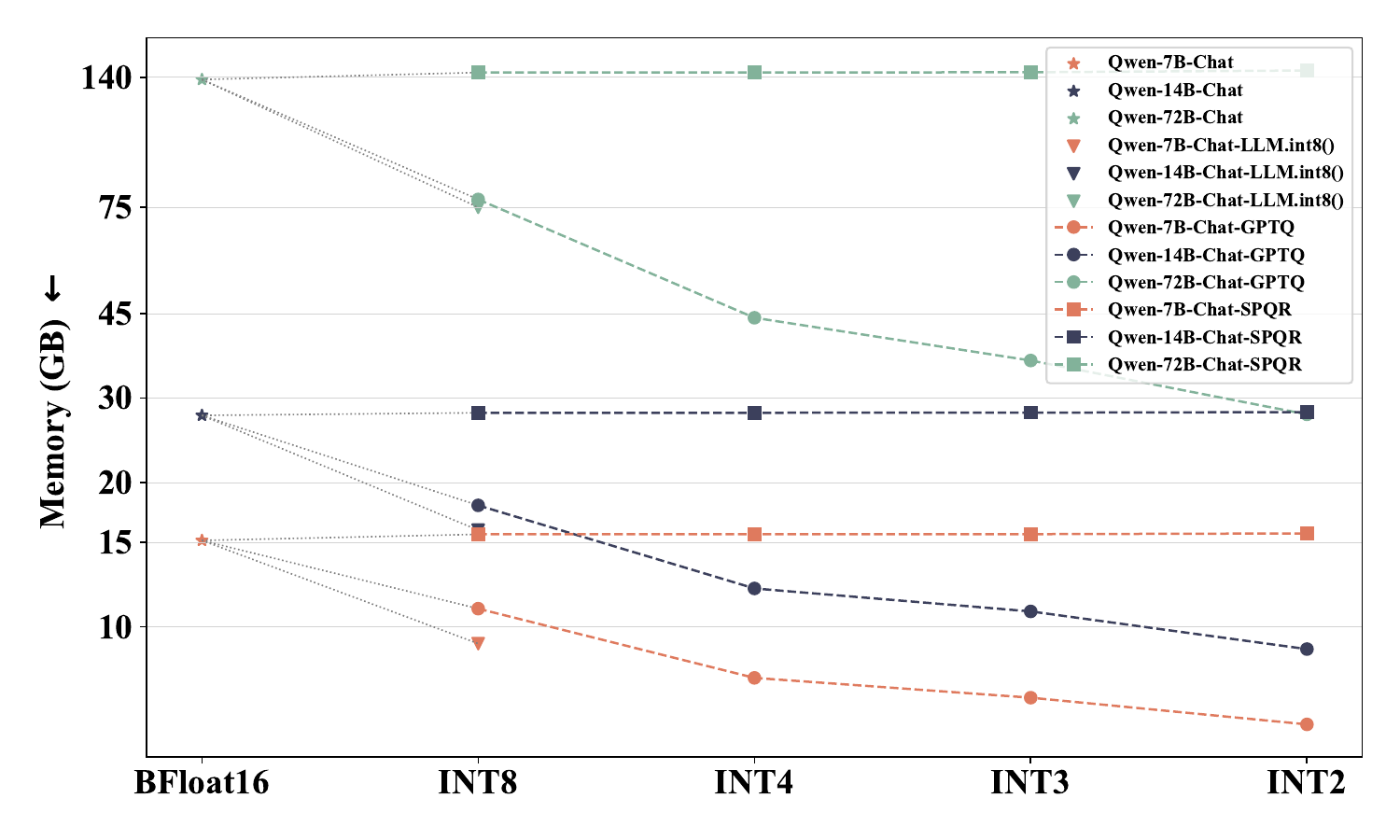}
    \caption{Memory}
    \label{figure:memory_and_speed:memory}
\end{subfigure}
\begin{subfigure}{0.475\textwidth}
    \centering
    \includegraphics[width=\linewidth]{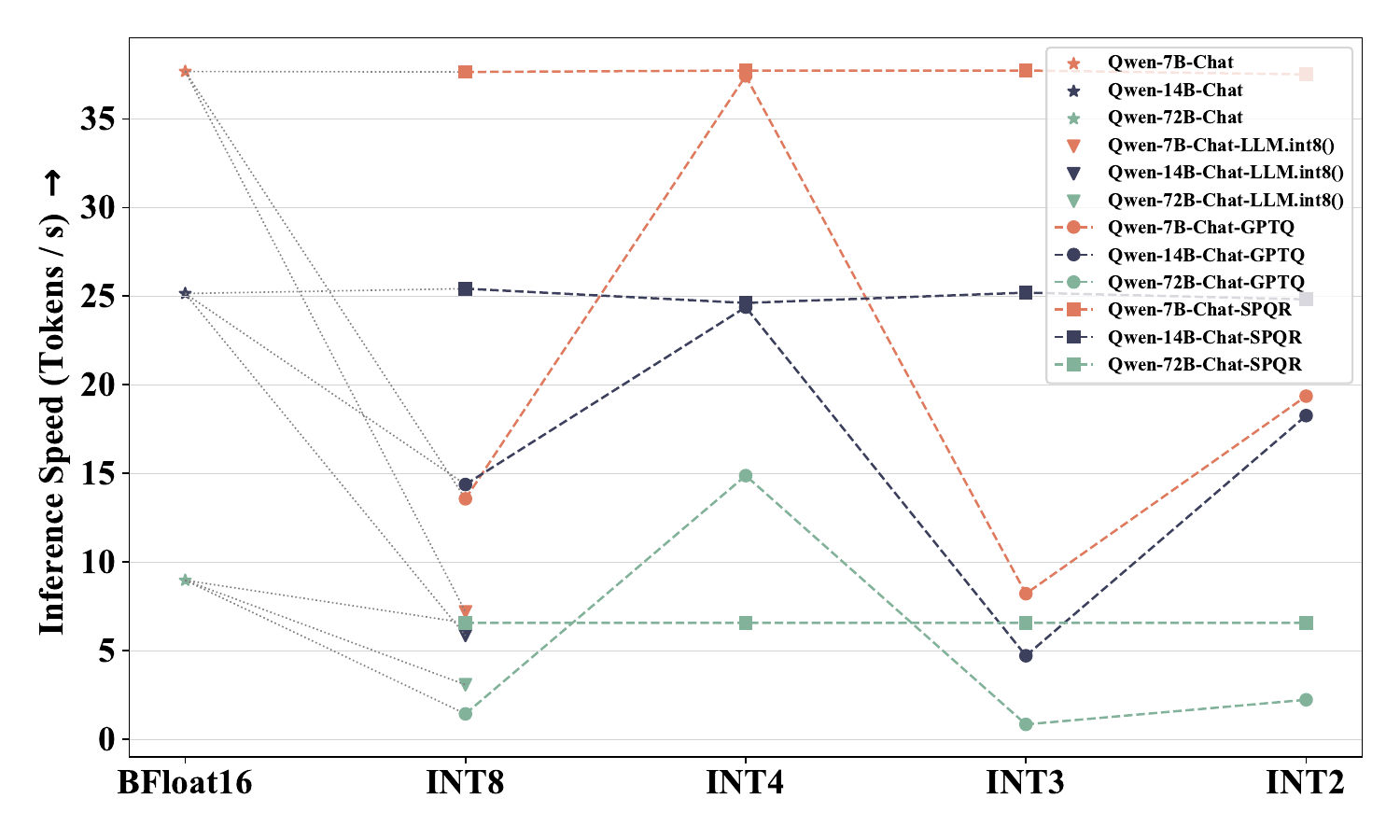}
    \caption{Speed}
    \label{figure:memory_and_speed:speed}
\end{subfigure}
\caption{\textbf{Left:} memory consumption comparison between Qwen-Chat series models and their quantized counterparts. The y-axis is presented on a logarithmic scale to clearly demonstrate the variation in memory consumption for LLMs with smaller parameter scales (7B, 14B) as the number of quantized bits decreases. \textbf{Right:} comparison of inference speed between Qwen-Chat series models and their quantized counterparts. These experiments are conducted with an input of 256 tokens and a generation of 512 tokens on A100 80GB SXM GPUs.}
\label{figure:memory_and_speed}
\end{figure*}

\paragraph{Identifying and isolating outlier weights is crucial for SpQR to effectively quantize LLMs to an extreme level of 2 bits.} Experimental results indicate a sharp decline in the performance of LLMs quantized to 2 bits by GPTQ, to the extent that they fail to produce coherent text. In contrast, LLMs quantized to 2 bits by SpQR exhibit a relatively moderate performance across all evaluated benchmarks. SpQR introduces two innovative strategies to enhance the performance of quantized LLMs, distinguishing it from GPTQ: (1) the adoption of an extremely small group size coupled with bilevel quantization, and (2) the isolation of unstructured outlier weights, maintaining these weights at a higher precision (16-bit) during computations. To study the impact of these strategies, we conducted two controlled experiments: (1) increasing the group size of SpQR from 16 to 128, matching the group size utilized by GPTQ, while still isolating the outlier weights. (2) keeping the small group size but not isolating outlier weights. Experimental results are shown in Table~\ref{tab:outlier-or-group_size}. We observe a significant increase in perplexity across three benchmarks when the outlier weights are not isolated, even with a small group size. Conversely, increasing the group size resulted in only a marginal increase in perplexity. Furthermore, we analyzed the proportion of outlier weights stored in high precision for the quantized LLMs, with the results presented in Table~\ref{tab:outlier-proportion}. These findings indicate an inverse relationship between the number of quantized bits and the percentage of outlier weights, with a consistent percentage of outlier weights across different model scales at the same quantization level. Consequently, it is concluded that the isolation of outlier weights and their preservation in high precision is indispensable for SpQR to effectively quantize LLMs to an extreme level of 2 bits.

\begin{table}[!t]
\centering
\tiny
\begin{tabular}{ccccc}
\toprule
\multicolumn{1}{c}{\textbf{Model}} & \textbf{Quantization Config} & \textbf{WikiText} & \textbf{C4} & \textbf{PTB} \\ \midrule
\multirow{3}{*}{Qwen-7B-Chat} & w2g16 w/ outlier & 10.05 & 14.19 & 17.07 \\
 & w2g16 w/o outlier & 17.96 & 21.86 & 27.98 \\
 & w2g128 w/ outlier & 10.58 & 14.48 & 17.40 \\ \midrule
\multirow{3}{*}{Qwen-14B-Chat} & w2g16 w/ outlier & 7.94 & 11.74 & 13.31 \\
 & w2g16 w/o outlier & 140.22 & 115.07 & 170.48 \\
 & w2g128 w/ outlier & 8.16 & 12.14 & 13.78 \\ \midrule
\multirow{3}{*}{Qwen-72B-Chat} & w2g16 w/ outlier & 7.01 & 9.78 & 11.92 \\
 & w2g16 w/o outlier & 10.49 & 14.07 & 16.11 \\
 & w2g128 w/ outlier & 7.44 & 10.34 & 12.27 \\ \bottomrule
\end{tabular}%
 \caption{Perplexity on WikiText2 \citep{DBLP:conf/iclr/MerityX0S17}, C4 \citep{DBLP:journals/jmlr/RaffelSRLNMZLL20}, and PTB \citep{DBLP:conf/naacl/MarcusKMMBFKS94} under different quantization configurations. ``w2g16'' denotes that weights are quantized to 2-bit with a group size of 16. ``w/ outlier'' indicates identifying outlier values which are not quantized while ``w/o outlier'' means not identifying outliers and the whole weight matrix is quantized. All experiments used bilevel 3-bit quantization, which quantizes the model's weights first and then quantizes group-wise statistics (scales and zeros).}
\label{tab:outlier-or-group_size}
\end{table}

\paragraph{In practical scenarios, the application of low-bit quantization necessitates substantial engineering effort and hardware support.} As illustrated in Figure~\ref{figure:memory_and_speed:memory}, both the GPTQ and LLM.int8() can effectively reduce memory consumption during LLMs inference, with the memory requirement diminishing as the number of quantized bits decreases. Conversely, despite the impressive performance of SpQR, it does not contribute to reducing memory consumption during LLMs inference. This is attributable to the implementation of SpQR employed in our study, which utilizes a high-precision format to represent quantized weights. It merely restricts the range of quantized weights to match that of the low-precision format, thereby mimicking the effect of representing quantized weights with low precision. Consequently, computations are executed under a high-precision format, resulting in no reduction in memory consumption. Furthermore, the efficient implementation of parallel computation in low-precision format is not yet supported by most computing libraries, such as PyTorch. This implies that the implementation of operators associated with low-precision format must be done manually, demanding a thorough understanding of computing hardware (e.g., GPU, TPU, etc.) and the dedication of considerable engineering effort to achieve efficient execution.

\begin{table}[!t]
\centering
\tiny
\begin{tabular}{ccc}
\toprule
\textbf{Model} & \textbf{Quantized Bit} & \textbf{Outlier Proportion} \\
\midrule
\multicolumn{1}{c}{\multirow{4}{*}{Qwen-7B-Chat}} & 8 & 0.003\% \\
 & 4 & 0.033\% \\
 & 3 & 1.676\% \\
 & 2 & 11.336\% \\
\midrule
\multicolumn{1}{c}{\multirow{4}{*}{Qwen-14B-Chat}} & 8 & 0.004\% \\
 & 4 & 0.036\% \\
 & 3 & 1.648\% \\
 & 2 & 11.148\% \\
\midrule
\multicolumn{1}{c}{\multirow{4}{*}{Qwen-72B-Chat}} & 8 & 0.003\% \\
 & 4 & 0.044\% \\
 & 3 & 1.682\% \\
 & 2 & 11.838\% \\
\bottomrule
\end{tabular}%
\caption{The proportion of outliers keeping high precision in LLMs quantized by SpQR.}
\label{tab:outlier-proportion}
\end{table}

Beyond memory consumption, Figure~\ref{figure:memory_and_speed:speed} reveals that while GPTQ and LLM.int8(), whose underlying implementation used in our study perform computation in low-precision format, lead to notable memory savings compared to their non-quantized counterparts, the inference speed of LLMs quantized by GPTQ and LLM.int8() is slower compared to their non-quantized counterparts, except in the case of 4-bit quantization. This slowdown is primarily due to the fact that only the weights of the LLMs use the low-precision format representation, while activations still employ the high-precision format representation. The acceleration of computation between this mixed precision format is not supported by the hardware used in our experiments. However, in the case of 4-bit quantization, only the LLM with 72B parameters exhibits a significant speed-up compared to its non-quantized counterpart, while others show similar inference speeds to their non-quantized counterparts. We hypothesize that this may be due to characteristics of the hardware, such as memory bandwidth \citep{DBLP:journals/corr/abs-1911-02150}. In summary, both the efficient implementation of parallel computation in low-precision format, which requires considerable engineering effort, and the acceleration of computation supported by associated hardware are essential for quantization techniques to effectively reduce memory usage and accelerate decoding during inference.

\paragraph{At similar levels of memory consumption, LLMs quantized to lower bit precision with a larger parameter scale can be preferred over LLMs with a smaller parameter scale, considering their performance capabilities.} As illustrated in Figure~~\ref{figure:memory_and_speed:memory}, the memory consumption during inference for Qwen-14B-Chat with 8-bit or 4-bit quantization by GPTQ is similar to that of Qwen-7B-Chat. However, the former outperforms the latter in most of the benchmarks evaluated. Additionally, Qwen-14B-Chat with 4-bit or 3-bit quantization can be competitive with Qwen-7B-Chat with 8-bit quantization. Nonetheless, while quantized LLMs offer advantages in terms of memory efficiency, quantization can also result in reduced inference speed. Therefore, these quantization approaches are most suitable for scenarios where memory is limited and inference speed is a secondary consideration.

\section{Conclusion}

We have presented a comprehensive evaluation of quantization strategies for LLMs, demonstrating the trade-offs between model efficiency and performance degradation across various benchmarks. By employing a structured evaluation framework that assesses models in terms of knowledge \& capacity, alignment, and efficiency, we aim to offer valuable insights into the scalability and practical application of quantized LLMs. Experimental findings indicate that while 4-bit quantization maintains performance close to non-quantized counterparts, a notable performance discrepancy emerges as quantization decreases to 3 bits or lower. Moreover, the results suggest that perplexity can be a reliable performance indicator for quantized LLMs on various evaluation benchmarks. SpQR effectively quantizes LLMs to an extreme level of 2 bits by isolating outlier weights and maintaining high precision during computation. When memory constraints exist and inference speed is a secondary concern, LLMs quantized to lower bit precision with a larger parameter scale can be preferred over smaller models. Additionally, we highlight the need for engineering effort and hardware support to efficiently deploy quantized LLMs in real-world scenarios.

\section*{Limitations}
We have utilized ten distinct benchmarks, encompassing knowledge \& capacity and alignment, for our evaluation. However, LLMs are pre-trained on vast amounts of data. This could potentially lead to the contamination of some test examples in the benchmarks we used with pre-training data, possibly resulting in an overestimation of the LLMs' performance \citep{DBLP:journals/corr/abs-2311-04850,DBLP:journals/corr/abs-2309-10677,DBLP:journals/corr/abs-2310-17623}. Consequently, it remains unclear whether the evaluated experimental results on these benchmarks could be generalized to other benchmarks. Identifying and eliminating these contaminated examples poses a significant challenge, and we leave it as our further work. Furthermore, due to limited computational resources, our experiments were confined to the Qwen-Chat series of models \citep{DBLP:journals/corr/abs-2309-16609}, which have diverse parameter scales and are trained on an extensive multilingual corpus that is dominated by both English and Chinese. The experimental results and findings from the Qwen series of models may not necessarily generalize to other LLMs, owing to various factors such as differences in training data, hyperparameters, and architectures.

\section*{Ethical Considerations}
In this study, we employ the BBQ \citep{DBLP:conf/acl/ParrishCNPPTHB22} and TruthfulQA \citep{DBLP:conf/acl/LinHE22} benchmarks to investigate the potential impact of quantization on the alignment of LLMs with human values. Our focus is on assessing the social bias and truthfulness of both quantized and non-quantized versions of these models.

The experimental results, as illustrated in Figure~\ref{figure:bbq}, reveal no consistent trend of increase or decrease in social bias when LLMs are quantized to fewer bits. However, it is noteworthy that quantization can either exacerbate or alleviate the social bias of the quantized LLMs in comparison to their non-quantized counterparts. Furthermore, Figure~\ref{figure:truthful_qa} demonstrates that the truthfulness of LLMs can also be influenced by quantization. Specifically, when LLMs are quantized to 2 bits using GPTQ, there is a significant decrease in the truthfulness of the quantized LLMs.

In conclusion, our findings suggest that in addition to commonly evaluated dimensions such as knowledge \& capacity and efficiency, the alignment of LLMs with human values, which is a dimension often overlooked in previous studies of LLM quantization, deserves greater attention.

\section*{Acknowledgements}
The present research was supported by the National Key Research and Development Program of China (Grant No. 2023YFE0116400). We would like to thank the anonymous reviewers for their insightful comments.

\bibliography{custom}

\begin{thebibliography}{81}
\expandafter\ifx\csname natexlab\endcsname\relax\def\natexlab#1{#1}\fi

\bibitem[{Askell et~al.(2021)Askell, Bai, Chen, Drain, Ganguli, Henighan, Jones, Joseph, Mann, DasSarma, Elhage, Hatfield{-}Dodds, Hernandez, Kernion, Ndousse, Olsson, Amodei, Brown, Clark, McCandlish, Olah, and Kaplan}]{DBLP:journals/corr/abs-2112-00861}
Amanda Askell, Yuntao Bai, Anna Chen, Dawn Drain, Deep Ganguli, Tom Henighan, Andy Jones, Nicholas Joseph, Benjamin Mann, Nova DasSarma, Nelson Elhage, Zac Hatfield{-}Dodds, Danny Hernandez, Jackson Kernion, Kamal Ndousse, Catherine Olsson, Dario Amodei, Tom~B. Brown, Jack Clark, Sam McCandlish, Chris Olah, and Jared Kaplan. 2021.
\newblock \href {http://arxiv.org/abs/2112.00861} {A general language assistant as a laboratory for alignment}.
\newblock \emph{CoRR}, abs/2112.00861.

\bibitem[{Bai et~al.(2023)Bai, Bai, Chu, Cui, Dang, Deng, Fan, Ge, Han, Huang, Hui, Ji, Li, Lin, Lin, Liu, Liu, Lu, Lu, Ma, Men, Ren, Ren, Tan, Tan, Tu, Wang, Wang, Wang, Wu, Xu, Xu, Yang, Yang, Yang, Yang, Yao, Yu, Yuan, Yuan, Zhang, Zhang, Zhang, Zhang, Zhou, Zhou, Zhou, and Zhu}]{DBLP:journals/corr/abs-2309-16609}
Jinze Bai, Shuai Bai, Yunfei Chu, Zeyu Cui, Kai Dang, Xiaodong Deng, Yang Fan, Wenbin Ge, Yu~Han, Fei Huang, Binyuan Hui, Luo Ji, Mei Li, Junyang Lin, Runji Lin, Dayiheng Liu, Gao Liu, Chengqiang Lu, Keming Lu, Jianxin Ma, Rui Men, Xingzhang Ren, Xuancheng Ren, Chuanqi Tan, Sinan Tan, Jianhong Tu, Peng Wang, Shijie Wang, Wei Wang, Shengguang Wu, Benfeng Xu, Jin Xu, An~Yang, Hao Yang, Jian Yang, Shusheng Yang, Yang Yao, Bowen Yu, Hongyi Yuan, Zheng Yuan, Jianwei Zhang, Xingxuan Zhang, Yichang Zhang, Zhenru Zhang, Chang Zhou, Jingren Zhou, Xiaohuan Zhou, and Tianhang Zhu. 2023.
\newblock \href {https://doi.org/10.48550/ARXIV.2309.16609} {Qwen technical report}.
\newblock \emph{CoRR}, abs/2309.16609.

\bibitem[{Bai et~al.(2022)Bai, Jones, Ndousse, Askell, Chen, DasSarma, Drain, Fort, Ganguli, Henighan, Joseph, Kadavath, Kernion, Conerly, Showk, Elhage, Hatfield{-}Dodds, Hernandez, Hume, Johnston, Kravec, Lovitt, Nanda, Olsson, Amodei, Brown, Clark, McCandlish, Olah, Mann, and Kaplan}]{DBLP:journals/corr/abs-2204-05862}
Yuntao Bai, Andy Jones, Kamal Ndousse, Amanda Askell, Anna Chen, Nova DasSarma, Dawn Drain, Stanislav Fort, Deep Ganguli, Tom Henighan, Nicholas Joseph, Saurav Kadavath, Jackson Kernion, Tom Conerly, Sheer~El Showk, Nelson Elhage, Zac Hatfield{-}Dodds, Danny Hernandez, Tristan Hume, Scott Johnston, Shauna Kravec, Liane Lovitt, Neel Nanda, Catherine Olsson, Dario Amodei, Tom~B. Brown, Jack Clark, Sam McCandlish, Chris Olah, Benjamin Mann, and Jared Kaplan. 2022.
\newblock \href {https://doi.org/10.48550/ARXIV.2204.05862} {Training a helpful and harmless assistant with reinforcement learning from human feedback}.
\newblock \emph{CoRR}, abs/2204.05862.

\bibitem[{Bang et~al.(2023)Bang, Cahyawijaya, Lee, Dai, Su, Wilie, Lovenia, Ji, Yu, Chung, Do, Xu, and Fung}]{DBLP:journals/corr/abs-2302-04023}
Yejin Bang, Samuel Cahyawijaya, Nayeon Lee, Wenliang Dai, Dan Su, Bryan Wilie, Holy Lovenia, Ziwei Ji, Tiezheng Yu, Willy Chung, Quyet~V. Do, Yan Xu, and Pascale Fung. 2023.
\newblock \href {https://doi.org/10.48550/ARXIV.2302.04023} {A multitask, multilingual, multimodal evaluation of chatgpt on reasoning, hallucination, and interactivity}.
\newblock \emph{CoRR}, abs/2302.04023.

\bibitem[{Bowman et~al.(2015)Bowman, Angeli, Potts, and Manning}]{DBLP:conf/emnlp/BowmanAPM15}
Samuel~R. Bowman, Gabor Angeli, Christopher Potts, and Christopher~D. Manning. 2015.
\newblock \href {https://doi.org/10.18653/V1/D15-1075} {A large annotated corpus for learning natural language inference}.
\newblock In \emph{Proceedings of the 2015 Conference on Empirical Methods in Natural Language Processing, {EMNLP} 2015, Lisbon, Portugal, September 17-21, 2015}, pages 632--642. The Association for Computational Linguistics.

\bibitem[{Brown et~al.(2020)Brown, Mann, Ryder, Subbiah, Kaplan, Dhariwal, Neelakantan, Shyam, Sastry, Askell, Agarwal, Herbert{-}Voss, Krueger, Henighan, Child, Ramesh, Ziegler, Wu, Winter, Hesse, Chen, Sigler, Litwin, Gray, Chess, Clark, Berner, McCandlish, Radford, Sutskever, and Amodei}]{DBLP:conf/nips/BrownMRSKDNSSAA20}
Tom~B. Brown, Benjamin Mann, Nick Ryder, Melanie Subbiah, Jared Kaplan, Prafulla Dhariwal, Arvind Neelakantan, Pranav Shyam, Girish Sastry, Amanda Askell, Sandhini Agarwal, Ariel Herbert{-}Voss, Gretchen Krueger, Tom Henighan, Rewon Child, Aditya Ramesh, Daniel~M. Ziegler, Jeffrey Wu, Clemens Winter, Christopher Hesse, Mark Chen, Eric Sigler, Mateusz Litwin, Scott Gray, Benjamin Chess, Jack Clark, Christopher Berner, Sam McCandlish, Alec Radford, Ilya Sutskever, and Dario Amodei. 2020.
\newblock \href {https://proceedings.neurips.cc/paper/2020/hash/1457c0d6bfcb4967418bfb8ac142f64a-Abstract.html} {Language models are few-shot learners}.
\newblock In \emph{Advances in Neural Information Processing Systems 33: Annual Conference on Neural Information Processing Systems 2020, NeurIPS 2020, December 6-12, 2020, virtual}.

\bibitem[{Chang et~al.(2024)Chang, Wang, Wang, Wu, Yang, Zhu, Chen, Yi, Wang, Wang, Ye, Zhang, Chang, Yu, Yang, and Xie}]{10.1145/3641289}
Yupeng Chang, Xu~Wang, Jindong Wang, Yuan Wu, Linyi Yang, Kaijie Zhu, Hao Chen, Xiaoyuan Yi, Cunxiang Wang, Yidong Wang, Wei Ye, Yue Zhang, Yi~Chang, Philip~S. Yu, Qiang Yang, and Xing Xie. 2024.
\newblock \href {https://doi.org/10.1145/3641289} {A survey on evaluation of large language models}.
\newblock \emph{ACM Trans. Intell. Syst. Technol.}
\newblock Just Accepted.

\bibitem[{Cobbe et~al.(2021)Cobbe, Kosaraju, Bavarian, Chen, Jun, Kaiser, Plappert, Tworek, Hilton, Nakano, Hesse, and Schulman}]{DBLP:journals/corr/abs-2110-14168}
Karl Cobbe, Vineet Kosaraju, Mohammad Bavarian, Mark Chen, Heewoo Jun, Lukasz Kaiser, Matthias Plappert, Jerry Tworek, Jacob Hilton, Reiichiro Nakano, Christopher Hesse, and John Schulman. 2021.
\newblock \href {http://arxiv.org/abs/2110.14168} {Training verifiers to solve math word problems}.
\newblock \emph{CoRR}, abs/2110.14168.

\bibitem[{Costa{-}juss{\`{a}} et~al.(2022)Costa{-}juss{\`{a}}, Cross, {\c{C}}elebi, Elbayad, Heafield, Heffernan, Kalbassi, Lam, Licht, Maillard, Sun, Wang, Wenzek, Youngblood, Akula, Barrault, Gonzalez, Hansanti, Hoffman, Jarrett, Sadagopan, Rowe, Spruit, Tran, Andrews, Ayan, Bhosale, Edunov, Fan, Gao, Goswami, Guzm{\'{a}}n, Koehn, Mourachko, Ropers, Saleem, Schwenk, and Wang}]{DBLP:journals/corr/abs-2207-04672}
Marta~R. Costa{-}juss{\`{a}}, James Cross, Onur {\c{C}}elebi, Maha Elbayad, Kenneth Heafield, Kevin Heffernan, Elahe Kalbassi, Janice Lam, Daniel Licht, Jean Maillard, Anna Sun, Skyler Wang, Guillaume Wenzek, Al~Youngblood, Bapi Akula, Lo{\"{\i}}c Barrault, Gabriel~Mejia Gonzalez, Prangthip Hansanti, John Hoffman, Semarley Jarrett, Kaushik~Ram Sadagopan, Dirk Rowe, Shannon Spruit, Chau Tran, Pierre Andrews, Necip~Fazil Ayan, Shruti Bhosale, Sergey Edunov, Angela Fan, Cynthia Gao, Vedanuj Goswami, Francisco Guzm{\'{a}}n, Philipp Koehn, Alexandre Mourachko, Christophe Ropers, Safiyyah Saleem, Holger Schwenk, and Jeff Wang. 2022.
\newblock \href {https://doi.org/10.48550/ARXIV.2207.04672} {No language left behind: Scaling human-centered machine translation}.
\newblock \emph{CoRR}, abs/2207.04672.

\bibitem[{Dettmers et~al.(2022)Dettmers, Lewis, Belkada, and Zettlemoyer}]{DBLP:conf/nips/DettmersLBZ22}
Tim Dettmers, Mike Lewis, Younes Belkada, and Luke Zettlemoyer. 2022.
\newblock \href {http://papers.nips.cc/paper\_files/paper/2022/hash/c3ba4962c05c49636d4c6206a97e9c8a-Abstract-Conference.html} {Gpt3.int8(): 8-bit matrix multiplication for transformers at scale}.
\newblock In \emph{Advances in Neural Information Processing Systems 35: Annual Conference on Neural Information Processing Systems 2022, NeurIPS 2022, New Orleans, LA, USA, November 28 - December 9, 2022}.

\bibitem[{Dettmers et~al.(2023{\natexlab{a}})Dettmers, Pagnoni, Holtzman, and Zettlemoyer}]{DBLP:journals/corr/abs-2305-14314}
Tim Dettmers, Artidoro Pagnoni, Ari Holtzman, and Luke Zettlemoyer. 2023{\natexlab{a}}.
\newblock \href {https://doi.org/10.48550/ARXIV.2305.14314} {Qlora: Efficient finetuning of quantized llms}.
\newblock \emph{CoRR}, abs/2305.14314.

\bibitem[{Dettmers et~al.(2023{\natexlab{b}})Dettmers, Svirschevski, Egiazarian, Kuznedelev, Frantar, Ashkboos, Borzunov, Hoefler, and Alistarh}]{DBLP:journals/corr/abs-2306-03078}
Tim Dettmers, Ruslan Svirschevski, Vage Egiazarian, Denis Kuznedelev, Elias Frantar, Saleh Ashkboos, Alexander Borzunov, Torsten Hoefler, and Dan Alistarh. 2023{\natexlab{b}}.
\newblock \href {https://doi.org/10.48550/ARXIV.2306.03078} {Spqr: {A} sparse-quantized representation for near-lossless {LLM} weight compression}.
\newblock \emph{CoRR}, abs/2306.03078.

\bibitem[{Du et~al.(2022)Du, Huang, Dai, Tong, Lepikhin, Xu, Krikun, Zhou, Yu, Firat, Zoph, Fedus, Bosma, Zhou, Wang, Wang, Webster, Pellat, Robinson, Meier{-}Hellstern, Duke, Dixon, Zhang, Le, Wu, Chen, and Cui}]{DBLP:conf/icml/DuHDTLXKZYFZFBZ22}
Nan Du, Yanping Huang, Andrew~M. Dai, Simon Tong, Dmitry Lepikhin, Yuanzhong Xu, Maxim Krikun, Yanqi Zhou, Adams~Wei Yu, Orhan Firat, Barret Zoph, Liam Fedus, Maarten~P. Bosma, Zongwei Zhou, Tao Wang, Yu~Emma Wang, Kellie Webster, Marie Pellat, Kevin Robinson, Kathleen~S. Meier{-}Hellstern, Toju Duke, Lucas Dixon, Kun Zhang, Quoc~V. Le, Yonghui Wu, Zhifeng Chen, and Claire Cui. 2022.
\newblock \href {https://proceedings.mlr.press/v162/du22c.html} {Glam: Efficient scaling of language models with mixture-of-experts}.
\newblock In \emph{International Conference on Machine Learning, {ICML} 2022, 17-23 July 2022, Baltimore, Maryland, {USA}}, volume 162 of \emph{Proceedings of Machine Learning Research}, pages 5547--5569. {PMLR}.

\bibitem[{Frantar et~al.(2022)Frantar, Ashkboos, Hoefler, and Alistarh}]{DBLP:journals/corr/abs-2210-17323}
Elias Frantar, Saleh Ashkboos, Torsten Hoefler, and Dan Alistarh. 2022.
\newblock \href {https://doi.org/10.48550/ARXIV.2210.17323} {{GPTQ:} accurate post-training quantization for generative pre-trained transformers}.
\newblock \emph{CoRR}, abs/2210.17323.

\bibitem[{Frantar et~al.(2023)Frantar, Ashkboos, Hoefler, and Alistarh}]{DBLP:conf/iclr/FrantarAHA23}
Elias Frantar, Saleh Ashkboos, Torsten Hoefler, and Dan Alistarh. 2023.
\newblock \href {https://openreview.net/pdf?id=tcbBPnfwxS} {{OPTQ:} accurate quantization for generative pre-trained transformers}.
\newblock In \emph{The Eleventh International Conference on Learning Representations, {ICLR} 2023, Kigali, Rwanda, May 1-5, 2023}. OpenReview.net.

\bibitem[{Gehman et~al.(2020)Gehman, Gururangan, Sap, Choi, and Smith}]{DBLP:conf/emnlp/GehmanGSCS20}
Samuel Gehman, Suchin Gururangan, Maarten Sap, Yejin Choi, and Noah~A. Smith. 2020.
\newblock \href {https://doi.org/10.18653/V1/2020.FINDINGS-EMNLP.301} {Realtoxicityprompts: Evaluating neural toxic degeneration in language models}.
\newblock In \emph{Findings of the Association for Computational Linguistics: {EMNLP} 2020, Online Event, 16-20 November 2020}, volume {EMNLP} 2020 of \emph{Findings of {ACL}}, pages 3356--3369. Association for Computational Linguistics.

\bibitem[{Gholami et~al.(2021)Gholami, Kim, Dong, Yao, Mahoney, and Keutzer}]{DBLP:journals/corr/abs-2103-13630}
Amir Gholami, Sehoon Kim, Zhen Dong, Zhewei Yao, Michael~W. Mahoney, and Kurt Keutzer. 2021.
\newblock \href {http://arxiv.org/abs/2103.13630} {A survey of quantization methods for efficient neural network inference}.
\newblock \emph{CoRR}, abs/2103.13630.

\bibitem[{Goyal et~al.(2022)Goyal, Gao, Chaudhary, Chen, Wenzek, Ju, Krishnan, Ranzato, Guzm{\'{a}}n, and Fan}]{DBLP:journals/tacl/GoyalGCCWJKRGF22}
Naman Goyal, Cynthia Gao, Vishrav Chaudhary, Peng{-}Jen Chen, Guillaume Wenzek, Da~Ju, Sanjana Krishnan, Marc'Aurelio Ranzato, Francisco Guzm{\'{a}}n, and Angela Fan. 2022.
\newblock \href {https://doi.org/10.1162/TACL\_A\_00474} {The flores-101 evaluation benchmark for low-resource and multilingual machine translation}.
\newblock \emph{Trans. Assoc. Comput. Linguistics}, 10:522--538.

\bibitem[{Guo et~al.(2023)Guo, Jin, Liu, Huang, Shi, Supryadi, Yu, Liu, Li, Xiong, and Xiong}]{DBLP:journals/corr/abs-2310-19736}
Zishan Guo, Renren Jin, Chuang Liu, Yufei Huang, Dan Shi, Supryadi, Linhao Yu, Yan Liu, Jiaxuan Li, Bojian Xiong, and Deyi Xiong. 2023.
\newblock \href {https://doi.org/10.48550/ARXIV.2310.19736} {Evaluating large language models: {A} comprehensive survey}.
\newblock \emph{CoRR}, abs/2310.19736.

\bibitem[{Hendrycks et~al.(2021)Hendrycks, Burns, Basart, Zou, Mazeika, Song, and Steinhardt}]{DBLP:conf/iclr/HendrycksBBZMSS21}
Dan Hendrycks, Collin Burns, Steven Basart, Andy Zou, Mantas Mazeika, Dawn Song, and Jacob Steinhardt. 2021.
\newblock \href {https://openreview.net/forum?id=d7KBjmI3GmQ} {Measuring massive multitask language understanding}.
\newblock In \emph{9th International Conference on Learning Representations, {ICLR} 2021, Virtual Event, Austria, May 3-7, 2021}. OpenReview.net.

\bibitem[{Hermann et~al.(2015)Hermann, Kocisk{\'{y}}, Grefenstette, Espeholt, Kay, Suleyman, and Blunsom}]{DBLP:conf/nips/HermannKGEKSB15}
Karl~Moritz Hermann, Tom{\'{a}}s Kocisk{\'{y}}, Edward Grefenstette, Lasse Espeholt, Will Kay, Mustafa Suleyman, and Phil Blunsom. 2015.
\newblock \href {https://proceedings.neurips.cc/paper/2015/hash/afdec7005cc9f14302cd0474fd0f3c96-Abstract.html} {Teaching machines to read and comprehend}.
\newblock In \emph{Advances in Neural Information Processing Systems 28: Annual Conference on Neural Information Processing Systems 2015, December 7-12, 2015, Montreal, Quebec, Canada}, pages 1693--1701.

\bibitem[{Hu et~al.(2023)Hu, Liu, Han, Zhang, He, Zhao, Lin, Ding, Ou, Zeng, Liu, and Sun}]{DBLP:journals/corr/abs-2310-03262}
Shengding Hu, Xin Liu, Xu~Han, Xinrong Zhang, Chaoqun He, Weilin Zhao, Yankai Lin, Ning Ding, Zebin Ou, Guoyang Zeng, Zhiyuan Liu, and Maosong Sun. 2023.
\newblock \href {https://doi.org/10.48550/ARXIV.2310.03262} {Unlock predictable scaling from emergent abilities}.
\newblock \emph{CoRR}, abs/2310.03262.

\bibitem[{Huang and Xiong(2023)}]{DBLP:journals/corr/abs-2306-16244}
Yufei Huang and Deyi Xiong. 2023.
\newblock \href {https://doi.org/10.48550/ARXIV.2306.16244} {{CBBQ:} {A} chinese bias benchmark dataset curated with human-ai collaboration for large language models}.
\newblock \emph{CoRR}, abs/2306.16244.

\bibitem[{Huang et~al.(2023)Huang, Bai, Zhu, Zhang, Zhang, Su, Liu, Lv, Zhang, jiayi lei, Fu, Sun, and He}]{huang2023ceval}
Yuzhen Huang, Yuzhuo Bai, Zhihao Zhu, Junlei Zhang, Jinghan Zhang, Tangjun Su, Junteng Liu, Chuancheng Lv, Yikai Zhang, jiayi lei, Yao Fu, Maosong Sun, and Junxian He. 2023.
\newblock \href {https://openreview.net/forum?id=fOrm2rGX2r} {C-eval: A multi-level multi-discipline chinese evaluation suite for foundation models}.
\newblock In \emph{Thirty-seventh Conference on Neural Information Processing Systems Datasets and Benchmarks Track}.

\bibitem[{Jacob et~al.(2018)Jacob, Kligys, Chen, Zhu, Tang, Howard, Adam, and Kalenichenko}]{DBLP:conf/cvpr/JacobKCZTHAK18}
Benoit Jacob, Skirmantas Kligys, Bo~Chen, Menglong Zhu, Matthew Tang, Andrew~G. Howard, Hartwig Adam, and Dmitry Kalenichenko. 2018.
\newblock \href {https://doi.org/10.1109/CVPR.2018.00286} {Quantization and training of neural networks for efficient integer-arithmetic-only inference}.
\newblock In \emph{2018 {IEEE} Conference on Computer Vision and Pattern Recognition, {CVPR} 2018, Salt Lake City, UT, USA, June 18-22, 2018}, pages 2704--2713. Computer Vision Foundation / {IEEE} Computer Society.

\bibitem[{Jiang et~al.(2023)Jiang, Wang, Zeng, Zhong, Li, Mi, Shang, Jiang, Liu, and Wang}]{DBLP:journals/corr/abs-2310-20410}
Yuxin Jiang, Yufei Wang, Xingshan Zeng, Wanjun Zhong, Liangyou Li, Fei Mi, Lifeng Shang, Xin Jiang, Qun Liu, and Wei Wang. 2023.
\newblock \href {https://doi.org/10.48550/ARXIV.2310.20410} {Followbench: {A} multi-level fine-grained constraints following benchmark for large language models}.
\newblock \emph{CoRR}, abs/2310.20410.

\bibitem[{Kim et~al.(2023)Kim, Lee, Kim, Park, Yoo, Kwon, and Lee}]{DBLP:journals/corr/abs-2305-14152}
Jeonghoon Kim, Jung~Hyun Lee, Sungdong Kim, Joonsuk Park, Kang~Min Yoo, Se~Jung Kwon, and Dongsoo Lee. 2023.
\newblock \href {https://doi.org/10.48550/ARXIV.2305.14152} {Memory-efficient fine-tuning of compressed large language models via sub-4-bit integer quantization}.
\newblock \emph{CoRR}, abs/2305.14152.

\bibitem[{Lai et~al.(2023)Lai, Ngo, Veyseh, Man, Dernoncourt, Bui, and Nguyen}]{DBLP:conf/emnlp/LaiNVMDBN23}
Viet~Dac Lai, Nghia~Trung Ngo, Amir Pouran~Ben Veyseh, Hieu Man, Franck Dernoncourt, Trung Bui, and Thien~Huu Nguyen. 2023.
\newblock \href {https://aclanthology.org/2023.findings-emnlp.878} {Chatgpt beyond english: Towards a comprehensive evaluation of large language models in multilingual learning}.
\newblock In \emph{Findings of the Association for Computational Linguistics: {EMNLP} 2023, Singapore, December 6-10, 2023}, pages 13171--13189. Association for Computational Linguistics.

\bibitem[{Laskar et~al.(2023)Laskar, Bari, Rahman, Bhuiyan, Joty, and Huang}]{DBLP:conf/acl/LaskarBRBJH23}
Md. Tahmid~Rahman Laskar, M.~Saiful Bari, Mizanur Rahman, Md~Amran~Hossen Bhuiyan, Shafiq Joty, and Jimmy~X. Huang. 2023.
\newblock \href {https://doi.org/10.18653/V1/2023.FINDINGS-ACL.29} {A systematic study and comprehensive evaluation of chatgpt on benchmark datasets}.
\newblock In \emph{Findings of the Association for Computational Linguistics: {ACL} 2023, Toronto, Canada, July 9-14, 2023}, pages 431--469. Association for Computational Linguistics.

\bibitem[{Lee et~al.(2023)Lee, Jin, Kim, Kim, and Park}]{DBLP:journals/corr/abs-2306-02272}
Changhun Lee, Jungyu Jin, Taesu Kim, Hyungjun Kim, and Eunhyeok Park. 2023.
\newblock \href {https://doi.org/10.48550/ARXIV.2306.02272} {{OWQ:} lessons learned from activation outliers for weight quantization in large language models}.
\newblock \emph{CoRR}, abs/2306.02272.

\bibitem[{Li et~al.(2023)Li, Zhang, Koto, Yang, Zhao, Gong, Duan, and Baldwin}]{DBLP:journals/corr/abs-2306-09212}
Haonan Li, Yixuan Zhang, Fajri Koto, Yifei Yang, Hai Zhao, Yeyun Gong, Nan Duan, and Timothy Baldwin. 2023.
\newblock \href {https://doi.org/10.48550/ARXIV.2306.09212} {{CMMLU:} measuring massive multitask language understanding in chinese}.
\newblock \emph{CoRR}, abs/2306.09212.

\bibitem[{Li(2023)}]{DBLP:journals/corr/abs-2309-10677}
Yucheng Li. 2023.
\newblock \href {https://doi.org/10.48550/ARXIV.2309.10677} {Estimating contamination via perplexity: Quantifying memorisation in language model evaluation}.
\newblock \emph{CoRR}, abs/2309.10677.

\bibitem[{Liang et~al.(2022)Liang, Bommasani, Lee, Tsipras, Soylu, Yasunaga, Zhang, Narayanan, Wu, Kumar, Newman, Yuan, Yan, Zhang, Cosgrove, Manning, R{\'{e}}, Acosta{-}Navas, Hudson, Zelikman, Durmus, Ladhak, Rong, Ren, Yao, Wang, Santhanam, Orr, Zheng, Y{\"{u}}ksekg{\"{o}}n{\"{u}}l, Suzgun, Kim, Guha, Chatterji, Khattab, Henderson, Huang, Chi, Xie, Santurkar, Ganguli, Hashimoto, Icard, Zhang, Chaudhary, Wang, Li, Mai, Zhang, and Koreeda}]{DBLP:journals/corr/abs-2211-09110}
Percy Liang, Rishi Bommasani, Tony Lee, Dimitris Tsipras, Dilara Soylu, Michihiro Yasunaga, Yian Zhang, Deepak Narayanan, Yuhuai Wu, Ananya Kumar, Benjamin Newman, Binhang Yuan, Bobby Yan, Ce~Zhang, Christian Cosgrove, Christopher~D. Manning, Christopher R{\'{e}}, Diana Acosta{-}Navas, Drew~A. Hudson, Eric Zelikman, Esin Durmus, Faisal Ladhak, Frieda Rong, Hongyu Ren, Huaxiu Yao, Jue Wang, Keshav Santhanam, Laurel~J. Orr, Lucia Zheng, Mert Y{\"{u}}ksekg{\"{o}}n{\"{u}}l, Mirac Suzgun, Nathan Kim, Neel Guha, Niladri~S. Chatterji, Omar Khattab, Peter Henderson, Qian Huang, Ryan Chi, Sang~Michael Xie, Shibani Santurkar, Surya Ganguli, Tatsunori Hashimoto, Thomas Icard, Tianyi Zhang, Vishrav Chaudhary, William Wang, Xuechen Li, Yifan Mai, Yuhui Zhang, and Yuta Koreeda. 2022.
\newblock \href {https://doi.org/10.48550/ARXIV.2211.09110} {Holistic evaluation of language models}.
\newblock \emph{CoRR}, abs/2211.09110.

\bibitem[{Lin et~al.(2023)Lin, Tang, Tang, Yang, Dang, and Han}]{DBLP:journals/corr/abs-2306-00978}
Ji~Lin, Jiaming Tang, Haotian Tang, Shang Yang, Xingyu Dang, and Song Han. 2023.
\newblock \href {https://doi.org/10.48550/ARXIV.2306.00978} {{AWQ:} activation-aware weight quantization for {LLM} compression and acceleration}.
\newblock \emph{CoRR}, abs/2306.00978.

\bibitem[{Lin et~al.(2022)Lin, Hilton, and Evans}]{DBLP:conf/acl/LinHE22}
Stephanie Lin, Jacob Hilton, and Owain Evans. 2022.
\newblock \href {https://doi.org/10.18653/V1/2022.ACL-LONG.229} {Truthfulqa: Measuring how models mimic human falsehoods}.
\newblock In \emph{Proceedings of the 60th Annual Meeting of the Association for Computational Linguistics (Volume 1: Long Papers), {ACL} 2022, Dublin, Ireland, May 22-27, 2022}, pages 3214--3252. Association for Computational Linguistics.

\bibitem[{Liu et~al.(2024{\natexlab{a}})Liu, Jin, Ren, and Xiong}]{DBLP:conf/coling/LiuJRX24}
Chuang Liu, Renren Jin, Yuqi Ren, and Deyi Xiong. 2024{\natexlab{a}}.
\newblock \href {https://aclanthology.org/2024.lrec-main.916} {{LHMKE:} {A} large-scale holistic multi-subject knowledge evaluation benchmark for chinese large language models}.
\newblock In \emph{Proceedings of the 2024 Joint International Conference on Computational Linguistics, Language Resources and Evaluation, {LREC/COLING} 2024, 20-25 May, 2024, Torino, Italy}, pages 10476--10487. {ELRA} and {ICCL}.

\bibitem[{Liu et~al.(2023{\natexlab{a}})Liu, Jin, Ren, Yu, Dong, Peng, Zhang, Peng, Zhang, Lyu, Su, Liu, and Xiong}]{DBLP:journals/corr/abs-2305-10263}
Chuang Liu, Renren Jin, Yuqi Ren, Linhao Yu, Tianyu Dong, Xiaohan Peng, Shuting Zhang, Jianxiang Peng, Peiyi Zhang, Qingqing Lyu, Xiaowen Su, Qun Liu, and Deyi Xiong. 2023{\natexlab{a}}.
\newblock \href {https://doi.org/10.48550/ARXIV.2305.10263} {{M3KE:} {A} massive multi-level multi-subject knowledge evaluation benchmark for chinese large language models}.
\newblock \emph{CoRR}, abs/2305.10263.

\bibitem[{Liu et~al.(2023{\natexlab{b}})Liu, Liu, Gao, Gao, Zhao, Li, Ding, and Wen}]{DBLP:journals/corr/abs-2307-08072}
Peiyu Liu, Zikang Liu, Ze{-}Feng Gao, Dawei Gao, Wayne~Xin Zhao, Yaliang Li, Bolin Ding, and Ji{-}Rong Wen. 2023{\natexlab{b}}.
\newblock \href {https://doi.org/10.48550/ARXIV.2307.08072} {Do emergent abilities exist in quantized large language models: An empirical study}.
\newblock \emph{CoRR}, abs/2307.08072.

\bibitem[{Liu et~al.(2023{\natexlab{c}})Liu, Lei, Wang, Huang, Feng, Wen, Cheng, Ke, Xu, Tam, Zhang, Sun, Wang, Zhang, Huang, Dong, and Tang}]{DBLP:journals/corr/abs-2311-18743}
Xiao Liu, Xuanyu Lei, Shengyuan Wang, Yue Huang, Zhuoer Feng, Bosi Wen, Jiale Cheng, Pei Ke, Yifan Xu, Weng~Lam Tam, Xiaohan Zhang, Lichao Sun, Hongning Wang, Jing Zhang, Minlie Huang, Yuxiao Dong, and Jie Tang. 2023{\natexlab{c}}.
\newblock \href {https://doi.org/10.48550/ARXIV.2311.18743} {Alignbench: Benchmarking chinese alignment of large language models}.
\newblock \emph{CoRR}, abs/2311.18743.

\bibitem[{Liu et~al.(2024{\natexlab{b}})Liu, Jin, Shi, Yao, and Xiong}]{DBLP:journals/corr/abs-2403-07747}
Yan Liu, Renren Jin, Lin Shi, Zheng Yao, and Deyi Xiong. 2024{\natexlab{b}}.
\newblock \href {https://doi.org/10.48550/ARXIV.2403.07747} {Finemath: {A} fine-grained mathematical evaluation benchmark for chinese large language models}.
\newblock \emph{CoRR}, abs/2403.07747.

\bibitem[{Liu et~al.(2023{\natexlab{d}})Liu, Yao, Ton, Zhang, Guo, Cheng, Klochkov, Taufiq, and Li}]{DBLP:journals/corr/abs-2308-05374}
Yang Liu, Yuanshun Yao, Jean{-}Francois Ton, Xiaoying Zhang, Ruocheng Guo, Hao Cheng, Yegor Klochkov, Muhammad~Faaiz Taufiq, and Hang Li. 2023{\natexlab{d}}.
\newblock \href {https://doi.org/10.48550/ARXIV.2308.05374} {Trustworthy llms: a survey and guideline for evaluating large language models' alignment}.
\newblock \emph{CoRR}, abs/2308.05374.

\bibitem[{Liu et~al.(2023{\natexlab{e}})Liu, Oguz, Zhao, Chang, Stock, Mehdad, Shi, Krishnamoorthi, and Chandra}]{DBLP:journals/corr/abs-2305-17888}
Zechun Liu, Barlas Oguz, Changsheng Zhao, Ernie Chang, Pierre Stock, Yashar Mehdad, Yangyang Shi, Raghuraman Krishnamoorthi, and Vikas Chandra. 2023{\natexlab{e}}.
\newblock \href {https://doi.org/10.48550/ARXIV.2305.17888} {{LLM-QAT:} data-free quantization aware training for large language models}.
\newblock \emph{CoRR}, abs/2305.17888.

\bibitem[{Lu et~al.(2023)Lu, Bigoulaeva, Sachdeva, Madabushi, and Gurevych}]{DBLP:journals/corr/abs-2309-01809}
Sheng Lu, Irina Bigoulaeva, Rachneet Sachdeva, Harish~Tayyar Madabushi, and Iryna Gurevych. 2023.
\newblock \href {https://doi.org/10.48550/ARXIV.2309.01809} {Are emergent abilities in large language models just in-context learning?}
\newblock \emph{CoRR}, abs/2309.01809.

\bibitem[{Mao et~al.(2023)Mao, Chen, Zhang, Guerin, and Cambria}]{DBLP:journals/corr/abs-2308-12488}
Rui Mao, Guanyi Chen, Xulang Zhang, Frank Guerin, and Erik Cambria. 2023.
\newblock \href {https://doi.org/10.48550/ARXIV.2308.12488} {Gpteval: {A} survey on assessments of chatgpt and {GPT-4}}.
\newblock \emph{CoRR}, abs/2308.12488.

\bibitem[{Marcus et~al.(1994)Marcus, Kim, Marcinkiewicz, MacIntyre, Bies, Ferguson, Katz, and Schasberger}]{DBLP:conf/naacl/MarcusKMMBFKS94}
Mitchell~P. Marcus, Grace Kim, Mary~Ann Marcinkiewicz, Robert MacIntyre, Ann Bies, Mark Ferguson, Karen Katz, and Britta Schasberger. 1994.
\newblock \href {https://aclanthology.org/H94-1020/} {The penn treebank: Annotating predicate argument structure}.
\newblock In \emph{Human Language Technology, Proceedings of a Workshop held at Plainsboro, New Jerey, USA, March 8-11, 1994}. Morgan Kaufmann.

\bibitem[{Merity et~al.(2017)Merity, Xiong, Bradbury, and Socher}]{DBLP:conf/iclr/MerityX0S17}
Stephen Merity, Caiming Xiong, James Bradbury, and Richard Socher. 2017.
\newblock \href {https://openreview.net/forum?id=Byj72udxe} {Pointer sentinel mixture models}.
\newblock In \emph{5th International Conference on Learning Representations, {ICLR} 2017, Toulon, France, April 24-26, 2017, Conference Track Proceedings}. OpenReview.net.

\bibitem[{Nallapati et~al.(2016)Nallapati, Zhou, dos Santos, G{\"{u}}l{\c{c}}ehre, and Xiang}]{DBLP:conf/conll/NallapatiZSGX16}
Ramesh Nallapati, Bowen Zhou, C{\'{\i}}cero~Nogueira dos Santos, {\c{C}}aglar G{\"{u}}l{\c{c}}ehre, and Bing Xiang. 2016.
\newblock \href {https://doi.org/10.18653/V1/K16-1028} {Abstractive text summarization using sequence-to-sequence rnns and beyond}.
\newblock In \emph{Proceedings of the 20th {SIGNLL} Conference on Computational Natural Language Learning, CoNLL 2016, Berlin, Germany, August 11-12, 2016}, pages 280--290. {ACL}.

\bibitem[{Narayan et~al.(2018)Narayan, Cohen, and Lapata}]{DBLP:conf/emnlp/NarayanCL18}
Shashi Narayan, Shay~B. Cohen, and Mirella Lapata. 2018.
\newblock \href {https://doi.org/10.18653/V1/D18-1206} {Don't give me the details, just the summary! topic-aware convolutional neural networks for extreme summarization}.
\newblock In \emph{Proceedings of the 2018 Conference on Empirical Methods in Natural Language Processing, Brussels, Belgium, October 31 - November 4, 2018}, pages 1797--1807. Association for Computational Linguistics.

\bibitem[{Oren et~al.(2023)Oren, Meister, Chatterji, Ladhak, and Hashimoto}]{DBLP:journals/corr/abs-2310-17623}
Yonatan Oren, Nicole Meister, Niladri~S. Chatterji, Faisal Ladhak, and Tatsunori~B. Hashimoto. 2023.
\newblock \href {https://doi.org/10.48550/ARXIV.2310.17623} {Proving test set contamination in black box language models}.
\newblock \emph{CoRR}, abs/2310.17623.

\bibitem[{Ouyang et~al.(2022)Ouyang, Wu, Jiang, Almeida, Wainwright, Mishkin, Zhang, Agarwal, Slama, Ray, Schulman, Hilton, Kelton, Miller, Simens, Askell, Welinder, Christiano, Leike, and Lowe}]{DBLP:conf/nips/Ouyang0JAWMZASR22}
Long Ouyang, Jeffrey Wu, Xu~Jiang, Diogo Almeida, Carroll~L. Wainwright, Pamela Mishkin, Chong Zhang, Sandhini Agarwal, Katarina Slama, Alex Ray, John Schulman, Jacob Hilton, Fraser Kelton, Luke Miller, Maddie Simens, Amanda Askell, Peter Welinder, Paul~F. Christiano, Jan Leike, and Ryan Lowe. 2022.
\newblock \href {http://papers.nips.cc/paper\_files/paper/2022/hash/b1efde53be364a73914f58805a001731-Abstract-Conference.html} {Training language models to follow instructions with human feedback}.
\newblock In \emph{Advances in Neural Information Processing Systems 35: Annual Conference on Neural Information Processing Systems 2022, NeurIPS 2022, New Orleans, LA, USA, November 28 - December 9, 2022}.

\bibitem[{Parrish et~al.(2022)Parrish, Chen, Nangia, Padmakumar, Phang, Thompson, Htut, and Bowman}]{DBLP:conf/acl/ParrishCNPPTHB22}
Alicia Parrish, Angelica Chen, Nikita Nangia, Vishakh Padmakumar, Jason Phang, Jana Thompson, Phu~Mon Htut, and Samuel~R. Bowman. 2022.
\newblock \href {https://doi.org/10.18653/V1/2022.FINDINGS-ACL.165} {{BBQ:} {A} hand-built bias benchmark for question answering}.
\newblock In \emph{Findings of the Association for Computational Linguistics: {ACL} 2022, Dublin, Ireland, May 22-27, 2022}, pages 2086--2105. Association for Computational Linguistics.

\bibitem[{Peng et~al.(2023)Peng, Li, He, Galley, and Gao}]{DBLP:journals/corr/abs-2304-03277}
Baolin Peng, Chunyuan Li, Pengcheng He, Michel Galley, and Jianfeng Gao. 2023.
\newblock \href {https://doi.org/10.48550/ARXIV.2304.03277} {Instruction tuning with {GPT-4}}.
\newblock \emph{CoRR}, abs/2304.03277.

\bibitem[{Qin et~al.(2023)Qin, Liang, Ye, Zhu, Yan, Lu, Lin, Cong, Tang, Qian, Zhao, Tian, Xie, Zhou, Gerstein, Li, Liu, and Sun}]{DBLP:journals/corr/abs-2307-16789}
Yujia Qin, Shihao Liang, Yining Ye, Kunlun Zhu, Lan Yan, Yaxi Lu, Yankai Lin, Xin Cong, Xiangru Tang, Bill Qian, Sihan Zhao, Runchu Tian, Ruobing Xie, Jie Zhou, Mark Gerstein, Dahai Li, Zhiyuan Liu, and Maosong Sun. 2023.
\newblock \href {https://doi.org/10.48550/ARXIV.2307.16789} {Toolllm: Facilitating large language models to master 16000+ real-world apis}.
\newblock \emph{CoRR}, abs/2307.16789.

\bibitem[{Raffel et~al.(2020)Raffel, Shazeer, Roberts, Lee, Narang, Matena, Zhou, Li, and Liu}]{DBLP:journals/jmlr/RaffelSRLNMZLL20}
Colin Raffel, Noam Shazeer, Adam Roberts, Katherine Lee, Sharan Narang, Michael Matena, Yanqi Zhou, Wei Li, and Peter~J. Liu. 2020.
\newblock \href {http://jmlr.org/papers/v21/20-074.html} {Exploring the limits of transfer learning with a unified text-to-text transformer}.
\newblock \emph{J. Mach. Learn. Res.}, 21:140:1--140:67.

\bibitem[{Ren et~al.(2023)Ren, Zhou, Meng, Huang, Wang, Wang, Li, Zhang, Podolskiy, Arshinov, Bout, Piontkovskaya, Wei, Jiang, Su, Liu, and Yao}]{DBLP:journals/corr/abs-2303-10845}
Xiaozhe Ren, Pingyi Zhou, Xinfan Meng, Xinjing Huang, Yadao Wang, Weichao Wang, Pengfei Li, Xiaoda Zhang, Alexander Podolskiy, Grigory Arshinov, Andrey Bout, Irina Piontkovskaya, Jiansheng Wei, Xin Jiang, Teng Su, Qun Liu, and Jun Yao. 2023.
\newblock \href {https://doi.org/10.48550/ARXIV.2303.10845} {Pangu-{\(\Sigma\)}: Towards trillion parameter language model with sparse heterogeneous computing}.
\newblock \emph{CoRR}, abs/2303.10845.

\bibitem[{Scao et~al.(2022)Scao, Fan, Akiki, Pavlick, Ilic, Hesslow, Castagn{\'{e}}, Luccioni, Yvon, Gall{\'{e}}, Tow, Rush, Biderman, Webson, Ammanamanchi, Wang, Sagot, Muennighoff, del Moral, Ruwase, Bawden, Bekman, McMillan{-}Major, Beltagy, Nguyen, Saulnier, Tan, Suarez, Sanh, Lauren{\c{c}}on, Jernite, Launay, Mitchell, Raffel, Gokaslan, Simhi, Soroa, Aji, Alfassy, Rogers, Nitzav, Xu, Mou, Emezue, Klamm, Leong, van Strien, Adelani, and et~al.}]{DBLP:journals/corr/abs-2211-05100}
Teven~Le Scao, Angela Fan, Christopher Akiki, Ellie Pavlick, Suzana Ilic, Daniel Hesslow, Roman Castagn{\'{e}}, Alexandra~Sasha Luccioni, Fran{\c{c}}ois Yvon, Matthias Gall{\'{e}}, Jonathan Tow, Alexander~M. Rush, Stella Biderman, Albert Webson, Pawan~Sasanka Ammanamanchi, Thomas Wang, Beno{\^{\i}}t Sagot, Niklas Muennighoff, Albert~Villanova del Moral, Olatunji Ruwase, Rachel Bawden, Stas Bekman, Angelina McMillan{-}Major, Iz~Beltagy, Huu Nguyen, Lucile Saulnier, Samson Tan, Pedro~Ortiz Suarez, Victor Sanh, Hugo Lauren{\c{c}}on, Yacine Jernite, Julien Launay, Margaret Mitchell, Colin Raffel, Aaron Gokaslan, Adi Simhi, Aitor Soroa, Alham~Fikri Aji, Amit Alfassy, Anna Rogers, Ariel~Kreisberg Nitzav, Canwen Xu, Chenghao Mou, Chris Emezue, Christopher Klamm, Colin Leong, Daniel van Strien, David~Ifeoluwa Adelani, and et~al. 2022.
\newblock \href {https://doi.org/10.48550/ARXIV.2211.05100} {{BLOOM:} {A} 176b-parameter open-access multilingual language model}.
\newblock \emph{CoRR}, abs/2211.05100.

\bibitem[{Schaeffer et~al.(2023)Schaeffer, Miranda, and Koyejo}]{schaeffer2023are}
Rylan Schaeffer, Brando Miranda, and Sanmi Koyejo. 2023.
\newblock \href {https://openreview.net/forum?id=ITw9edRDlD} {Are emergent abilities of large language models a mirage?}
\newblock In \emph{Thirty-seventh Conference on Neural Information Processing Systems}.

\bibitem[{See et~al.(2017)See, Liu, and Manning}]{DBLP:conf/acl/SeeLM17}
Abigail See, Peter~J. Liu, and Christopher~D. Manning. 2017.
\newblock \href {https://doi.org/10.18653/V1/P17-1099} {Get to the point: Summarization with pointer-generator networks}.
\newblock In \emph{Proceedings of the 55th Annual Meeting of the Association for Computational Linguistics, {ACL} 2017, Vancouver, Canada, July 30 - August 4, Volume 1: Long Papers}, pages 1073--1083. Association for Computational Linguistics.

\bibitem[{Shazeer(2019)}]{DBLP:journals/corr/abs-1911-02150}
Noam Shazeer. 2019.
\newblock \href {http://arxiv.org/abs/1911.02150} {Fast transformer decoding: One write-head is all you need}.
\newblock \emph{CoRR}, abs/1911.02150.

\bibitem[{Shen et~al.(2023)Shen, Li, and Xiong}]{DBLP:journals/corr/abs-2312-16132}
Tianhao Shen, Sun Li, and Deyi Xiong. 2023.
\newblock \href {https://doi.org/10.48550/ARXIV.2312.16132} {Roleeval: {A} bilingual role evaluation benchmark for large language models}.
\newblock \emph{CoRR}, abs/2312.16132.

\bibitem[{Shi et~al.(2024)Shi, You, Huang, Li, and Xiong}]{DBLP:conf/aaai/ShiYHLX24}
Dan Shi, Chaobin You, Jiantao Huang, Taihao Li, and Deyi Xiong. 2024.
\newblock \href {https://doi.org/10.1609/AAAI.V38I17.29861} {{CORECODE:} {A} common sense annotated dialogue dataset with benchmark tasks for chinese large language models}.
\newblock In \emph{Thirty-Eighth {AAAI} Conference on Artificial Intelligence, {AAAI} 2024, Thirty-Sixth Conference on Innovative Applications of Artificial Intelligence, {IAAI} 2024, Fourteenth Symposium on Educational Advances in Artificial Intelligence, {EAAI} 2014, February 20-27, 2024, Vancouver, Canada}, pages 18952--18960. {AAAI} Press.

\bibitem[{Srivastava et~al.(2022)Srivastava, Rastogi, Rao, Shoeb, Abid, Fisch, Brown, Santoro, Gupta, Garriga{-}Alonso, Kluska, Lewkowycz, Agarwal, Power, Ray, Warstadt, Kocurek, Safaya, Tazarv, Xiang, Parrish, Nie, Hussain, Askell, Dsouza, Rahane, Iyer, Andreassen, Santilli, Stuhlm{\"{u}}ller, Dai, La, Lampinen, Zou, Jiang, Chen, Vuong, Gupta, Gottardi, Norelli, Venkatesh, Gholamidavoodi, Tabassum, Menezes, Kirubarajan, Mullokandov, Sabharwal, Herrick, Efrat, Erdem, Karakas, and et~al.}]{DBLP:journals/corr/abs-2206-04615}
Aarohi Srivastava, Abhinav Rastogi, Abhishek Rao, Abu Awal~Md Shoeb, Abubakar Abid, Adam Fisch, Adam~R. Brown, Adam Santoro, Aditya Gupta, Adri{\`{a}} Garriga{-}Alonso, Agnieszka Kluska, Aitor Lewkowycz, Akshat Agarwal, Alethea Power, Alex Ray, Alex Warstadt, Alexander~W. Kocurek, Ali Safaya, Ali Tazarv, Alice Xiang, Alicia Parrish, Allen Nie, Aman Hussain, Amanda Askell, Amanda Dsouza, Ameet Rahane, Anantharaman~S. Iyer, Anders Andreassen, Andrea Santilli, Andreas Stuhlm{\"{u}}ller, Andrew~M. Dai, Andrew La, Andrew~K. Lampinen, Andy Zou, Angela Jiang, Angelica Chen, Anh Vuong, Animesh Gupta, Anna Gottardi, Antonio Norelli, Anu Venkatesh, Arash Gholamidavoodi, Arfa Tabassum, Arul Menezes, Arun Kirubarajan, Asher Mullokandov, Ashish Sabharwal, Austin Herrick, Avia Efrat, Aykut Erdem, Ayla Karakas, and et~al. 2022.
\newblock \href {https://doi.org/10.48550/ARXIV.2206.04615} {Beyond the imitation game: Quantifying and extrapolating the capabilities of language models}.
\newblock \emph{CoRR}, abs/2206.04615.

\bibitem[{Taori et~al.(2023)Taori, Gulrajani, Zhang, Dubois, Li, Guestrin, Liang, and Hashimoto}]{alpaca}
Rohan Taori, Ishaan Gulrajani, Tianyi Zhang, Yann Dubois, Xuechen Li, Carlos Guestrin, Percy Liang, and Tatsunori~B. Hashimoto. 2023.
\newblock Stanford alpaca: An instruction-following llama model.
\newblock \url{https://github.com/tatsu-lab/stanford_alpaca}.

\bibitem[{Touvron et~al.(2023{\natexlab{a}})Touvron, Lavril, Izacard, Martinet, Lachaux, Lacroix, Rozi{\`{e}}re, Goyal, Hambro, Azhar, Rodriguez, Joulin, Grave, and Lample}]{DBLP:journals/corr/abs-2302-13971}
Hugo Touvron, Thibaut Lavril, Gautier Izacard, Xavier Martinet, Marie{-}Anne Lachaux, Timoth{\'{e}}e Lacroix, Baptiste Rozi{\`{e}}re, Naman Goyal, Eric Hambro, Faisal Azhar, Aur{\'{e}}lien Rodriguez, Armand Joulin, Edouard Grave, and Guillaume Lample. 2023{\natexlab{a}}.
\newblock \href {https://doi.org/10.48550/ARXIV.2302.13971} {Llama: Open and efficient foundation language models}.
\newblock \emph{CoRR}, abs/2302.13971.

\bibitem[{Touvron et~al.(2023{\natexlab{b}})Touvron, Martin, Stone, Albert, Almahairi, Babaei, Bashlykov, Batra, Bhargava, Bhosale, Bikel, Blecher, Canton{-}Ferrer, Chen, Cucurull, Esiobu, Fernandes, Fu, Fu, Fuller, Gao, Goswami, Goyal, Hartshorn, Hosseini, Hou, Inan, Kardas, Kerkez, Khabsa, Kloumann, Korenev, Koura, Lachaux, Lavril, Lee, Liskovich, Lu, Mao, Martinet, Mihaylov, Mishra, Molybog, Nie, Poulton, Reizenstein, Rungta, Saladi, Schelten, Silva, Smith, Subramanian, Tan, Tang, Taylor, Williams, Kuan, Xu, Yan, Zarov, Zhang, Fan, Kambadur, Narang, Rodriguez, Stojnic, Edunov, and Scialom}]{DBLP:journals/corr/abs-2307-09288}
Hugo Touvron, Louis Martin, Kevin Stone, Peter Albert, Amjad Almahairi, Yasmine Babaei, Nikolay Bashlykov, Soumya Batra, Prajjwal Bhargava, Shruti Bhosale, Dan Bikel, Lukas Blecher, Cristian Canton{-}Ferrer, Moya Chen, Guillem Cucurull, David Esiobu, Jude Fernandes, Jeremy Fu, Wenyin Fu, Brian Fuller, Cynthia Gao, Vedanuj Goswami, Naman Goyal, Anthony Hartshorn, Saghar Hosseini, Rui Hou, Hakan Inan, Marcin Kardas, Viktor Kerkez, Madian Khabsa, Isabel Kloumann, Artem Korenev, Punit~Singh Koura, Marie{-}Anne Lachaux, Thibaut Lavril, Jenya Lee, Diana Liskovich, Yinghai Lu, Yuning Mao, Xavier Martinet, Todor Mihaylov, Pushkar Mishra, Igor Molybog, Yixin Nie, Andrew Poulton, Jeremy Reizenstein, Rashi Rungta, Kalyan Saladi, Alan Schelten, Ruan Silva, Eric~Michael Smith, Ranjan Subramanian, Xiaoqing~Ellen Tan, Binh Tang, Ross Taylor, Adina Williams, Jian~Xiang Kuan, Puxin Xu, Zheng Yan, Iliyan Zarov, Yuchen Zhang, Angela Fan, Melanie Kambadur, Sharan Narang, Aur{\'{e}}lien Rodriguez, Robert Stojnic, Sergey Edunov,
  and Thomas Scialom. 2023{\natexlab{b}}.
\newblock \href {https://doi.org/10.48550/ARXIV.2307.09288} {Llama 2: Open foundation and fine-tuned chat models}.
\newblock \emph{CoRR}, abs/2307.09288.

\bibitem[{Wei et~al.(2022)Wei, Tay, Bommasani, Raffel, Zoph, Borgeaud, Yogatama, Bosma, Zhou, Metzler, Chi, Hashimoto, Vinyals, Liang, Dean, and Fedus}]{DBLP:journals/tmlr/WeiTBRZBYBZMCHVLDF22}
Jason Wei, Yi~Tay, Rishi Bommasani, Colin Raffel, Barret Zoph, Sebastian Borgeaud, Dani Yogatama, Maarten Bosma, Denny Zhou, Donald Metzler, Ed~H. Chi, Tatsunori Hashimoto, Oriol Vinyals, Percy Liang, Jeff Dean, and William Fedus. 2022.
\newblock \href {https://openreview.net/forum?id=yzkSU5zdwD} {Emergent abilities of large language models}.
\newblock \emph{Trans. Mach. Learn. Res.}, 2022.

\bibitem[{Wei et~al.(2023)Wei, Zhang, Li, Zhang, Gong, Guo, and Liu}]{DBLP:conf/emnlp/WeiZLZGG023}
Xiuying Wei, Yunchen Zhang, Yuhang Li, Xiangguo Zhang, Ruihao Gong, Jinyang Guo, and Xianglong Liu. 2023.
\newblock \href {https://aclanthology.org/2023.emnlp-main.102} {Outlier suppression+: Accurate quantization of large language models by equivalent and effective shifting and scaling}.
\newblock In \emph{Proceedings of the 2023 Conference on Empirical Methods in Natural Language Processing, {EMNLP} 2023, Singapore, December 6-10, 2023}, pages 1648--1665. Association for Computational Linguistics.

\bibitem[{Wolf et~al.(2020)Wolf, Debut, Sanh, Chaumond, Delangue, Moi, Cistac, Rault, Louf, Funtowicz, Davison, Shleifer, von Platen, Ma, Jernite, Plu, Xu, Scao, Gugger, Drame, Lhoest, and Rush}]{DBLP:conf/emnlp/WolfDSCDMCRLFDS20}
Thomas Wolf, Lysandre Debut, Victor Sanh, Julien Chaumond, Clement Delangue, Anthony Moi, Pierric Cistac, Tim Rault, R{\'{e}}mi Louf, Morgan Funtowicz, Joe Davison, Sam Shleifer, Patrick von Platen, Clara Ma, Yacine Jernite, Julien Plu, Canwen Xu, Teven~Le Scao, Sylvain Gugger, Mariama Drame, Quentin Lhoest, and Alexander~M. Rush. 2020.
\newblock \href {https://doi.org/10.18653/V1/2020.EMNLP-DEMOS.6} {Transformers: State-of-the-art natural language processing}.
\newblock In \emph{Proceedings of the 2020 Conference on Empirical Methods in Natural Language Processing: System Demonstrations, {EMNLP} 2020 - Demos, Online, November 16-20, 2020}, pages 38--45. Association for Computational Linguistics.

\bibitem[{Xia et~al.(2023)Xia, Artetxe, Zhou, Lin, Pasunuru, Chen, Zettlemoyer, and Stoyanov}]{DBLP:conf/acl/XiaAZLPCZS23}
Mengzhou Xia, Mikel Artetxe, Chunting Zhou, Xi~Victoria Lin, Ramakanth Pasunuru, Danqi Chen, Luke Zettlemoyer, and Veselin Stoyanov. 2023.
\newblock \href {https://doi.org/10.18653/V1/2023.ACL-LONG.767} {Training trajectories of language models across scales}.
\newblock In \emph{Proceedings of the 61st Annual Meeting of the Association for Computational Linguistics (Volume 1: Long Papers), {ACL} 2023, Toronto, Canada, July 9-14, 2023}, pages 13711--13738. Association for Computational Linguistics.

\bibitem[{Xiao et~al.(2023)Xiao, Lin, Seznec, Wu, Demouth, and Han}]{DBLP:conf/icml/XiaoLSWDH23}
Guangxuan Xiao, Ji~Lin, Micka{\"{e}}l Seznec, Hao Wu, Julien Demouth, and Song Han. 2023.
\newblock \href {https://proceedings.mlr.press/v202/xiao23c.html} {Smoothquant: Accurate and efficient post-training quantization for large language models}.
\newblock In \emph{International Conference on Machine Learning, {ICML} 2023, 23-29 July 2023, Honolulu, Hawaii, {USA}}, volume 202 of \emph{Proceedings of Machine Learning Research}, pages 38087--38099. {PMLR}.

\bibitem[{Yang et~al.(2023)Yang, Chiang, Zheng, Gonzalez, and Stoica}]{DBLP:journals/corr/abs-2311-04850}
Shuo Yang, Wei{-}Lin Chiang, Lianmin Zheng, Joseph~E. Gonzalez, and Ion Stoica. 2023.
\newblock \href {https://doi.org/10.48550/ARXIV.2311.04850} {Rethinking benchmark and contamination for language models with rephrased samples}.
\newblock \emph{CoRR}, abs/2311.04850.

\bibitem[{Yao et~al.(2022)Yao, Aminabadi, Zhang, Wu, Li, and He}]{DBLP:conf/nips/YaoAZWLH22}
Zhewei Yao, Reza~Yazdani Aminabadi, Minjia Zhang, Xiaoxia Wu, Conglong Li, and Yuxiong He. 2022.
\newblock \href {http://papers.nips.cc/paper\_files/paper/2022/hash/adf7fa39d65e2983d724ff7da57f00ac-Abstract-Conference.html} {Zeroquant: Efficient and affordable post-training quantization for large-scale transformers}.
\newblock In \emph{Advances in Neural Information Processing Systems 35: Annual Conference on Neural Information Processing Systems 2022, NeurIPS 2022, New Orleans, LA, USA, November 28 - December 9, 2022}.

\bibitem[{Yin et~al.(2023)Yin, Sun, Guo, Wu, Qiu, and Huang}]{DBLP:conf/acl/YinSGWQH23}
Zhangyue Yin, Qiushi Sun, Qipeng Guo, Jiawen Wu, Xipeng Qiu, and Xuanjing Huang. 2023.
\newblock \href {https://doi.org/10.18653/V1/2023.FINDINGS-ACL.551} {Do large language models know what they don't know?}
\newblock In \emph{Findings of the Association for Computational Linguistics: {ACL} 2023, Toronto, Canada, July 9-14, 2023}, pages 8653--8665. Association for Computational Linguistics.

\bibitem[{Yu et~al.(2024)Yu, Liu, and Xiong}]{DBLP:conf/coling/Yu0X24}
Linhao Yu, Qun Liu, and Deyi Xiong. 2024.
\newblock \href {https://aclanthology.org/2024.lrec-main.915} {{LFED:} {A} literary fiction evaluation dataset for large language models}.
\newblock In \emph{Proceedings of the 2024 Joint International Conference on Computational Linguistics, Language Resources and Evaluation, {LREC/COLING} 2024, 20-25 May, 2024, Torino, Italy}, pages 10466--10475. {ELRA} and {ICCL}.

\bibitem[{Zeng(2023)}]{DBLP:journals/corr/abs-2304-12986}
Hui Zeng. 2023.
\newblock \href {https://doi.org/10.48550/ARXIV.2304.12986} {Measuring massive multitask chinese understanding}.
\newblock \emph{CoRR}, abs/2304.12986.

\bibitem[{Zhao et~al.(2023)Zhao, Zhou, Li, Tang, Wang, Hou, Min, Zhang, Zhang, Dong, Du, Yang, Chen, Chen, Jiang, Ren, Li, Tang, Liu, Liu, Nie, and Wen}]{DBLP:journals/corr/abs-2303-18223}
Wayne~Xin Zhao, Kun Zhou, Junyi Li, Tianyi Tang, Xiaolei Wang, Yupeng Hou, Yingqian Min, Beichen Zhang, Junjie Zhang, Zican Dong, Yifan Du, Chen Yang, Yushuo Chen, Zhipeng Chen, Jinhao Jiang, Ruiyang Ren, Yifan Li, Xinyu Tang, Zikang Liu, Peiyu Liu, Jian{-}Yun Nie, and Ji{-}Rong Wen. 2023.
\newblock \href {https://doi.org/10.48550/ARXIV.2303.18223} {A survey of large language models}.
\newblock \emph{CoRR}, abs/2303.18223.

\bibitem[{Zhou et~al.(2023)Zhou, Lu, Mishra, Brahma, Basu, Luan, Zhou, and Hou}]{DBLP:journals/corr/abs-2311-07911}
Jeffrey Zhou, Tianjian Lu, Swaroop Mishra, Siddhartha Brahma, Sujoy Basu, Yi~Luan, Denny Zhou, and Le~Hou. 2023.
\newblock \href {https://doi.org/10.48550/ARXIV.2311.07911} {Instruction-following evaluation for large language models}.
\newblock \emph{CoRR}, abs/2311.07911.

\bibitem[{Zhu et~al.(2024)Zhu, Cui, and Xiong}]{DBLP:conf/coling/ZhuCX24}
Shaolin Zhu, Menglong Cui, and Deyi Xiong. 2024.
\newblock \href {https://aclanthology.org/2024.lrec-main.1444} {Towards robust in-context learning for machine translation with large language models}.
\newblock In \emph{Proceedings of the 2024 Joint International Conference on Computational Linguistics, Language Resources and Evaluation, {LREC/COLING} 2024, 20-25 May, 2024, Torino, Italy}, pages 16619--16629. {ELRA} and {ICCL}.

\bibitem[{Zhu et~al.(2023)Zhu, Li, Liu, Ma, and Wang}]{DBLP:journals/corr/abs-2308-07633}
Xunyu Zhu, Jian Li, Yong Liu, Can Ma, and Weiping Wang. 2023.
\newblock \href {https://doi.org/10.48550/ARXIV.2308.07633} {A survey on model compression for large language models}.
\newblock \emph{CoRR}, abs/2308.07633.

\bibitem[{Ziyu et~al.(2023)Ziyu, Qiguang, Longxuan, Mingda, Yi, Yushan, Haopeng, Weinan, and Liu}]{ziyu-etal-2023-lens}
Zhuang Ziyu, Chen Qiguang, Ma~Longxuan, Li~Mingda, Han Yi, Qian Yushan, Bai Haopeng, Zhang Weinan, and Ting Liu. 2023.
\newblock \href {https://aclanthology.org/2023.ccl-2.8} {Through the lens of core competency: Survey on evaluation of large language models}.
\newblock In \emph{Proceedings of the 22nd Chinese National Conference on Computational Linguistics (Volume 2: Frontier Forum)}, pages 88--109, Harbin, China. Chinese Information Processing Society of China.

\bibitem[{Zou et~al.(2023)Zou, Phan, Chen, Campbell, Guo, Ren, Pan, Yin, Mazeika, Dombrowski, Goel, Li, Byun, Wang, Mallen, Basart, Koyejo, Song, Fredrikson, Kolter, and Hendrycks}]{DBLP:journals/corr/abs-2310-01405}
Andy Zou, Long Phan, Sarah Chen, James Campbell, Phillip Guo, Richard Ren, Alexander Pan, Xuwang Yin, Mantas Mazeika, Ann{-}Kathrin Dombrowski, Shashwat Goel, Nathaniel Li, Michael~J. Byun, Zifan Wang, Alex Mallen, Steven Basart, Sanmi Koyejo, Dawn Song, Matt Fredrikson, J.~Zico Kolter, and Dan Hendrycks. 2023.
\newblock \href {https://doi.org/10.48550/ARXIV.2310.01405} {Representation engineering: {A} top-down approach to {AI} transparency}.
\newblock \emph{CoRR}, abs/2310.01405.

\end{thebibliography}

\clearpage
\appendix

\section{Quantization Strategies}\label{appendix_sec:quantization_strategies}

\paragraph{LLM.int8()} \citet{DBLP:conf/nips/DettmersLBZ22} is the earliest proposed method among the methods we use. It was implemented in bitsandbytes\footnote{\url{https://github.com/TimDettmers/bitsandbytes}} and deeply integrated with Huggingface Transformers. LLM.int8() proposes a vector-wise quantization approach, and stores the outlier submatrices in FP16 format while regular submatrices are in int8. In the matrix multiplication operation, the FP16 submatrix and the int8 submatrix are computed separately. This protects the outlier value, but the inference speed will decrease. 

\paragraph{GPTQ} \citet{DBLP:conf/iclr/FrantarAHA23} is a popular quantization method. Due to the outstanding contribution of the third-party library AutoGPTQ,\footnote{\url{https://github.com/AutoGPTQ/AutoGPTQ}} which provides CUDA implementation of quantization operators, it can also be easily applied to the model. GPTQ quantizes a weight matrix column by column and uses the Hessian matrix to adjust the unquantized parts of a weight matrix to minimize the loss caused by quantizing some parameters. 

\paragraph{SpQR} \citet{DBLP:journals/corr/abs-2306-03078} cleverly combines GPTQ \citep{DBLP:conf/iclr/FrantarAHA23} and outlier value protection to further improve quantization performance. It uses a smaller group size and saves outliers through a sparse matrix. Currently, this method has not yet implemented the CUDA operator. So we need to use floating point numbers to simulate the integer quantization, which is called fake quantization. The code used in our experiments was modified from the official code\footnote{\url{https://github.com/Vahe1994/SpQR}} and adapted to Qwen.

\section{Benchmarks}\label{appendix_sec:benchmarks}

\paragraph{MMLU} \citet{DBLP:conf/iclr/HendrycksBBZMSS21} serves as a comprehensive benchmark to measure the knowledge acquired by LLMs during their pretraining phase through zero- and few-shot learning. It encompasses 57 disciplines that cover diverse areas including STEM, humanities, social sciences, law, and ethics. These disciplines collectively evaluate the breadth and depth of a model's understanding across numerous academic and professional domains.

\paragraph{C-EVAL} \citet{huang2023ceval} is a comprehensive Chinese evaluation suite specifically tailored to assess the advanced knowledge and reasoning capabilities of LLMs within the Chinese context. Analogous to MMLU \citep{DBLP:conf/iclr/HendrycksBBZMSS21}, it comprises 52 disciplines, ranging from humanities to science and engineering, categorized within four difficulty levels: middle school, high school, college, and professional.

\paragraph{FLORES-200} \citet{DBLP:journals/corr/abs-2207-04672} is a high-quality benchmark for machine translation that encompasses 204 languages, doubling the language coverage of its predecessor, FLORES-101 \citep{DBLP:journals/tacl/GoyalGCCWJKRGF22}. Every sentence in each language has been translated into the others by professional translators. This unique feature establishes FLORES-200 as a many-to-many translation benchmark. Consequently, it is particularly well-suited for the evaluation of translation directions in which both the source and target languages are involved in the FLORES-200 benchmark.

\paragraph{CNN/DailyMail} \citet{DBLP:conf/conll/NallapatiZSGX16,DBLP:conf/acl/SeeLM17} is a valuable resource for abstractive multi-sentence summarization. It is derived from a previous dataset created by \citet{DBLP:conf/nips/HermannKGEKSB15} for passage-based question-answering, using human-generated abstractive summary bullets from news stories on the CNN and Daily Mail websites. These summaries are originally used as questions with a masked entity, paired with corresponding passages from which systems are expected to generate answers. CNN/DailyMail is constructed by restoring all the original summary bullets for each story, treating them as separate sentences to form coherent, multi-sentence summaries. CNN/DailyMail consists of a large number of instances, including 286,817 training instances, 13,368 validation instances, and 11,487 test instances. The test instances are utilized in our evaluation experiments.

\paragraph{XSum} \citet{DBLP:conf/emnlp/NarayanCL18} is a fundamental resource for the development and assessment of abstractive single-document summarization systems. It is derived from online articles sourced from the British Broadcasting Corporation (BBC), which typically include professionally written introductory sentences serving as concise one-sentence summaries that encapsulate the essence of the entire article. XSum covers a wide range of domains, including news, politics, sports, weather, and more. Notably, the documents and summaries in XSum are shorter compared to CNN/DailyMail. Furthermore, the summaries in XSum are significantly more abstractive, as evidenced by a notable percentage of novel n-grams that are not present within the source documents. The dataset has been randomly partitioned into training (90\%), validation (5\%), and test (5\%) splits. The evaluation experiments in our work are conducted using the test set.

\paragraph{GSM8K} \citet{DBLP:journals/corr/abs-2110-14168} is a collection of 8,500 high-quality grade school math word problems designed to evaluate the multi-step mathematical reasoning abilities of LLMs. The dataset has been meticulously curated to ensure high linguistic diversity. The problems included in GSM8K only involve relatively simple math concepts that a bright middle school student can solve using basic arithmetic operations such as addition, subtraction, multiplication, and division over a sequence of 2 to 8 steps.

\paragraph{SNLI} \citet{DBLP:conf/emnlp/BowmanAPM15} is a large-scale, human-annotated collection of sentence pairs specifically designed for training and evaluating machine learning models on the task of natural language inference (NLI). All sentences in SNLI are written by human contributors within a grounded context based on image captioning, ensuring that they reflect naturalistic language use rather than being algorithmically generated. Each sentence pair within the dataset is labeled as either an entailment, a contradiction, or neutral. SNLI has been partitioned into training, development, and test splits. Both the development and test splits encompass 10,000 examples each. The test split, in particular, is utilized in our evaluation experiments.

\paragraph{FollowBench} \citet{DBLP:journals/corr/abs-2310-20410} is a comprehensive benchmark that focuses on evaluating the instruction-following capabilities of LLMs through a variety of fine-grained constraints. It encompasses five distinct fine-grained constraints: content, situation, style, format, and example. This benchmark is specifically designed to address the limitations of existing evaluation benchmarks, which primarily assess the quality of responses without measuring their adherence to specific instruction constraints. FollowBench is available in two language splits, English and Chinese, with the English split used in our evaluation experiments.

\paragraph{TruthfulQA} \citet{DBLP:conf/acl/LinHE22} is a benchmark designed to assess the truthfulness of LLMs. It is composed of 817 questions across 38 categories, including health, law, finance, and politics. These questions are crafted in such a way that they can elicit false answers based on common misconceptions or false beliefs that some humans might also give. TruthfulQA incorporates two distinct tasks, namely, generation and multiple-choice. Both tasks utilize the same sets of questions and reference answers, thereby ensuring consistency in evaluation. Following \citet{DBLP:journals/corr/abs-2310-01405}, we assess models on the multiple-choice task.

\paragraph{BBQ} \citet{DBLP:conf/acl/ParrishCNPPTHB22} is a benchmark for evaluating the degree of social biases present in LLMs, specifically about question-answering tasks. It assesses biases towards protected groups across nine social dimensions that are particularly relevant in US English-speaking contexts. This benchmark includes a variety of question sets, including ambiguous contexts where the answer is not clear, and disambiguated ones where a correct response can be determined with great certainty. Each example within the dataset comprises clusters of four multiple-choice questions, encompassing both negative and non-negative variants, and is presented with or without a disambiguating context. Negative questions aim to test stereotypes that reflect societal prejudices, while non-negative questions complement this by assessing whether model responses show a bias towards particular labels.

\section{Prompts}\label{appendix_sec:prompts}

\begin{figure*}[!t]
\centering
\includegraphics[width=\linewidth]{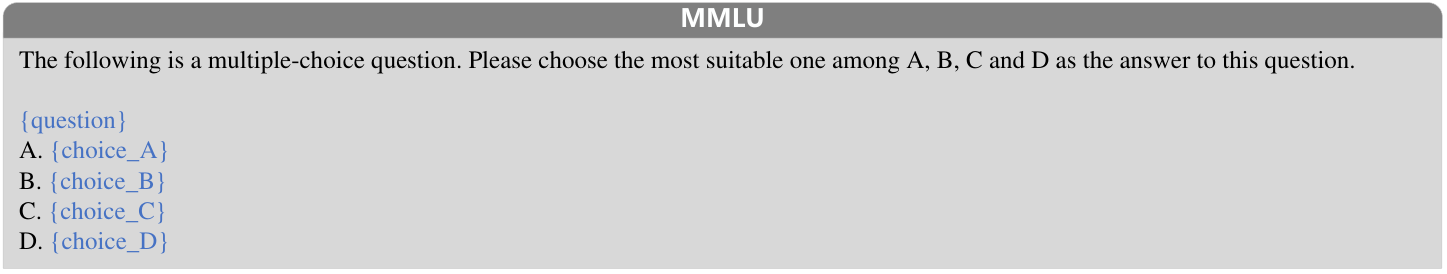} 
\caption{Prompt used for MMLU \citep{DBLP:conf/iclr/HendrycksBBZMSS21} benchmark.}
\label{prompts_mmlu}
\end{figure*}

\begin{figure*}[!t]
\centering
\includegraphics[width=\linewidth]{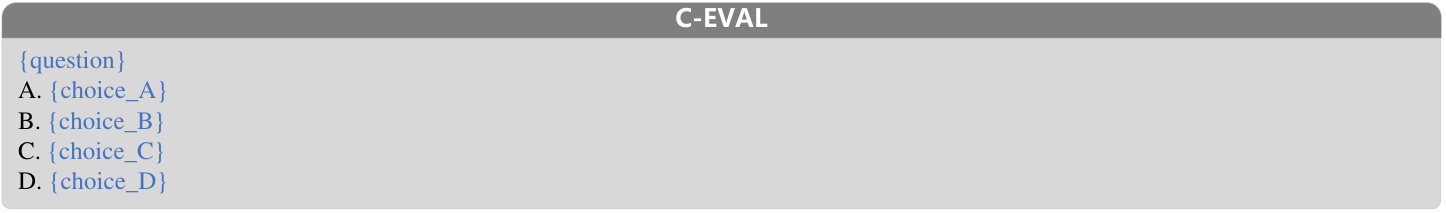} 
\caption{Prompt used for C-EVAL \citep{huang2023ceval} benchmark.}
\label{prompt_ceval}
\end{figure*}

\begin{figure*}[!t]
\centering
\includegraphics[width=\linewidth]{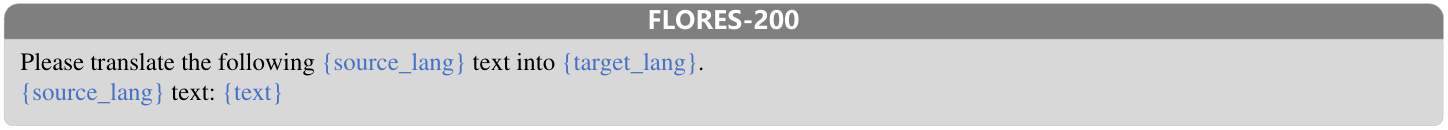} 
\caption{Prompt used for FLORES-200 \citep{DBLP:journals/corr/abs-2207-04672} benchmark.}
\label{prompt_flores200}
\end{figure*}

\begin{figure*}[!t]
\centering
\includegraphics[width=\linewidth]{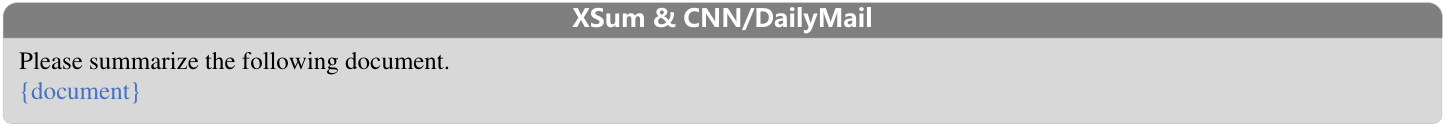} 
\caption{Prompt used for XSum \citep{DBLP:conf/emnlp/NarayanCL18} and CNN/DailyMail \citep{DBLP:conf/acl/SeeLM17} benchmarks.}
\label{prompt_summarization}
\end{figure*}

\begin{figure*}[!t]
\centering
\includegraphics[width=\linewidth]{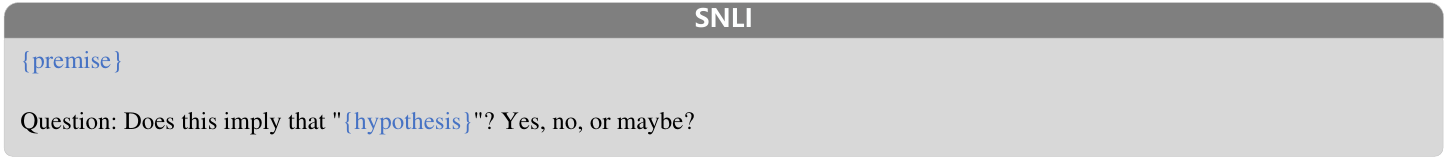} 
\caption{Prompt used for SNLI \citep{DBLP:conf/emnlp/BowmanAPM15} benchmark.}
\label{prompt_snli}
\end{figure*}

\begin{figure*}[!t]
\centering
\includegraphics[width=\linewidth]{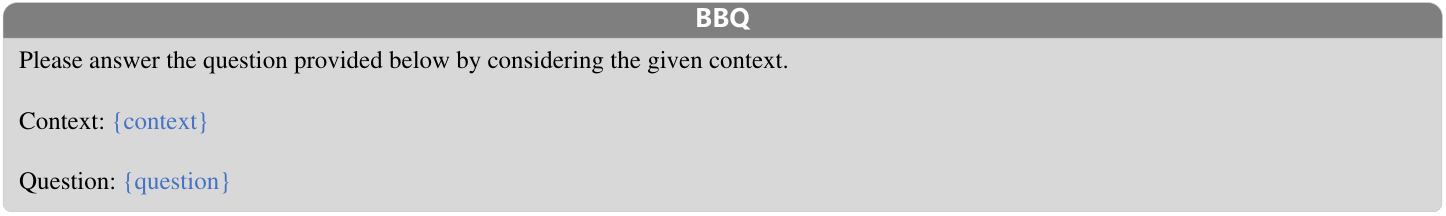} 
\caption{Prompt used for BBQ \citep{DBLP:conf/acl/ParrishCNPPTHB22} benchmark.}
\label{prompt_bbq}
\end{figure*}

The prompts employed in our evaluation experiments across various benchmarks are illustrated in Figures~\ref{prompts_mmlu} to~\ref{prompt_bbq}. Notably, for the GSM8K and TruthfulQA benchmarks, the questions are used directly as input for the LLMs. Furthermore, for the FollowBench benchmark, we utilized the official implementation, resulting in prompts that are consistent with those described in \citet{DBLP:journals/corr/abs-2310-20410}.

\section{Detailed Experimental Results}\label{appendix_sec:detailed_experimental_results}
\begin{figure*}[!t]
\centering
\begin{subfigure}{0.325\textwidth}
    \centering
    \includegraphics[width=\linewidth]{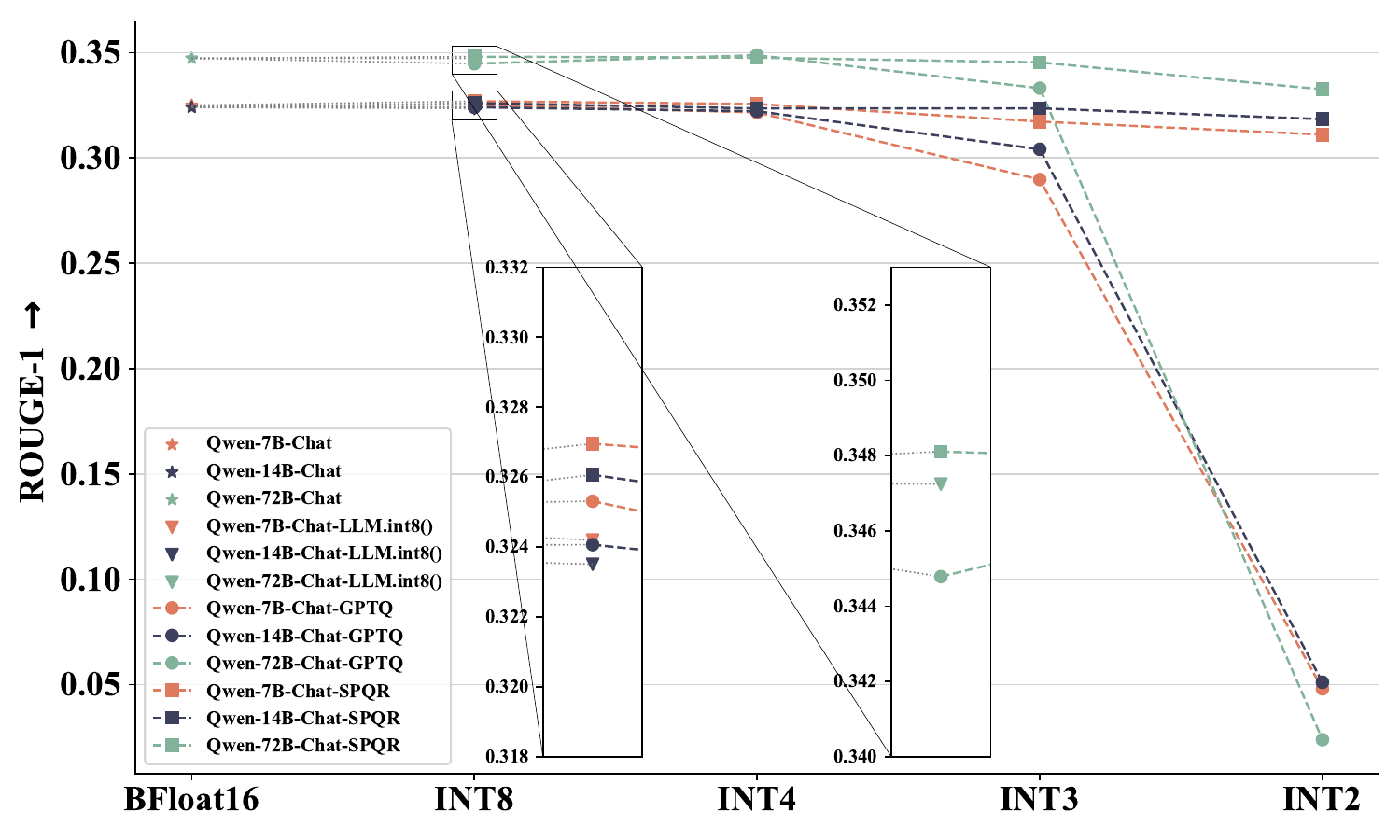}
    \caption{ROUGE-1}
\end{subfigure}
\begin{subfigure}{0.325\textwidth}
    \centering
    \includegraphics[width=\linewidth]{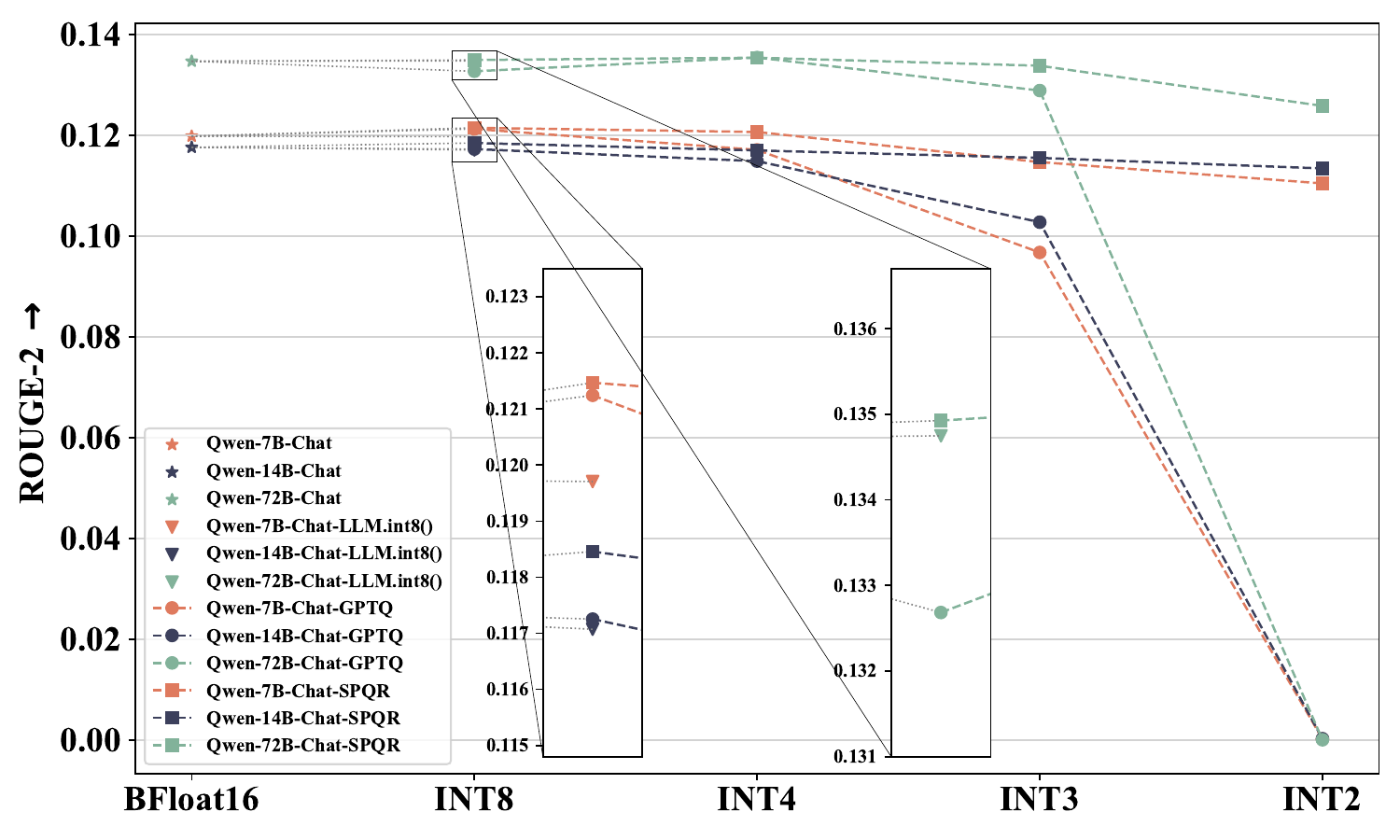}
    \caption{ROUGE-2}
\end{subfigure}
\begin{subfigure}{0.325\textwidth}
    \centering
    \includegraphics[width=\linewidth]{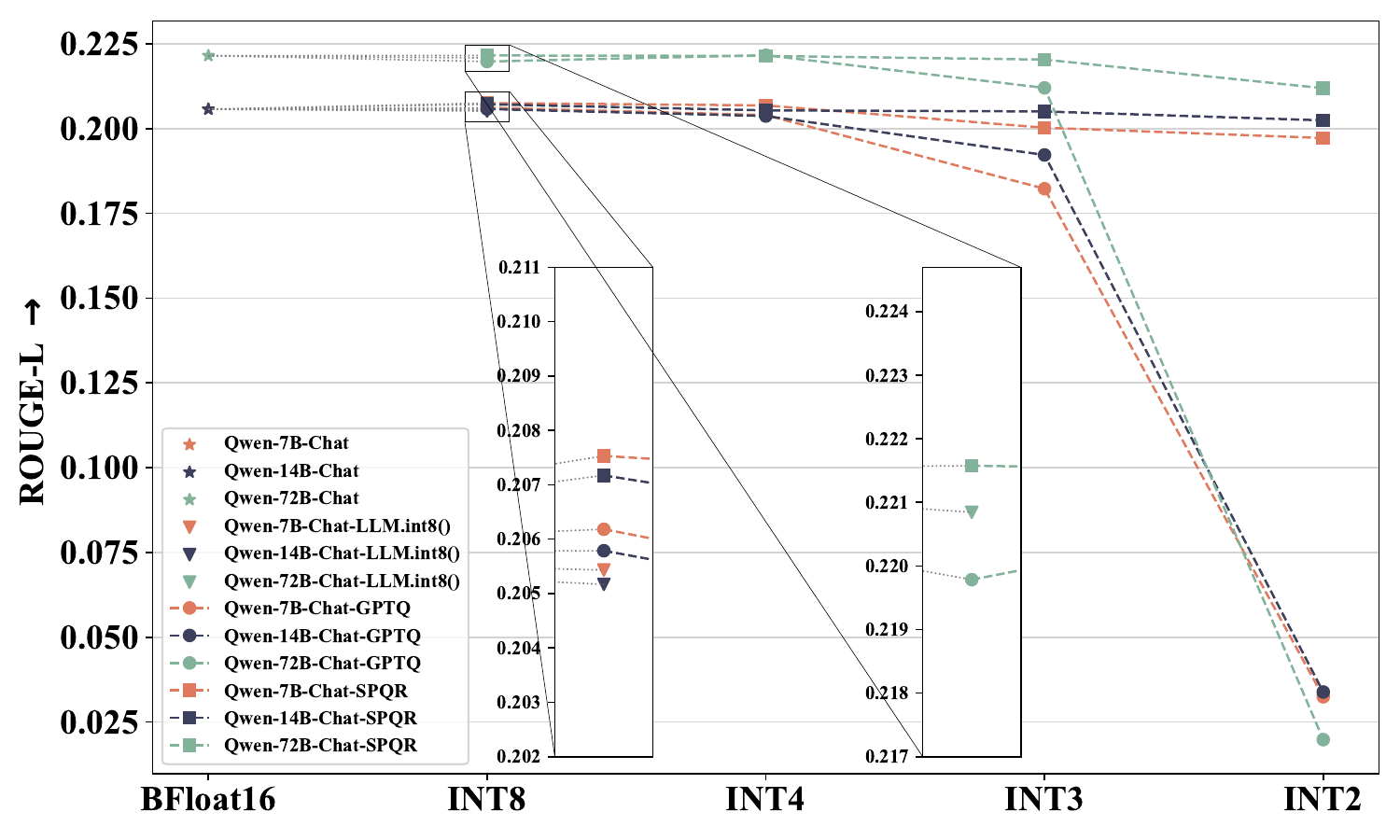}
    \caption{ROUGE-L}
\end{subfigure}
\caption{ROUGE-1 (a), ROUGE-2 (b), and ROUGE-L (c) scores for the Qwen-Chat series models and their quantized counterparts on the test sets of CNN/DailyMail \citep{DBLP:conf/acl/SeeLM17}.}
\label{figure:cnndm}
\end{figure*}

\begin{figure*}[!t]
\centering
\begin{subfigure}{0.475\textwidth}
    \centering
    \includegraphics[width=\linewidth]{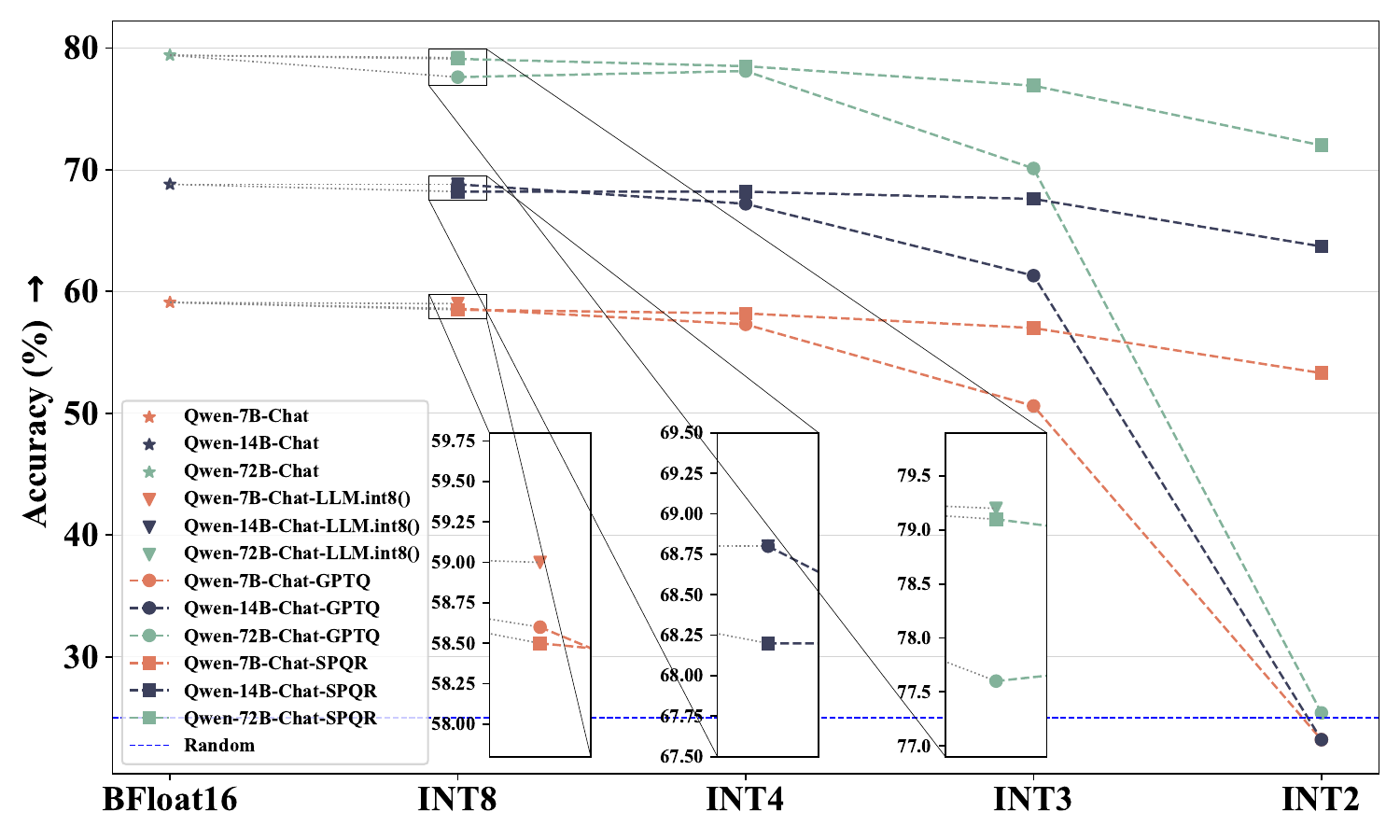}
\caption{C-EVAL}
\label{figure:ceval_avg}
\end{subfigure}
\begin{subfigure}{0.475\textwidth}
    \centering
    \includegraphics[width=\linewidth]{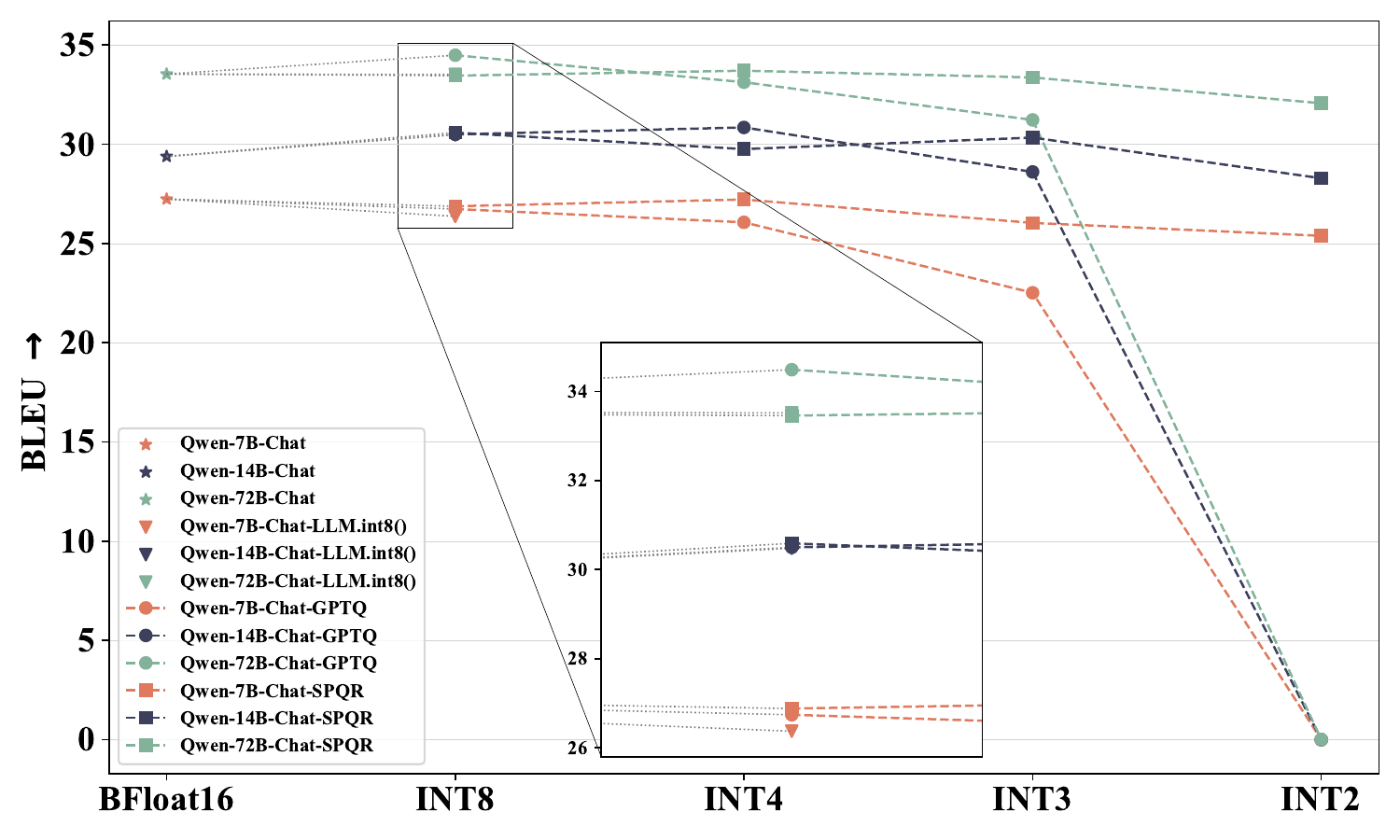}
\caption{Zh $\rightarrow$ En}
\label{figure:flores_200_zh2en}
\end{subfigure}
\caption{Performance of the Qwen-Chat series models and their quantized counterparts on the C-EVAL \citep{huang2023ceval} benchmark (a) and Chinese-to-English (Zh $\rightarrow$ En) translation task of the FLORES-200 \citep{DBLP:journals/corr/abs-2207-04672} (b) benchmark.}
\label{figure:ceval_avg_flores_200_zh2en}
\end{figure*}

\begin{figure*}[!t]
\centering
\begin{subfigure}{0.475\textwidth}
    \centering
    \includegraphics[width=\linewidth]{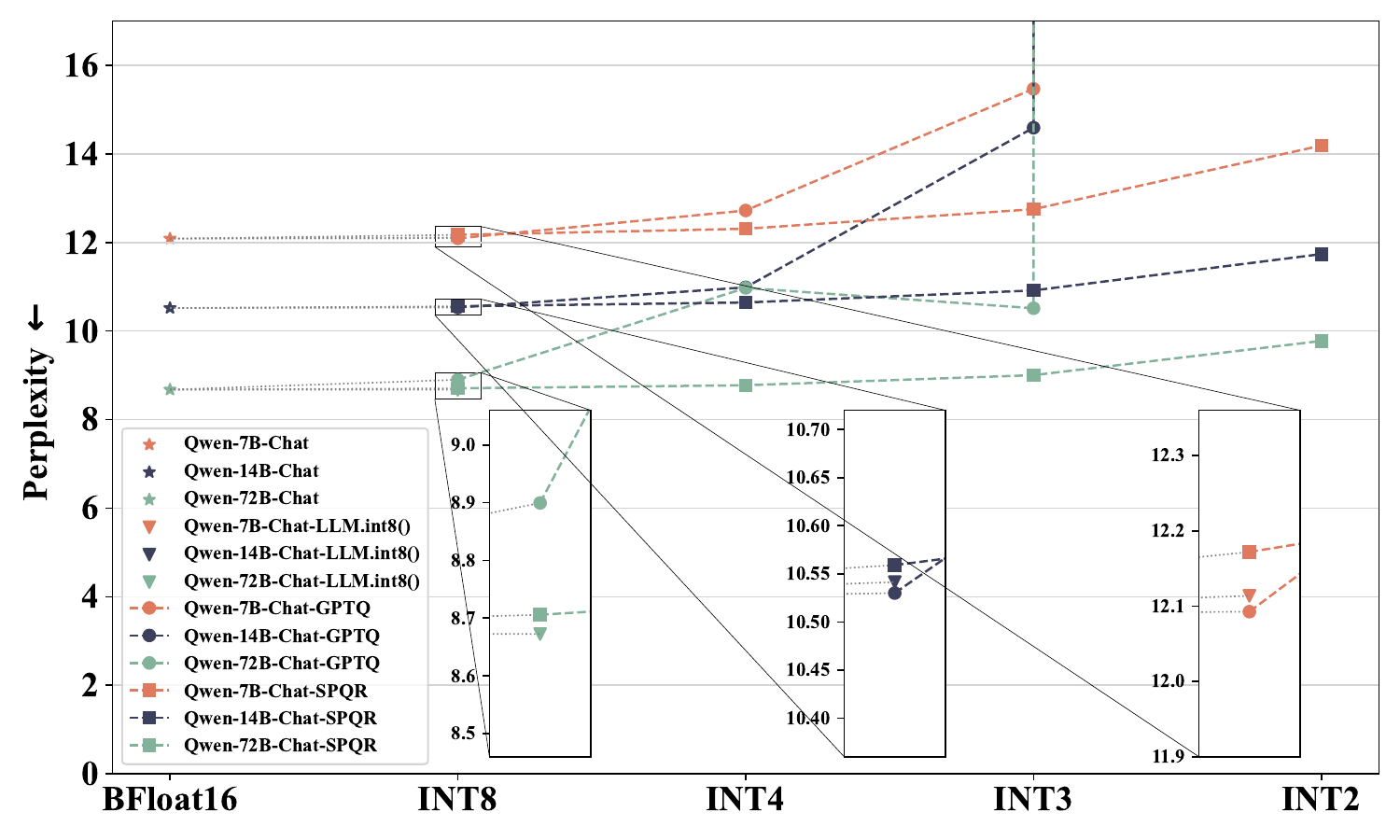}
    \caption{C4}
\end{subfigure}
\begin{subfigure}{0.475\textwidth}
    \centering
    \includegraphics[width=\linewidth]{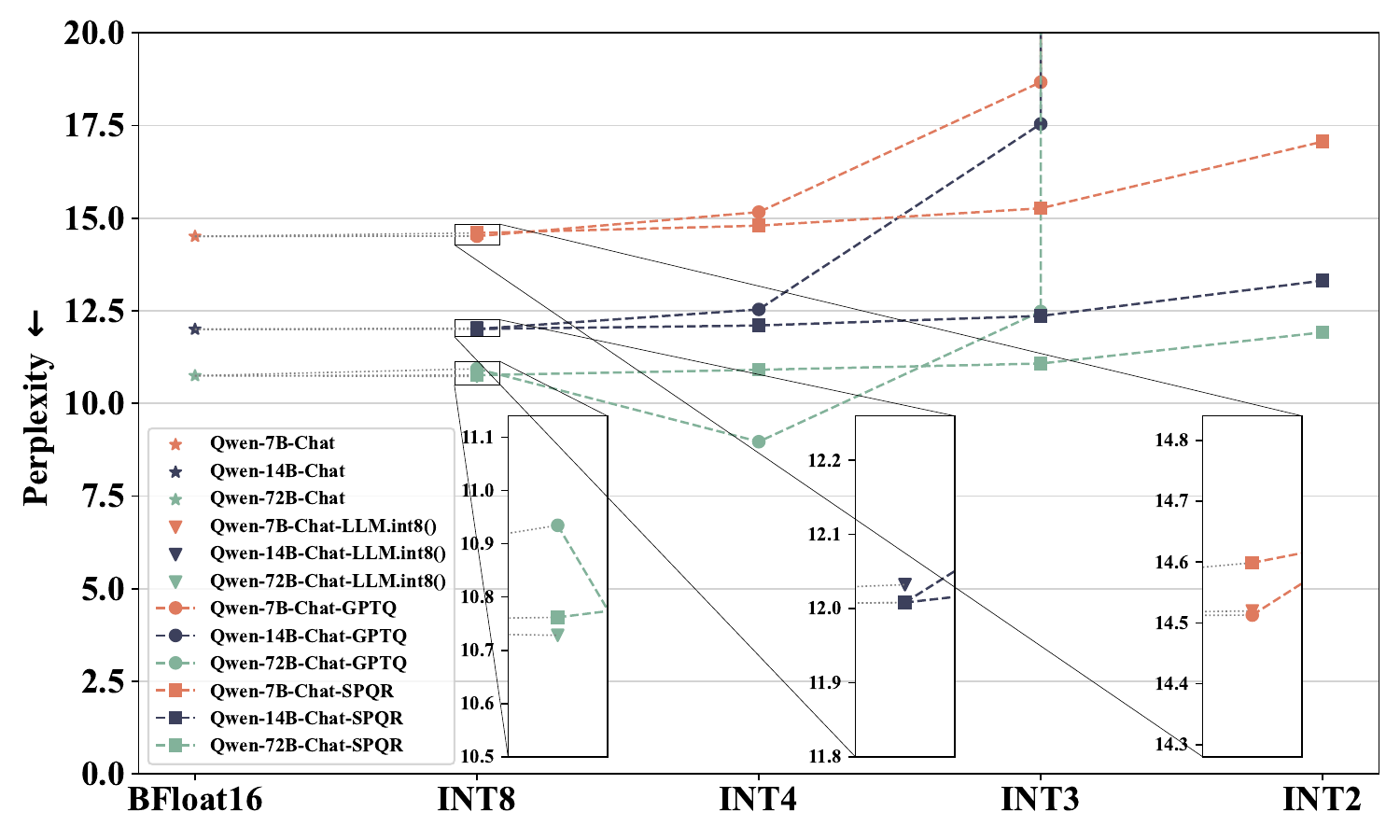}
    \caption{PTB}
\end{subfigure}
\caption{Perplexity of Qwen-Chat Series models and their quantized counterparts on the C4 \citep{DBLP:journals/jmlr/RaffelSRLNMZLL20} and PTB datasets \citep{DBLP:conf/naacl/MarcusKMMBFKS94}.}
\label{figure:ppl_c4_ptb}
\end{figure*}

The performance of the Qwen-Chat series models, along with their quantized counterparts, is depicted in the following figures: CNN/DailyMail test sets \citep{DBLP:conf/acl/SeeLM17} (Figure~\ref{figure:cnndm}), C-EVAL benchmark \citep{huang2023ceval} (Figure~\ref{figure:ceval_avg}), Chinese to English translation on the FLORES-200 benchmark \citep{DBLP:journals/corr/abs-2207-04672} (Figure~\ref{figure:flores_200_zh2en}), and perplexity on the C4 \citep{DBLP:journals/jmlr/RaffelSRLNMZLL20} and PTB \citep{DBLP:conf/naacl/MarcusKMMBFKS94} datasets (Figure~\ref{figure:ppl_c4_ptb}). Detailed experimental results for all evaluated benchmarks, as well as data on memory consumption and decoding speed during inference, are provided in Table~\ref{tab:flores_200} through~\ref{tab:bbq}. In these tables, the best results achieved by the quantized models are highlighted in bold, while underlined results indicate that the performance of the quantized model surpasses that of the BFloat16 baseline.

\begin{table*}[!ht]
\centering
% \resizebox{\textwidth}{!}{
\begin{tabular}{ccccc}
    \toprule
    \textbf{Model} & \textbf{Datatype} & \textbf{Quantization Method} & \textbf{BLEU (En $\rightarrow$ Zh)} & \textbf{BLEU (Zh $\rightarrow$ En)} \\
    \midrule
    \multicolumn{1}{c}{\multirow{10}{*}{Qwen-7B-Chat}} & BFloat16 & - & 32.02 &  27.24 \\
    \cmidrule{2-5}
    \multicolumn{1}{c}{} & \multicolumn{1}{c}{\multirow{3}{*}{INT-8}} & LLM.int8() & 31.56 & 26.37$^\star$ \\
    \multicolumn{1}{c}{} & \multicolumn{1}{c}{} & GPTQ & {\ul \textbf{32.30}} & 26.74 \\
    \multicolumn{1}{c}{} & \multicolumn{1}{c}{} & SpQR & {\ul 32.13} & \textbf{26.88} \\
    \cmidrule{2-5}
    \multicolumn{1}{c}{} & \multicolumn{1}{c}{\multirow{2}{*}{INT-4}} & GPTQ & 30.79$^\star$ & 26.07$^\star$ \\
    \multicolumn{1}{c}{} & \multicolumn{1}{c}{} & SpQR & \textbf{31.96} & \textbf{27.22} \\
    \cmidrule{2-5}
    \multicolumn{1}{c}{} & \multicolumn{1}{c}{\multirow{2}{*}{INT-3}} & GPTQ & 25.91$^\star$ & 22.52$^\star$ \\
    \multicolumn{1}{c}{} & \multicolumn{1}{c}{} & SpQR & \textbf{30.73}$^\star$ & \textbf{26.04}$^\star$ \\
    \cmidrule{2-5}
    \multicolumn{1}{c}{} & \multicolumn{1}{c}{\multirow{2}{*}{INT-2}} & GPTQ & 0.01$^\star$ & 0.00$^\star$ \\
    \multicolumn{1}{c}{} & \multicolumn{1}{c}{} & SpQR & \textbf{29.04}$^\star$ & \textbf{25.39}$^\star$ \\
    \midrule
    \multicolumn{1}{c}{\multirow{10}{*}{Qwen-14B-Chat}} & BFloat16 & - & 32.87 & 29.39 \\
    \cmidrule{2-5}
    \multicolumn{1}{c}{} & \multicolumn{1}{c}{\multirow{3}{*}{INT-8}} & LLM.int8() & 32.78 & {\ul 30.48} \\
    \multicolumn{1}{c}{} & \multicolumn{1}{c}{} & GPTQ & 33.18 & {\ul 30.50} \\
    \multicolumn{1}{c}{} & \multicolumn{1}{c}{} & SpQR & {\ul \textbf{33.35}} & {\ul \textbf{30.59}} \\
    \cmidrule{2-5}
    \multicolumn{1}{c}{} & \multicolumn{1}{c}{\multirow{2}{*}{INT-4}} & GPTQ & 31.95$^\star$ & {\ul \textbf{30.85}}$^\star$ \\
    \multicolumn{1}{c}{} & \multicolumn{1}{c}{} & SpQR & {\ul \textbf{33.18}} & {\ul 29.76} \\
    \cmidrule{2-5}
    \multicolumn{1}{c}{} & \multicolumn{1}{c}{\multirow{2}{*}{INT-3}} & GPTQ & 29.23$^\star$ & 28.61 \\
    \multicolumn{1}{c}{} & \multicolumn{1}{c}{} & SpQR & \textbf{31.84}$^\star$ & {\ul \textbf{30.34}} \\
    \cmidrule{2-5}
    \multicolumn{1}{c}{} & \multicolumn{1}{c}{\multirow{2}{*}{INT-2}} & GPTQ & 0.01$^\star$ & 0.00$^\star$ \\
    \multicolumn{1}{c}{} & \multicolumn{1}{c}{} & SpQR & \textbf{30.10}$^\star$ & \textbf{28.29}$^\star$ \\
    \midrule
    \multicolumn{1}{c}{\multirow{10}{*}{Qwen-72B-Chat}} & BFloat16 & - & 36.08 & 33.54 \\
    \cmidrule{2-5}
    \multicolumn{1}{c}{} & \multicolumn{1}{c}{\multirow{3}{*}{INT-8}} & LLM.int8() & 35.26$^\star$ & 33.52 \\
    \multicolumn{1}{c}{} & \multicolumn{1}{c}{} & GPTQ & 35.15$^\star$ & {\ul \textbf{34.49}}$^\star$ \\
    \multicolumn{1}{c}{} & \multicolumn{1}{c}{} & SpQR & \textbf{35.87} & 33.46 \\
    \cmidrule{2-5}
    \multicolumn{1}{c}{} & \multicolumn{1}{c}{\multirow{2}{*}{INT-4}} & GPTQ & 34.93$^\star$ & 33.13 \\
    \multicolumn{1}{c}{} & \multicolumn{1}{c}{} & SpQR & \textbf{35.73} & {\ul \textbf{33.71}} \\
    \cmidrule{2-5}
    \multicolumn{1}{c}{} & \multicolumn{1}{c}{\multirow{2}{*}{INT-3}} & GPTQ & 30.73$^\star$ & 31.23$^\star$ \\
    \multicolumn{1}{c}{} & \multicolumn{1}{c}{} & SpQR & \textbf{34.79}$^\star$ & {\ul \textbf{33.37}}$^\star$ \\
    \cmidrule{2-5}
    \multicolumn{1}{c}{} & \multicolumn{1}{c}{\multirow{2}{*}{INT-2}} & GPTQ & 0.01$^\star$ & 0.01$^\star$ \\
    \multicolumn{1}{c}{} & \multicolumn{1}{c}{} & SpQR & \textbf{33.41}$^\star$ & \textbf{32.07}$^\star$ \\
    \bottomrule
\end{tabular}
% }
\caption{BLEU scores of Qwen-Chat series models and their quantized counterparts for English-to-Chinese and Chinese-to-English translation tasks of the FLORES-200 benchmark \citep{DBLP:journals/corr/abs-2207-04672}. The highest BLEU scores obtained by the quantized models are highlighted in bold. Underlined results denote instances where the quantized model outperforms the BFloat16 baseline. \textbf{Statistically significant differences between quantized LLMs and their non-quantized equivalents are indicated by $^\star p < 0.05$.}}
\label{tab:flores_200}
\end{table*}

\begin{table*}[!ht]
\centering
\resizebox{\textwidth}{!}{
\begin{tabular}{cccccc}
    \toprule
    \textbf{Model} & \textbf{Datatype} & \textbf{Quantization Method} & \textbf{Perplexity (WikiText2)} & \textbf{Perplexity (C4)} & \textbf{Perplexity (PTB)} \\
    \midrule
    \multicolumn{1}{c}{\multirow{10}{*}{Qwen-7B-Chat}} & BFloat16 & - & 8.67 & 12.09 & 14.51 \\
    \cmidrule{2-6}
    \multicolumn{1}{c}{} & \multicolumn{1}{c}{\multirow{3}{*}{INT-8}} & LLM.int8() & \textbf{8.68} & 12.11 & 14.52 \\
    \multicolumn{1}{c}{} & \multicolumn{1}{c}{} & GPTQ & \textbf{8.68} & \textbf{12.09} & \textbf{14.51} \\
    \multicolumn{1}{c}{} & \multicolumn{1}{c}{} & SpQR & 8.71 & 12.17 & 14.60 \\
    \cmidrule{2-6}
    \multicolumn{1}{c}{} & \multicolumn{1}{c}{\multirow{2}{*}{INT-4}} & GPTQ & 9.04 & 12.72 & 15.16 \\
    \multicolumn{1}{c}{} & \multicolumn{1}{c}{} & SpQR & \textbf{8.82} & \textbf{12.31} & \textbf{14.80} \\
    \cmidrule{2-6}
    \multicolumn{1}{c}{} & \multicolumn{1}{c}{\multirow{2}{*}{INT-3}} & GPTQ & 11.17 & 15.47 & 18.67 \\
    \multicolumn{1}{c}{} & \multicolumn{1}{c}{} & SpQR & \textbf{9.17} & \textbf{12.75} & \textbf{15.27} \\
    \cmidrule{2-6}
    \multicolumn{1}{c}{} & \multicolumn{1}{c}{\multirow{2}{*}{INT-2}} & GPTQ & 123030.34 & 41936.28 & 88223.58 \\
    \multicolumn{1}{c}{} & \multicolumn{1}{c}{} & SpQR & \textbf{10.05} & \textbf{14.19} & \textbf{17.07} \\
    \midrule
    \multicolumn{1}{c}{\multirow{10}{*}{Qwen-14B-Chat}} & BFloat16 & - & 6.99 & 10.52 & 12.00 \\
    \cmidrule{2-6}
    \multicolumn{1}{c}{} & \multicolumn{1}{c}{\multirow{3}{*}{INT-8}} & LLM.int8() & 7.00 & 10.54 & 12.03 \\
    \multicolumn{1}{c}{} & \multicolumn{1}{c}{} & GPTQ & 7.00 & \textbf{10.53} & \textbf{12.01} \\
    \multicolumn{1}{c}{} & \multicolumn{1}{c}{} & SpQR & \textbf{6.99} & 10.56 & \textbf{12.01} \\
    \cmidrule{2-6}
    \multicolumn{1}{c}{} & \multicolumn{1}{c}{\multirow{2}{*}{INT-4}} & GPTQ & 7.35 & 10.99 & 12.54 \\
    \multicolumn{1}{c}{} & \multicolumn{1}{c}{} & SpQR & \textbf{7.07} & \textbf{10.64} & \textbf{12.10} \\
    \cmidrule{2-6}
    \multicolumn{1}{c}{} & \multicolumn{1}{c}{\multirow{2}{*}{INT-3}} & GPTQ & 9.68 & 14.59 & 17.54 \\
    \multicolumn{1}{c}{} & \multicolumn{1}{c}{} & SpQR & \textbf{7.31} & \textbf{10.92} & \textbf{12.36} \\
    \cmidrule{2-6}
    \multicolumn{1}{c}{} & \multicolumn{1}{c}{\multirow{2}{*}{INT-2}} & GPTQ & 200643.66 & 153141.75 & 224832.00 \\
    \multicolumn{1}{c}{} & \multicolumn{1}{c}{} & SpQR & \textbf{7.94} & \textbf{11.74} & \textbf{13.31} \\
    \midrule
    \multicolumn{1}{c}{\multirow{10}{*}{Qwen-72B-Chat}} & BFloat16 & - & 6.15 & 8.68 & 10.75 \\
    \cmidrule{2-6}
    \multicolumn{1}{c}{} & \multicolumn{1}{c}{\multirow{3}{*}{INT-8}} & LLM.int8() & {\ul \textbf{6.14}} & {\ul \textbf{8.67}} & {\ul \textbf{10.73}} \\
    \multicolumn{1}{c}{} & \multicolumn{1}{c}{} & GPTQ & 6.28 & 8.90 & 10.93 \\
    \multicolumn{1}{c}{} & \multicolumn{1}{c}{} & SpQR & 6.16 & 8.71 & 10.76 \\
    \cmidrule{2-6}
    \multicolumn{1}{c}{} & \multicolumn{1}{c}{\multirow{2}{*}{INT-4}} & GPTQ & 6.37 & 8.97 & 10.97 \\
    \multicolumn{1}{c}{} & \multicolumn{1}{c}{} & SpQR & \textbf{6.23} & \textbf{8.77} & \textbf{10.91} \\
    \cmidrule{2-6}
    \multicolumn{1}{c}{} & \multicolumn{1}{c}{\multirow{2}{*}{INT-3}} & GPTQ & 7.58 & 10.51 & 12.48 \\
    \multicolumn{1}{c}{} & \multicolumn{1}{c}{} & SpQR & \textbf{6.43} & \textbf{9.00} & \textbf{11.08} \\
    \cmidrule{2-6}
    \multicolumn{1}{c}{} & \multicolumn{1}{c}{\multirow{2}{*}{INT-2}} & GPTQ & 52688.75 & 38123.23 & 55330.14 \\
    \multicolumn{1}{c}{} & \multicolumn{1}{c}{} & SpQR & \textbf{7.01} & \textbf{9.78} & \textbf{11.92} \\
    \bottomrule
\end{tabular}
}
\caption{The perplexity of the Qwen-Chat series models and their quantized counterparts on WikiText2 \citep{DBLP:conf/iclr/MerityX0S17}, C4 \citep{DBLP:journals/jmlr/RaffelSRLNMZLL20}, and PTB \citep{DBLP:conf/naacl/MarcusKMMBFKS94}.  The best results achieved by the quantized models are highlighted in bold, while underlined results indicate that the performance of the quantized model surpasses that of the BFloat16 baseline.}
\label{tab:ppl}
\end{table*}

\begin{table*}[!ht]
\centering
% \resizebox{\textwidth}{!}{
\begin{tabular}{ccccc}
\toprule
\textbf{Model} & \textbf{Datatype} & \textbf{Quantization Method} & \textbf{Memory} & \textbf{Speed} \\
\midrule
\multicolumn{1}{c}{\multirow{10}{*}{Qwen-7B-Chat}} & BFloat16 & - & 15.14 & 37.67 \\
\cmidrule{2-5}
\multicolumn{1}{c}{} & \multicolumn{1}{c}{\multirow{3}{*}{INT-8}} & LLM.int8 & \textbf{9.23} & 7.19 \\
\multicolumn{1}{c}{} & \multicolumn{1}{c}{} & GPTQ & 10.91 & 13.57 \\
\multicolumn{1}{c}{} & \multicolumn{1}{c}{} & SpQR & 15.60 & \textbf{37.65} \\
\cmidrule{2-5}
\multicolumn{1}{c}{} & \multicolumn{1}{c}{\multirow{2}{*}{INT-4}} & GPTQ & \textbf{7.83} & 37.43 \\
\multicolumn{1}{c}{} & \multicolumn{1}{c}{} & SpQR & 15.60 & \textbf{37.73} \\
\cmidrule{2-5}
\multicolumn{1}{c}{} & \multicolumn{1}{c}{\multirow{2}{*}{INT-3}} & GPTQ & \textbf{7.12} & 8.21 \\
\multicolumn{1}{c}{} & \multicolumn{1}{c}{} & SpQR & 15.61 & \textbf{37.73} \\
\cmidrule{2-5}
\multicolumn{1}{c}{} & \multicolumn{1}{c}{\multirow{2}{*}{INT-2}} & GPTQ & \textbf{6.26} & 19.36 \\
\multicolumn{1}{c}{} & \multicolumn{1}{c}{} & SpQR & 15.66 & \textbf{37.51} \\
\midrule
\multicolumn{1}{c}{\multirow{10}{*}{Qwen-14B-Chat}} & BFloat16 & - & 27.60 & 25.15 \\
\cmidrule{2-5}
\multicolumn{1}{c}{} & \multicolumn{1}{c}{\multirow{3}{*}{INT-8}} & LLM.int8 & \textbf{15.91} & 5.85 \\
\multicolumn{1}{c}{} & \multicolumn{1}{c}{} & GPTQ & 17.92 & 14.37 \\
\multicolumn{1}{c}{} & \multicolumn{1}{c}{} & SpQR & 27.95 & \textbf{25.42} \\
\cmidrule{2-5}
\multicolumn{1}{c}{} & \multicolumn{1}{c}{\multirow{2}{*}{INT-4}} & GPTQ & \textbf{12.03} & 24.38 \\
\multicolumn{1}{c}{} & \multicolumn{1}{c}{} & SpQR & 27.95 & \textbf{24.62} \\
\cmidrule{2-5}
\multicolumn{1}{c}{} & \multicolumn{1}{c}{\multirow{2}{*}{INT-3}} & GPTQ & \textbf{10.77} & 4.71 \\
\multicolumn{1}{c}{} & \multicolumn{1}{c}{} & SpQR & 27.97 & \textbf{25.20} \\
\cmidrule{2-5}
\multicolumn{1}{c}{} & \multicolumn{1}{c}{\multirow{2}{*}{INT-2}} & GPTQ & \textbf{8.99} & 18.26 \\
\multicolumn{1}{c}{} & \multicolumn{1}{c}{} & SpQR & 28.04 & \textbf{24.82} \\
\midrule
\multicolumn{1}{c}{\multirow{10}{*}{Qwen-72B-Chat}} & BFloat16 & - & 138.44 & 8.97 \\
\cmidrule{2-5}
\multicolumn{1}{c}{} & \multicolumn{1}{c}{\multirow{3}{*}{INT-8}} & LLM.int8 & \textbf{74.96} & 3.07 \\
\multicolumn{1}{c}{} & \multicolumn{1}{c}{} & GPTQ & 77.85 & 1.43 \\
\multicolumn{1}{c}{} & \multicolumn{1}{c}{} & SpQR & 143.20 & \textbf{6.57} \\
\cmidrule{2-5}
\multicolumn{1}{c}{} & \multicolumn{1}{c}{\multirow{2}{*}{INT-4}} & GPTQ & \textbf{44.11} & \textbf{14.88} \\
\multicolumn{1}{c}{} & \multicolumn{1}{c}{} & SpQR & 143.21 & 6.56 \\
\cmidrule{2-5}
\multicolumn{1}{c}{} & \multicolumn{1}{c}{\multirow{2}{*}{INT-3}} & GPTQ & \textbf{35.93} & 0.84 \\
\multicolumn{1}{c}{} & \multicolumn{1}{c}{} & SpQR & 143.37 & \textbf{6.57} \\
\cmidrule{2-5}
\multicolumn{1}{c}{} & \multicolumn{1}{c}{\multirow{2}{*}{INT-2}} & GPTQ & \textbf{27.74} & 2.23 \\
\multicolumn{1}{c}{} & \multicolumn{1}{c}{} & SpQR & 144.59 & \textbf{6.56} \\
\bottomrule
\end{tabular}
% }
\caption{Memory consumption (in GB) and decoding speed (tokens generated per second) of Qwen-Chat series models and their quantized counterparts during inference. The best results achieved by the quantized models are highlighted in bold.}
\label{tab:flores_ppl_memory_speed}
\end{table*}

\begin{table*}[htbp]
  \centering
  \resizebox{\textwidth}{!}{
    \begin{tabular}{cccccccc}
      \toprule
      \textbf{Model} & \textbf{Datatype} & \textbf{Quantization Method} & \textbf{Accuracy (STEM)} & \textbf{Accuracy (Humanities)} & \textbf{Accuracy (Other)} & \textbf{Accuracy (Social Science)} & \textbf{Accuracy (Average)} \\
      \midrule
      \multirow{10}[9]{*}{Qwen-7B-Chat} & BFloat16 & -     & 50.84  & 48.44  & 63.57  & 64.28  & 55.80  \\
  \cmidrule{2-8}          & \multirow{3}[2]{*}{INT-8} & LLM.int8 & 49.79  & 47.91  & 63.18  & 63.86  & 55.21  \\
            &       & GPTQ  & 50.49  & \textbf{48.03} & {\ul \textbf{63.60}} & \textbf{64.02} & \textbf{55.53} \\
            &       & SpQR  & {\ul \textbf{50.94}} & 47.97  & 62.99  & 63.96  & 55.46  \\
  \cmidrule{2-8}          & \multirow{2}[2]{*}{INT-4} & GPTQ  & 49.44  & 47.29  & 61.47  & 62.63  & 54.27$^\star$  \\
            &       & SpQR  & {\ul \textbf{51.00}} & \textbf{48.20} & \textbf{62.86} & \textbf{63.57} & \textbf{55.44} \\
  \cmidrule{2-8}          & \multirow{2}[2]{*}{INT-3} & GPTQ  & 46.46  & 44.55  & 57.39  & 60.51  & 51.32$^\star$  \\
            &       & SpQR  & \textbf{48.68} & \textbf{47.25} & \textbf{62.70} & \textbf{63.54} & \textbf{54.56$^\star$} \\
  \cmidrule{2-8}          & \multirow{2}[1]{*}{INT-2} & GPTQ  & 22.80  & 24.51  & 25.43  & 21.81  & 23.74$^\star$  \\
            &       & SpQR  & \textbf{46.75} & \textbf{45.55} & \textbf{60.35} & \textbf{61.07} & \textbf{52.49$^\star$} \\ \midrule
      \multirow{10}[8]{*}{Qwen-14B-Chat} & BFloat16 & -     & 61.62  & 56.43  & 71.36  & 73.32  & 64.60  \\
  \cmidrule{2-8}          & \multirow{3}[2]{*}{INT-8} & LLM.int8 & \textbf{61.50} & \textbf{56.20} & \textbf{71.03} & {\ul \textbf{73.68}} & \textbf{64.50} \\
            &       & GPTQ  & 61.02  & 56.13  & 70.87  & {\ul 73.58} & 64.31  \\
            &       & SpQR  & 60.99  & 55.94  & 70.68  & {\ul 73.38} & 64.16  \\
  \cmidrule{2-8}          & \multirow{2}[2]{*}{INT-4} & GPTQ  & 60.32  & 55.71  & 69.13  & 72.64  & 63.42$^\star$  \\
            &       & SpQR  & \textbf{60.70} & \textbf{56.43} & \textbf{70.23} & \textbf{72.99} & \textbf{64.07} \\
  \cmidrule{2-8}          & \multirow{2}[2]{*}{INT-3} & GPTQ  & 56.42  & 52.14  & 65.24  & 68.96  & 59.69$^\star$  \\
            &       & SpQR  & \textbf{59.78} & \textbf{55.81} & \textbf{68.88} & \textbf{72.86} & \textbf{63.33$^\star$} \\
  \cmidrule{2-8}          & \multirow{2}[1]{*}{INT-2} & GPTQ  & 22.42  & 24.87  & 24.04  & 23.95  & 23.94$^\star$  \\
            &       & SpQR  & \textbf{57.41} & \textbf{54.43} & \textbf{67.98} & \textbf{69.81} & \textbf{61.47$^\star$} \\ \midrule
      \multirow{10}[9]{*}{Qwen-72B-Chat} & BFloat16 & -     & 70.25  & 68.50  & 79.98  & 81.74  & 74.33  \\
  \cmidrule{2-8}          & \multirow{3}[2]{*}{INT-8} & LLM.int8 & \textbf{70.06} & 68.42  & 79.47  & 81.64  & 74.13  \\
            &       & GPTQ  & 69.46  & 67.38  & 79.40  & 81.51  & 73.60  \\
            &       & SpQR  & 69.77  & {\ul \textbf{68.52}} & {\ul \textbf{80.14}} & \textbf{81.67} & \textbf{74.26} \\
  \cmidrule{2-8}          & \multirow{2}[2]{*}{INT-4} & GPTQ  & 70.19  & 67.91  & 79.11  & 81.22  & 73.81  \\
            &       & SpQR  & {\ul \textbf{70.98}} & \textbf{68.18} & \textbf{79.95} & {\ul \textbf{81.80}} & {\ul \textbf{74.40}} \\
  \cmidrule{2-8}          & \multirow{2}[2]{*}{INT-3} & GPTQ  & 65.68  & 63.72  & 76.21  & 76.44  & 69.71$^\star$  \\
            &       & SpQR  & \textbf{68.41} & \textbf{68.31} & \textbf{79.47} & {\ul \textbf{81.90}} & \textbf{73.78} \\
  \cmidrule{2-8}          & \multirow{2}[2]{*}{INT-2} & GPTQ  & 23.25  & 27.01  & 24.65  & 23.24  & 24.82$^\star$  \\
            &       & SpQR  & \textbf{66.10} & \textbf{65.06} & \textbf{76.25} & \textbf{79.56} & \textbf{70.94$^\star$} \\
      \bottomrule
      \end{tabular}%
 }
    \caption{Accuracy of Qwen-Chat series models and their quantized counterparts across four broad disciplines (STEM, Humanities, Social Science, and Other) on the MMLU benchmark \citep{DBLP:conf/iclr/HendrycksBBZMSS21}, including overall average accuracy. The best results achieved by the quantized models are highlighted in bold, while underlined results indicate that the performance of the quantized model surpasses that of the BFloat16 baseline. \textbf{Statistically significant differences in the Accuracy (Average) column between quantized LLMs and their non-quantized equivalents are indicated $^\star p < 0.05$.}}
  \label{tab:mmlu}%
\end{table*}%

\begin{table*}[htbp]
  \centering
  \resizebox{\textwidth}{!}{
    \begin{tabular}{ccccccccc}
    \toprule
    \textbf{Model} & \textbf{Datatype} & \textbf{Quantization Method} & \textbf{Accuracy (STEM)} & \textbf{Accuracy (Social Science)} & \textbf{Accuracy (Humanities)} & \textbf{Accuracy (Other)} & \textbf{Accuracy (Hard)} & \textbf{Accuracy (Average)} \\
    \midrule
    \multirow{10}[9]{*}{Qwen-7B-Chat} & BFloat16 & -     & 54.4  & 71.9  & 63    & 52.3  & 40.4  & 59.1 \\
\cmidrule{2-9}          & \multirow{3}[2]{*}{INT-8} & LLM.int8 & \textbf{54.2} & 71.7  & 62.2  & {\ul \textbf{53}} & {\ul \textbf{41.2}} & \textbf{59} \\
          &       & GPTQ  & 53.8  & \textbf{71.8} & 61.8  & 52.2  & 39.6  & 58.6 \\
          &       & SpQR  & 53.4  & 71.3  & \textbf{62.6} & 52.3  & 40.2  & 58.5 \\
\cmidrule{2-9}          & \multirow{2}[2]{*}{INT-4} & GPTQ  & 52    & 70.6  & \textbf{61.4} & 50.7  & \textbf{39.2} & 57.3 \\
          &       & SpQR  & \textbf{53.3} & \textbf{71.5} & 61.3  & \textbf{51.9} & 38.4  & \textbf{58.2} \\
\cmidrule{2-9}          & \multirow{2}[2]{*}{INT-3} & GPTQ  & 44.8  & 64.1  & 54.1  & 45.4  & 32.8  & 50.6 \\
          &       & SpQR  & \textbf{51.8} & \textbf{70.2} & \textbf{60.7} & \textbf{50.5} & \textbf{38.4} & \textbf{57} \\
\cmidrule{2-9}          & \multirow{2}[1]{*}{INT-2} & GPTQ  & 22.9  & 22.7  & 23.6  & 24.1  & 21.1  & 23.2 \\
          &       & SpQR  & \textbf{47.4} & \textbf{66.8} & \textbf{56.9} & \textbf{48} & \textbf{33.1} & \textbf{53.3} \\ \midrule
    \multirow{10}[8]{*}{Qwen-14B-Chat} & BFloat16 & -     & 64.4  & 80.7  & 71.2  & 63.5  & 52.7  & 68.8 \\
\cmidrule{2-9}          & \multirow{3}[2]{*}{INT-8} & LLM.int8 & 64.4  & {\ul 80.9} & 71    & \textbf{63.4} & 52.2  & \textbf{68.8} \\
          &       & GPTQ  & {\ul \textbf{64.6}} & {\ul \textbf{81}} & \textbf{71.2} & 63.1  & \textbf{52.4} & \textbf{68.8} \\
          &       & SpQR  & 63.7  & 80.6  & 70.9  & 62.5  & 52.1  & 68.2 \\
\cmidrule{2-9}          & \multirow{2}[2]{*}{INT-4} & GPTQ  & 62.6  & 80    & 69.3  & 61.8  & 50.4  & 67.2 \\
          &       & SpQR  & \textbf{64.4} & \textbf{80.2} & \textbf{70.4} & \textbf{62.1} & \textbf{51.9} & \textbf{68.2} \\
\cmidrule{2-9}          & \multirow{2}[2]{*}{INT-3} & GPTQ  & 56.1  & 74.2  & 64    & 56.3  & 43.9  & 61.3 \\
          &       & SpQR  & \textbf{63.2} & \textbf{79.7} & \textbf{69.7} & \textbf{62.3} & \textbf{50.4} & \textbf{67.6} \\
\cmidrule{2-9}          & \multirow{2}[1]{*}{INT-2} & GPTQ  & 22.9  & 23.9  & 22.6  & 23.5  & 21.8  & 23.2 \\
          &       & SpQR  & \textbf{58.6} & \textbf{77.6} & \textbf{67.3} & \textbf{56.9} & \textbf{46.5} & \textbf{63.7} \\ \midrule
    \multirow{10}[9]{*}{Qwen-72B-Chat} & BFloat16 & -     & 74.4  & 89.5  & 80.7  & 78.2  & 61.6  & 79.4 \\
\cmidrule{2-9}          & \multirow{3}[2]{*}{INT-8} & LLM.int8 & 73.9  & {\ul \textbf{89.8}} & 80.5  & \textbf{77.8} & 60.6  & \textbf{79.2} \\
          &       & GPTQ  & 71.6  & 89.2  & 80    & 75.5  & 57.7  & 77.6 \\
          &       & SpQR  & \textbf{74} & 89.3  & \textbf{80.7} & 77.4  & \textbf{61.6} & 79.1 \\
\cmidrule{2-9}          & \multirow{2}[2]{*}{INT-4} & GPTQ  & \textbf{72.5} & 89.1  & 80    & 76.6  & 58.9  & 78.1 \\
          &       & SpQR  & \textbf{72.5} & {\ul \textbf{89.6}} & \textbf{80.3} & \textbf{77.7} & \textbf{59} & \textbf{78.5} \\
\cmidrule{2-9}          & \multirow{2}[2]{*}{INT-3} & GPTQ  & 65.3  & 82.5  & 72    & 65.5  & 51.7  & 70.1 \\
          &       & SpQR  & \textbf{72} & \textbf{87.1} & \textbf{79.2} & \textbf{74.6} & \textbf{57.9} & \textbf{76.9} \\
\cmidrule{2-9}          & \multirow{2}[2]{*}{INT-2} & GPTQ  & 25.3  & 25.7  & 25.8  & 25.2  & 25.2  & 25.4 \\
          &       & SpQR  & \textbf{66.7} & \textbf{84.5} & \textbf{74.4} & 67.9  & \textbf{52.7} & \textbf{72} \\
    \bottomrule
    \end{tabular}%
    }
    \caption{Accuracy of Qwen-Chat series models and their quantized counterparts Across four broad disciplines (STEM, Social Sciences, Humanities, and Other) on the C-EVAL Benchmark \citep{huang2023ceval}, including the C-EVAL Hard Subset and the average accuracy across all disciplines. The best results achieved by the quantized models are highlighted in bold, while underlined results indicate that the performance of the quantized model surpasses that of the BFloat16 baseline.}
    
    \label{tab:c-eval}%
\end{table*}%

\begin{table*}[htbp]
    \centering
  \resizebox{\textwidth}{!}{
      \begin{tabular}{ccccccc}
      \toprule
      \textbf{Model} & \textbf{Datatype} & \textbf{Quantization Method} & \textbf{ROUGE-1} & \textbf{ROUGE-2} & \textbf{ROUGE-L} \\
      \midrule
      \multirow{10}[9]{*}{Qwen-7B-Chat} & BFloat16 & -     & 0.19 & 0.05 & 0.13 \\
  \cmidrule{2-6}          & \multirow{3}[2]{*}{INT-8} & LLM.int8 & 0.19 & \underline{0.05} & \underline{0.13} \\
            &       & GPTQ  & \underline{\textbf{\textbf{0.19}}} & \underline{\textbf{0.05}} & \underline{\textbf{0.13}} \\
            &       & SpQR  & 0.19 & 0.05$^\star$ & 0.13 \\
  \cmidrule{2-6}          & \multirow{2}[2]{*}{INT-4} & GPTQ  & \textbf{0.19$^\star$} & \textbf{0.05$^\star$} & \textbf{0.13} \\
            &       & SpQR  & 0.19$^\star$ & 0.05$^\star$ & 0.13$^\star$ \\
  \cmidrule{2-6}          & \multirow{2}[2]{*}{INT-3} & GPTQ  & 0.17$^\star$ & 0.04$^\star$ & 0.12$^\star$ \\
            &       & SpQR  & \underline{\textbf{\textbf{0.19}}} & \textbf{0.05} & \underline{\textbf{\textbf{0.13$^\star$}}} \\
  \cmidrule{2-6}          & \multirow{2}[1]{*}{INT-2} & GPTQ  & 0.04$^\star$ & 0.00$^\star$ & 0.03$^\star$ \\
            &       & SpQR  & \textbf{0.17$^\star$} & \textbf{0.04$^\star$} & \textbf{0.12$^\star$} \\
  \midrule
      \multirow{10}[8]{*}{Qwen-14B-Chat} & BFloat16 & -     & 0.18 & 0.05 & 0.13 \\
  \cmidrule{2-6}          & \multirow{3}[2]{*}{INT-8} & LLM.int8 & \underline{\textbf{0.18}} & 0.05 & 0.13 \\
            &       & GPTQ  & 0.18 & \textbf{0.05} & 0.13 \\
            &       & SpQR  & \underline{\textbf{\textbf{0.18}}} & 0.05 & \textbf{0.13} \\
  \cmidrule{2-6}          & \multirow{2}[2]{*}{INT-4} & GPTQ  & \textbf{0.18$^\star$} & \textbf{0.05$^\star$} & \textbf{0.13$^\star$} \\
            &       & SpQR  & 0.18$^\star$ & 0.05$^\star$ & 0.13$^\star$ \\
  \cmidrule{2-6}          & \multirow{2}[2]{*}{INT-3} & GPTQ  & 0.17$^\star$ & 0.04$^\star$ & 0.12$^\star$ \\
            &       & SpQR  & \textbf{0.18$^\star$} & \textbf{0.05$^\star$} & \textbf{0.13$^\star$} \\
  \cmidrule{2-6}          & \multirow{2}[1]{*}{INT-2} & GPTQ  & 0.04$^\star$ & 0.00$^\star$ & 0.03$^\star$ \\
            &       & SpQR  & \textbf{0.18$^\star$} & \textbf{0.05$^\star$} & \textbf{0.12$^\star$} \\
  \midrule
      \multirow{10}[9]{*}{Qwen-72B-Chat} & BFloat16 & -     & 0.25 & 0.09 & 0.18 \\
  \cmidrule{2-6}          & \multirow{3}[2]{*}{INT-8} & LLM.int8 & \underline{0.25} & \underline{\textbf{\textbf{0.10}}} & \underline{0.19} \\
            &       & GPTQ  & \underline{\textbf{0.26}} & \underline{0.09} & \underline{\textbf{\textbf{0.19$^\star$}}} \\
            &       & SpQR  & 0.25 & \underline{0.09} & \underline{0.19} \\
  \cmidrule{2-6}          & \multirow{2}[2]{*}{INT-4} & GPTQ  & 0.25$^\star$ & 0.09$^\star$ & 0.18$^\star$ \\
            &       & SpQR  & \textbf{0.25$^\star$} & \textbf{0.09} & \textbf{0.18$^\star$} \\
  \cmidrule{2-6}          & \multirow{2}[2]{*}{INT-3} & GPTQ  & 0.20$^\star$ & 0.06$^\star$ & 0.14$^\star$ \\
            &       & SpQR  & \textbf{0.24$^\star$} & \textbf{0.08$^\star$} & \textbf{0.17$^\star$} \\
  \cmidrule{2-6}          & \multirow{2}[2]{*}{INT-2} & GPTQ  & 0.02$^\star$ & {0.00$^\star$} & 0.01$^\star$ \\
            &       & SpQR  & \textbf{0.22$^\star$} & \textbf{0.07$^\star$} & \textbf{0.16$^\star$} \\
      \bottomrule
      \end{tabular}%
  }
  \caption{ROUGE-1, ROUGE-2, and ROUGE-L scores for the Qwen-Chat series models and their quantized counterparts on the test sets of XSum \citep{DBLP:conf/emnlp/NarayanCL18}. The best results achieved by the quantized models are highlighted in bold, while underlined results indicate that the performance of the quantized model surpasses that of the BFloat16 baseline. \textbf{Statistically significant differences between quantized LLMs and their non-quantized equivalents are indicated by $^\star p < 0.05$.}}
    \label{tab:xsum}%
  \end{table*}%

  \begin{table*}[htbp]
    \centering
  \resizebox{\textwidth}{!}{
      \begin{tabular}{ccccccc}
      \toprule
      \textbf{Model} & \textbf{Datatype} & \textbf{Quantization Method} & {\textbf{ROUGE-1}} & {\textbf{ROUGE-2}} & {\textbf{ROUGE-L}} \\
      \midrule
      \multirow{10}[9]{*}{Qwen-7B-Chat} & BFloat16 & -     & 0.33 & 0.12 & 0.21 \\
  \cmidrule{2-6}          & \multirow{3}[2]{*}{INT-8} & LLM.int8 & 0.32 & 0.12 & 0.21 \\
            &       & GPTQ  & \underline{0.33} & \underline{0.12$^\star$} & \underline{0.21} \\
            &       & SpQR  & \underline{\textbf{\textbf{0.33$^\star$}}} & \underline{\textbf{\textbf{0.12$^\star$}}} & \underline{\textbf{\textbf{0.21$^\star$}}} \\
  \cmidrule{2-6}          & \multirow{2}[2]{*}{INT-4} & GPTQ  & 0.32$^\star$ & 0.12$^\star$ & 0.20$^\star$ \\
            &       & SpQR  & \underline{\textbf{\textbf{0.33}}} & \underline{\textbf{\textbf{0.12}}} & \underline{\textbf{\textbf{0.21}}} \\
  \cmidrule{2-6}          & \multirow{2}[2]{*}{INT-3} & GPTQ  & 0.29$^\star$ & 0.10$^\star$ & 0.18$^\star$ \\
            &       & SpQR  & \textbf{0.32$^\star$} & \textbf{0.11$^\star$} & \textbf{0.20$^\star$} \\
  \cmidrule{2-6}          & \multirow{2}[1]{*}{INT-2} & GPTQ  & 0.05$^\star$ & 0.00$^\star$ & 0.03$^\star$ \\
            &       & SpQR  & \textbf{0.31$^\star$} & \textbf{0.11$^\star$} & \textbf{0.20$^\star$} \\
  \midrule
      \multirow{10}[8]{*}{Qwen-14B-Chat} & BFloat16 & -     & 0.32 & 0.12 & 0.21 \\
  \cmidrule{2-6}          & \multirow{3}[2]{*}{INT-8} & LLM.int8 & 0.32 & 0.12 & 0.21 \\
            &       & GPTQ  & \underline{0.32} & 0.12 & \underline{0.21} \\
            &       & SpQR  & \underline{\textbf{\textbf{0.33$^\star$}}} & \underline{\textbf{\textbf{0.12}}} & \underline{\textbf{\textbf{0.21$^\star$}}} \\
  \cmidrule{2-6}          & \multirow{2}[2]{*}{INT-4} & GPTQ  & 0.32$^\star$ & 0.11$^\star$ & 0.20$^\star$ \\
            &       & SpQR  & \textbf{0.32} & \textbf{0.12} & \textbf{0.21} \\
  \cmidrule{2-6}          & \multirow{2}[2]{*}{INT-3} & GPTQ  & 0.30$^\star$ & 0.10$^\star$ & 0.19$^\star$ \\
            &       & SpQR  & \textbf{0.32} & \textbf{0.12$^\star$} & \textbf{0.21} \\
  \cmidrule{2-6}          & \multirow{2}[1]{*}{INT-2} & GPTQ  & 0.05$^\star$ & 0.00$^\star$ & 0.03$^\star$ \\
            &       & SpQR  & \textbf{0.32$^\star$} & \textbf{0.11$^\star$} & \textbf{0.20$^\star$} \\
  \midrule
      \multirow{10}[9]{*}{Qwen-72B-Chat} & BFloat16 & -     & 0.35 & 0.13 & 0.22 \\
  \cmidrule{2-6}          & \multirow{3}[2]{*}{INT-8} & LLM.int8 & 0.35 & \underline{0.13} & 0.22 \\
            &       & GPTQ  & 0.34$^\star$ & 0.13$^\star$ & 0.22$^\star$ \\
            &       & SpQR  & \underline{\textbf{\textbf{0.35}}} & \underline{\textbf{\textbf{0.13}}} & \underline{\textbf{\textbf{0.22}}} \\
  \cmidrule{2-6}          & \multirow{2}[2]{*}{INT-4} & GPTQ  & \underline{\textbf{0.35}} & \underline{\textbf{0.14}} & \underline{\textbf{\textbf{0.22}}} \\
            &       & SpQR  & \underline{0.35} & \underline{0.14} & 0.22 \\
  \cmidrule{2-6}          & \multirow{2}[2]{*}{INT-3} & GPTQ  & 0.33$^\star$ & 0.13$^\star$ & 0.21$^\star$ \\
            &       & SpQR  & \textbf{0.35$^\star$} & \textbf{0.13} & \textbf{0.22} \\
  \cmidrule{2-6}          & \multirow{2}[2]{*}{INT-2} & GPTQ  & 0.02$^\star$ & {0.00$^\star$} & 0.02$^\star$ \\
            &       & SpQR  & \textbf{0.33$^\star$} & \textbf{0.13$^\star$} & \textbf{0.21$^\star$} \\
      \bottomrule
      \end{tabular}%
  }
  \caption{ROUGE-1, ROUGE-2, and ROUGE-L scores of the Qwen-Chat series models and their quantized counterparts on the test sets of CNN/DailyMail \citep{DBLP:conf/acl/SeeLM17}. The best results achieved by the quantized models are highlighted in bold, while underlined results indicate that the performance of the quantized model surpasses that of the BFloat16 baseline. \textbf{Statistically significant differences between quantized LLMs and their non-quantized equivalents are indicated by $^\star p < 0.05$.}}
    \label{tab:cnn}%
  \end{table*}%

% Table generated by Excel2LaTeX from sheet ''
\begin{table*}[htbp]
    \centering
  \resizebox{\textwidth}{!}{
      \begin{tabular}{cccccc}
      \toprule
      \textbf{Model} & \textbf{Datatype} & \textbf{Quantization Method} & \textbf{Accuracy (GSM8K)} & \textbf{Accuracy (SNLI)} & \textbf{MC1 Accuracy (TruthfulQA)} \\
      \midrule
      \multirow{10}[9]{*}{Qwen-7B-Chat} & BFloat16 & -     & 0.51 & 0.82 & 0.38 \\
  \cmidrule{2-6}          & \multirow{3}[2]{*}{INT-8} & LLM.int8 & \underline{0.52} & \textbf{0.81$^\star$} & 0.37 \\
            &       & GPTQ  & \underline{\textbf{\textbf{0.54}}} & \textbf{0.81} & \textbf{0.38} \\
            &       & SpQR  & 0.51 & 0.80$^\star$ & \textbf{0.38} \\
  \cmidrule{2-6}          & \multirow{2}[2]{*}{INT-4} & GPTQ  & 0.47 & \textbf{0.80$^\star$} & 0.36 \\
            &       & SpQR  & \underline{\textbf{\textbf{0.52}}} & 0.80$^\star$ & \textbf{0.37} \\
  \cmidrule{2-6}          & \multirow{2}[2]{*}{INT-3} & GPTQ  & 0.39$^\star$ & \textbf{0.80$^\star$} & 0.36 \\
            &       & SpQR  & \textbf{0.49} & 0.80$^\star$ & \textbf{0.37} \\
  \cmidrule{2-6}          & \multirow{2}[1]{*}{INT-2} & GPTQ  & 0.04$^\star$ & 0.02$^\star$ & 0.30$^\star$ \\
            &       & SpQR  & \textbf{0.44$^\star$} & \textbf{0.80$^\star$} & \textbf{0.36} \\
  \midrule
      \multirow{10}[8]{*}{Qwen-14B-Chat} & BFloat16 & -     & 0.62 & 0.80   & 0.39 \\
  \cmidrule{2-6}          & \multirow{3}[2]{*}{INT-8} & LLM.int8 & 0.60   & 0.80 & \underline{\textbf{\textbf{0.40}}} \\
            &       & GPTQ  & 0.61 & \underline{0.80} & 0.39 \\
            &       & SpQR  & \textbf{0.62} & \underline{\textbf{\textbf{0.81}}} & \underline{0.39} \\
  \cmidrule{2-6}          & \multirow{2}[2]{*}{INT-4} & GPTQ  & 0.60 & 0.79$^\star$ & 0.38 \\
            &       & SpQR  & \textbf{0.61} & \underline{\textbf{\textbf{0.81}}} & \underline{\textbf{\textbf{0.39}}} \\
  \cmidrule{2-6}          
  & \multirow{2}[2]{*}{INT-3} & GPTQ  & 0.51$^\star$ & \underline{\textbf{\textbf{0.81$^\star$}}} & \textbf{0.39} \\
                      &       & SpQR  & \textbf{0.59} & 0.79 & 0.38 \\
  \cmidrule{2-6}          
  & \multirow{2}[1]{*}{INT-2} & GPTQ  & 0.06$^\star$ & 0.03$^\star$ & 0.29 \\
                      &       & SpQR  & \textbf{0.56$^\star$} & \textbf{0.78$^\star$} & \underline{\textbf{\textbf{0.39}}} \\
  \midrule
      \multirow{10}[9]{*}{Qwen-72B-Chat} & BFloat16 & -     & 0.78 & 0.84 & 0.43 \\
  \cmidrule{2-6}          & \multirow{3}[2]{*}{INT-8} & LLM.int8 & \underline{\textbf{\textbf{0.79}}} & 0.83 & 0.43 \\
            &       & GPTQ  & 0.77 & \underline{\textbf{\textbf{0.85$^\star$}}} & 0.42 \\
            &       & SpQR  & \underline{\textbf{0.79}} & 0.84 & \underline{\textbf{\textbf{0.43}}} \\
  \cmidrule{2-6}          & \multirow{2}[2]{*}{INT-4} & GPTQ  & \underline{\textbf{\textbf{0.79}}} & 0.83 & 0.43 \\
            &       & SpQR  & \underline{0.78} & \underline{\textbf{\textbf{0.84}}} & \underline{\textbf{\textbf{0.44}}} \\
  \cmidrule{2-6}          & \multirow{2}[2]{*}{INT-3} & GPTQ  & 0.71$^\star$ & 0.82$^\star$ & 0.41 \\
            &       & SpQR  & \textbf{0.76} & \textbf{0.84} & \textbf{0.42} \\
  \cmidrule{2-6}          & \multirow{2}[2]{*}{INT-2} & GPTQ  & 0.13$^\star$ & 0.01$^\star$ & 0.28$^\star$ \\
            &       & SpQR  & \textbf{0.72$^\star$} & \textbf{0.79$^\star$} & \textbf{0.41} \\
      \bottomrule
      \end{tabular}%
  }
    \caption{Performance of Qwen-Chat series models and their quantized counterparts on the TruthfulQA benchmark \citep{DBLP:conf/acl/LinHE22}, as well as the test sets of GSM8K \citep{DBLP:journals/corr/abs-2110-14168} and SNLI \citep{DBLP:conf/emnlp/BowmanAPM15}. The best results achieved by the quantized models are highlighted in bold, while underlined results indicate that the performance of the quantized model surpasses that of the BFloat16 baseline. \textbf{Statistically significant differences between quantized LLMs and their non-quantized equivalents are indicated by $^\star p < 0.05$.}}
    \label{tab:acc-task}%
  \end{table*}%

\begin{table*}[htbp]
  \centering
\resizebox{\textwidth}{!}{
    \begin{tabular}{cccccc}
    \toprule
    \textbf{Models} & \textbf{Datatypes} & \textbf{Quantization Methods} & \textbf{Hard Satisfaction Rate} & \textbf{Soft Satisfaction Rate} & \textbf{Consistent Satisfaction Levels} \\
    \midrule
    \multirow{10}[9]{*}{Qwen-7B-Chat} & BFloat16 & -     & 0.40  & 0.52  & 1.57  \\
\cmidrule{2-6}          & \multirow{3}[2]{*}{INT-8} & LLM.int8 & {\ul 0.41 } & {\ul 0.52 } & {\ul \textbf{1.62 }} \\
          &       & GPTQ  & {\ul \textbf{0.41 }} & {\ul \textbf{0.53 }} & 1.52  \\
          &       & SpQR  & 0.40  & {\ul 0.52 } & {\ul \textbf{1.62 }} \\
\cmidrule{2-6}          & \multirow{2}[2]{*}{INT-4} & GPTQ  & 0.40  & {\ul \textbf{0.52 }} & 1.38  \\
          &       & SpQR  & {\ul \textbf{0.40 }} & 0.52  & \textbf{1.48 } \\
\cmidrule{2-6}          & \multirow{2}[2]{*}{INT-3} & GPTQ  & 0.36  & 0.48  & 1.27  \\
          &       & SpQR  & \textbf{0.36 } & \textbf{0.49 } & \textbf{1.30 } \\
\cmidrule{2-6}          & \multirow{2}[1]{*}{INT-2} & GPTQ  & 0.00  & 0.01  & 0.00  \\
          &       & SpQR  & \textbf{0.38 } & \textbf{0.50 } & \textbf{1.55 } \\
\midrule
    \multirow{10}[8]{*}{Qwen-14B-Chat} & BFloat16 & -     & 0.47  & 0.57  & 1.73  \\
\cmidrule{2-6}          & \multirow{3}[2]{*}{INT-8} & LLM.int8 & {\ul 0.48 } & {\ul 0.58 } & {\ul 1.75 } \\
          &       & GPTQ  & {\ul \textbf{0.49 }} & {\ul \textbf{0.59 }} & {\ul \textbf{1.90 }} \\
          &       & SpQR  & {\ul 0.48 } & {\ul 0.58 } & {\ul 1.87 } \\
\cmidrule{2-6}          & \multirow{2}[2]{*}{INT-4} & GPTQ  & {\ul \textbf{0.49 }} & {\ul \textbf{0.57 }} & {\ul \textbf{1.90 }} \\
          &       & SpQR  & 0.47  & 0.56  & {\ul 1.78 } \\
\cmidrule{2-6}          & \multirow{2}[2]{*}{INT-3} & GPTQ  & 0.44  & 0.53  & 1.62  \\
          &       & SpQR  & {\ul \textbf{0.48 }} & {\ul \textbf{0.58 }} & {\ul \textbf{1.85 }} \\
\cmidrule{2-6}          & \multirow{2}[1]{*}{INT-2} & GPTQ  & 0.01  & 0.01  & 0.02  \\
          &       & SpQR  & {\ul \textbf{0.48 }} & {\ul \textbf{0.57 }} & {\ul \textbf{1.82 }} \\
\midrule
    \multirow{10}[9]{*}{Qwen-72B-Chat} & BFloat16 & -     & 0.53  & 0.62  & 2.15  \\
\cmidrule{2-6}          & \multirow{3}[2]{*}{INT-8} & LLM.int8 & {\ul \textbf{0.56 }} & {\ul \textbf{0.64 }} & {\ul 2.28 } \\
          &       & GPTQ  & {\ul 0.55 } & {\ul 0.63 } & 2.08  \\
          &       & SpQR  & {\ul 0.54 } & {\ul 0.63 } & {\ul \textbf{2.33 }} \\
\cmidrule{2-6}          & \multirow{2}[2]{*}{INT-4} & GPTQ  & 0.53  & 0.61  & \textbf{2.13 } \\
          &       & SpQR  & {\ul \textbf{0.54 }} & \textbf{0.62 } & 2.10  \\
\cmidrule{2-6}          & \multirow{2}[2]{*}{INT-3} & GPTQ  & 0.53  & 0.61  & 2.02  \\
          &       & SpQR  & \textbf{0.53 } & \textbf{0.61 } & {\ul \textbf{2.22 }} \\
\cmidrule{2-6}          & \multirow{2}[2]{*}{INT-2} & GPTQ  & 0.00  & 0.00  & 0.00  \\
          &       & SpQR  & {\ul \textbf{0.53 }} & \textbf{0.62 } & {\ul \textbf{2.20 }} \\
    \bottomrule
    \end{tabular}%
}
\caption{Average hard satisfaction rates, soft satisfaction rates, and consistent satisfaction levels across five difficulty levels for the Qwen-Chat series models and their quantized counterparts on the FollowBench \citep{DBLP:journals/corr/abs-2310-20410} benchmark. The best results achieved by the quantized models are highlighted in bold, while underlined results indicate that the performance of the quantized model surpasses that of the BFloat16 baseline.}
  \label{tab:addlabel}%
\end{table*}%

\begin{table*}[htbp]
  \centering
  
    \resizebox{\textwidth}{!}{
    \begin{tabular}{ccccc}
    \toprule
    \textbf{Model} & \textbf{Datatype} & \textbf{Quantization Method} & \textbf{Bias Score (Ambiguous Context)} & \textbf{Bias Score (Disambiguated Context)}  \\
    \midrule
    \multirow{10}[9]{*}{Qwen-7B-Chat} & BFloat16 & -     & 6.20   & 3.87  \\
\cmidrule{2-5}          & \multirow{3}[2]{*}{INT-8} & LLM.int8 & 6.41  & {\ul \textbf{3.49}} \\
          &       & GPTQ  & {\ul \textbf{5.98}} & {\ul 3.76} \\
          &       & SpQR  & 6.36  & 3.95 \\
\cmidrule{2-5}          & \multirow{2}[2]{*}{INT-4} & GPTQ  & {\ul 5.49} & \textbf{3.69} \\
          &       & SpQR  & 6.34  & {\ul 3.77} \\
\cmidrule{2-5}          & \multirow{2}[2]{*}{INT-3} & GPTQ  & \textbf{4.21} & 4.90  \\
          &       & SpQR  & 6.31  & \textbf{4.08} \\
\cmidrule{2-5}          & \multirow{2}[1]{*}{INT-2} & GPTQ  & {\ul \textbf{-0.54}} & {\ul \textbf{-0.97}} \\
          &       & SpQR  & {\ul 4.13} & 5.76 \\
\midrule
    \multirow{10}[8]{*}{Qwen-14B-Chat} & BFloat16 & -     & 8.35  & 3.69 \\
\cmidrule{2-5}          & \multirow{3}[2]{*}{INT-8} & LLM.int8 & {\ul \textbf{7.92}} & 3.89 \\
          &       & GPTQ  & {\ul 8.22} & 3.70 \\
          &       & SpQR  & 8.60   & {\ul \textbf{3.65}} \\
\cmidrule{2-5}          & \multirow{2}[2]{*}{INT-4} & GPTQ  & {\ul \textbf{7.82}} & 4.11 \\
          &       & SpQR  & {\ul 7.96} & \textbf{3.86} \\
\cmidrule{2-5}          & \multirow{2}[2]{*}{INT-3} & GPTQ  & 8.41  & 3.88 \\
          &       & SpQR  & {\ul \textbf{8.03}} & {\ul \textbf{3.11}} \\
\cmidrule{2-5}          & \multirow{2}[1]{*}{INT-2} & GPTQ  & {\ul \textbf{-0.17}} & {\ul \textbf{-0.81}} \\
          &       & SpQR  & {\ul 8.08} & 5.33 \\
\midrule
    \multirow{10}[9]{*}{Qwen-72B-Chat} & BFloat16 & -     & 9.07  & 1.57 \\
\cmidrule{2-5}          & \multirow{3}[2]{*}{INT-8} & LLM.int8 & {\ul \textbf{8.81}} & 1.68 \\
          &       & GPTQ  & {\ul 8.95} & {\ul \textbf{1.31}} \\
          &       & SpQR  & 9.07  & {\ul 1.51} \\
\cmidrule{2-5}          & \multirow{2}[2]{*}{INT-4} & GPTQ  & 10.11 & 1.76 \\
          &       & SpQR  & {\ul \textbf{8.73}} & {\ul \textbf{1.52}} \\
\cmidrule{2-5}          & \multirow{2}[2]{*}{INT-3} & GPTQ  & {\ul 8.77} & 3.11 \\
          &       & SpQR  & {\ul \textbf{8.24}} & \textbf{1.97} \\
\cmidrule{2-5}          & \multirow{2}[2]{*}{INT-2} & GPTQ  & {\ul \textbf{-0.36}} & {\ul \textbf{0.89}} \\
          &       & SpQR  & {\ul 7.67} & 1.69 \\
    \bottomrule
    \end{tabular}%
}
\caption{Bias scores of the Qwen-Chat series models and their quantized counterparts in ambiguous and disambiguated contexts on the BBQ benchmark \citep{DBLP:conf/acl/ParrishCNPPTHB22}. The best results achieved by the quantized models are highlighted in bold, while underlined results indicate that the performance of the quantized model surpasses that of the BFloat16 baseline.}
  \label{tab:bbq}%
\end{table*}%

\end{document}